\documentclass[10pt,a4paper,fleqn]{article}

\usepackage[
    a4paper,
    margin=1in
]{geometry}

\usepackage{setspace}
\usepackage{indentfirst}

\usepackage{amsmath}
\usepackage{amssymb}
\usepackage{amsfonts}
\usepackage{amsthm}
\usepackage{mathrsfs}
\usepackage{nicefrac}

\usepackage{graphicx}
\usepackage{booktabs}
\usepackage{tabularx}
\usepackage{multirow}
\usepackage{makecell}
\usepackage{float}
\usepackage{subcaption}
\usepackage{caption}

\usepackage{algorithm}
\usepackage{algpseudocode}
\usepackage{listings}

\usepackage{xcolor}
\usepackage{tikz}

\usepackage{csquotes}
\usepackage{textcomp}
\usepackage{enumitem}

\usepackage[title]{appendix}

\usepackage[authoryear,longnamesfirst]{natbib}

\usepackage{authblk}

\usepackage{hyperref}

\hypersetup{
    colorlinks=true,
    linkcolor=black,
    citecolor=black,
    urlcolor=black,
    filecolor=black
}

\usepackage{lineno}

\begin{document}

\title{
Sensitivity-Constrained Neural Operators for Data-Efficient Forward and
Inverse Modeling of Partial Differential Equation Systems
}

\author[1]{Abdolmehdi Behroozi}
\author[1]{Chaopeng Shen\thanks{Corresponding author: \texttt{cshen@engr.psu.edu}}}
\author[2]{Daniel Kifer}
\author[1]{Kathryn Lawson}

\affil[1]{
Department of Civil and Environmental Engineering,
Penn State University,
University Park, PA 16802, USA
}

\affil[2]{
School of Electrical Engineering and Computer Science,
Penn State University,
University Park, PA 16802, USA
}

\date{}

\maketitle

\begin{center}
\small
Abdolmehdi Behroozi: ORCID 0000-0002-7663-8727 \\
Chaopeng Shen: ORCID 0000-0002-0685-1901 \\
Daniel Kifer: ORCID 0000-0002-4611-7066 \\
Kathryn Lawson: ORCID 0000-0003-0075-7911
\end{center}

\begin{abstract}
Neural operators provide fast surrogates for partial differential equation (PDE) solvers, but their reliability can degrade when the inputs are high-dimensional spatial fields and when the surrogate is used for inverse or repeated inference. The central limitation is that state-only training constrains solution values but does not directly constrain the learned input--output response of the operator. This work studies sensitivity-constrained neural operators (SC-NOs), which make each simulated trajectory more informative by augmenting standard neural-operator training with sampled solver-derived Jacobian supervision. The method matches selected sensitivities obtained from differentiable solvers or discrete adjoints, allowing full-field response information to be amortized across minibatch training without imposing the complete Jacobian at every update. We evaluate SC-NO on controlled advection--diffusion and RANS--Spalart--Allmaras benchmarks, empirical input-dimensionality scaling tests, long-horizon autoregressive rollout, and a shallow-water Tohoku tsunami source-inversion case. Across the reported settings, sensitivity supervision improves forward prediction and produces larger gains in gradient-based inverse reconstruction of distributed fields. The scaling experiments show that sampled Jacobian supervision improves the observed accuracy--cost tradeoff for high-dimensional gridded inputs, while ablations indicate that state values and Jacobian information provide complementary forms of supervision. In the tsunami case, a trained SC-FNO reconstructs a gridded seafloor deformation source from sparse early gauge observations and forecasts subsequent wave propagation within a near-real-time proof-of-concept workflow. These results support sampled sensitivity supervision as a practical strategy for improving neural PDE surrogates when forward accuracy, inverse stability, robustness, and computational cost must be considered together.
\end{abstract}

\medskip



\section{Introduction}

High-fidelity partial differential equation (PDE) solvers are central to modern scientific and engineering prediction, including fluid mechanics, geophysical hazards, structural response, environmental transport, and subsurface flow. These solvers provide detailed representations of physical processes, but their computational cost remains a major barrier when rapid forward simulation, repeated inverse inference, or uncertainty-aware prediction is required \citep{leveque2002finite, palais2009differential, quarteroni2010numerical}. The difficulty is amplified when the governing inputs are high-dimensional spatial fields, such as heterogeneous material properties, geometries, bed topography, roughness maps, forcing fields, or source deformations. This computational bottleneck has motivated a broad class of reduced-order, data-driven, and surrogate modeling approaches that seek to approximate the input--output map of a high-fidelity solver at much lower evaluation cost. In this setting, the surrogate is trained to emulate the solution operator defined by the numerical PDE model, replacing repeated expensive solver calls with fast learned predictions. However, inversions or learning-based workflows may still require thousands of model evaluations \citep{li2021fourier}, and the number of degrees of freedom in the input can make purely empirical surrogate training increasingly data-intensive \citep{grady2023model, xiao2024fourier}.

Time-critical geophysical forecasting provides one important example of this broader challenge. In tsunami warning, flood forecasting, plume transport, and related applications, models must assimilate limited observations and produce forecasts within short decision windows. For tsunamis, real-time high-resolution modeling remains difficult \citep{Reymond2012}, so operational systems often rely on precomputed scenario databases and rapid scenario superposition \citep{gica2008development, fujita2024scenario}. Dynamic source-modeling approaches that embed fault slip within high-resolution elasticity solvers can better represent transient rupture effects \citep{vogl2017high}, but they remain computationally intensive. The 2011 Tohoku event also showed that early forecasts can underestimate wave heights when the source and propagation dynamics are not adequately resolved \citep{hoshiba2014earthquake, ozaki2012jma}. These examples illustrate a general computational tension: high-fidelity PDE models are needed most in settings where repeated, rapid forward and inverse evaluations are hardest to afford.

Neural operators offer a promising route for accelerating PDE surrogate modeling. Recent progress in artificial intelligence for scientific computing and differentiable physical modeling has created strong interest in fast surrogate models for PDE-governed systems \citep{zong2026mathematics}. Neural operators, including Fourier Neural Operators (FNOs) \citep{li2021fourier}, DeepONets \citep{lu2021learning}, Wavelet Neural Operators (WNOs) \citep{tripura2023wavelet}, and Convolutional Neural Operators (CNOs) \citep{raonic2023convolutional}, learn mappings between function spaces from numerical simulation data and can provide large acceleration relative to traditional solvers \citep{kovachki2023neural, wang2024deep, qin2024toward}. These models have been used in applications including lithography \citep{yang2022large}, weather forecasting \citep{kurth2023fourcastnet}, fluid mechanics \citep{han2022equivariant}, and subsurface CO$_2$ sequestration \citep{wen2023real}.

Despite this progress, standard neural operators remain vulnerable when the inputs are high-dimensional gridded fields. Purely data-driven training must infer how changes in distributed parameters, geometries, forcing fields, or initial conditions affect the output solution. Recent theoretical work has emphasized the parametric and data complexity of operator learning in such settings \citep{lanthaler2025parametric, kovachki2024data}. In practice, this means that neural operators may require large training datasets, may degrade when gridded inputs change substantially, and may become unreliable in inverse problems where accurate gradients with respect to inputs are essential \citep{kovachki2023neural}. Thus, the relevant issue is not only whether a surrogate predicts solution values accurately on held-out samples, but whether it captures the input--output sensitivity structure needed for stable inference beyond the training distribution.

A central failure mode of data-only neural operators is inaccurate sensitivity, even when forward predictions appear acceptable. A surrogate may reproduce solution fields on the training distribution while learning incorrect local derivatives with respect to the input. These gradient errors can degrade robustness to input perturbations, worsen out-of-distribution behavior, amplify errors during autoregressive rollout, and mislead gradient-based inversion \citep{choi2024applications, behroozi2025sensitivity}. Several approaches have incorporated additional physical or derivative information into surrogate training, including PDE-residual or physics-informed neural-operator losses \citep{li2024physics}, equivariant constraints \citep{han2022equivariant}, Sobolev training \citep{czarnecki2017sobolev}, Sensitivity-Constrained Fourier Neural Operators \citep{behroozi2025sensitivity}, derivative-enhanced DeepONets \citep{qiu2024derivative}, and derivative-informed neural operators \citep{cao2025derivative}. These methods show that derivative information can improve generalization, but many existing demonstrations remain focused on low-dimensional parameterizations or compressed representations. Full-field Jacobian supervision for gridded inputs remains difficult because the Jacobian can be extremely large and costly to impose directly during training.

This work studies sensitivity-constrained neural operators (SC-NOs) for high-dimensional forward and inverse PDE inference. The key idea is to train neural operators not only to match PDE solution fields, but also to match selected entries of physics-derived Jacobians obtained from differentiable solvers or discrete adjoints. Rather than enforcing the full Jacobian at every optimization step, the method samples subsets of Jacobian entries and resamples them across minibatches, amortizing sensitivity supervision over the training process. This provides a practical mechanism for incorporating full-field sensitivity information without requiring the entire Jacobian tensor to be used in every update.

The contribution is not a new neural operator architecture. Instead, the paper evaluates whether scalable sensitivity supervision improves the reliability of existing neural-operator families under high-dimensional gridded inputs. We first study controlled advection--diffusion and turbulent Navier--Stokes benchmarks to examine forward prediction, inverse reconstruction of distributed fields, sample efficiency, and accuracy--cost tradeoffs. We then analyze empirical scaling with input degrees of freedom, Jacobian sampling density, out-of-distribution behavior, and long-horizon rollout stability. Finally, we use a large-scale shallow-water tsunami source-inversion problem as a capstone application, where sparse early gauge observations are used to reconstruct a gridded seafloor deformation source and forecast subsequent wave propagation.

The claims in this paper are empirical and computational rather than asymptotic. We do not claim that sensitivity supervision solves the curse of dimensionality in a formal sample-complexity sense. Instead, the results show that sampled Jacobian supervision can substantially reduce the data burden and improve the observed accuracy--cost tradeoff for the high-dimensional PDE inference problems studied here. The findings support sensitivity supervision as a practical mechanism for improving neural PDE surrogates in settings where forward accuracy, inverse stability, out-of-distribution robustness, and time-critical inference must be considered together.

\section{Sensitivity-Constrained Neural Operators}
\label{sec:scno}

This section defines the sensitivity-constrained neural operator (SC-NO) framework used throughout the paper. The objective is not to introduce a new neural-operator architecture, but to augment existing operator families with supervision on physically derived input--output sensitivities. In this work, the sensitivity information is obtained from differentiable numerical solvers or discrete adjoints and is imposed through sampled Jacobian entries during training.

\subsection{Neural Operator Formulation}
\label{sec:no_formulation}

Let $\mathcal{A}$ denote a space of input functions and $\mathcal{U}$ a space of solution functions. For a PDE-governed system, the high-fidelity numerical solver defines a solution operator
\begin{equation}
    \mathcal{G}: \mathcal{A} \rightarrow \mathcal{U},
    \qquad
    u = \mathcal{G}(a),
\end{equation}
where $a$ represents the problem input and $u$ is the corresponding solution field. Depending on the benchmark, $a$ may include initial states, boundary or forcing information, spatially distributed coefficients, bed topography, or source deformation fields. We write the input generically as $a = (a_{\mathrm{ctx}}, p)$, where $a_{\mathrm{ctx}}$ denotes any state or contextual information provided to the model, and $p$ denotes the input field or parameter with respect to which sensitivities are supervised.

A neural operator approximates $\mathcal{G}$ by a parameterized map
\begin{equation}
    \mathcal{G}_{\theta}: \mathcal{A} \rightarrow \mathcal{U},
    \qquad
    \hat{u} = \mathcal{G}_{\theta}(a),
\end{equation}
where $\theta$ denotes trainable parameters. In a standard neural-operator layer, a latent representation $v_\ell$ at layer $\ell$ is updated through a combination of nonlocal and local transformations,
\begin{equation}
    v_{\ell+1}
    =
    \sigma\!\left(
    \mathcal{K}_{\theta}^{(\ell)} v_\ell
    +
    \mathcal{W}_{\theta}^{(\ell)} v_\ell
    \right),
\end{equation}
where $\mathcal{K}_{\theta}^{(\ell)}$ is a learned nonlocal operator, $\mathcal{W}_{\theta}^{(\ell)}$ is a local transformation, and $\sigma$ is a nonlinear activation. Different choices of $\mathcal{K}_{\theta}^{(\ell)}$ recover different operator architectures, such as Fourier neural operators, wavelet neural operators, or branch--trunk operator models. The sensitivity-constrained formulation below is independent of this architectural choice and can be applied to any differentiable neural operator.

\subsection{Sensitivity-Constrained Training Objective}
\label{sec:sensitivity_objective}

Standard neural-operator training minimizes a state-prediction loss between the surrogate solution $\hat{u}$ and the reference solver solution $u$. For a dataset of $N$ samples,
\begin{equation}
    \mathcal{D}
    =
    \left\{
    \left(a^{(i)}, u^{(i)}, J^{(i)}\right)
    \right\}_{i=1}^{N},
\end{equation}
where $u^{(i)} = \mathcal{G}(a^{(i)})$ and the solver-derived Jacobian $J^{(i)}$ with respect to the supervised input $p^{(i)}$ is defined by
\begin{equation}
    J^{(i)}
    =
    \frac{\partial \mathcal{G}(a^{(i)})}{\partial p^{(i)}}.
\end{equation}
The state loss for minibatch $\mathcal{B}$ is
\begin{equation}
    \mathcal{L}_{u}(\theta)
    =
    \frac{1}{|\mathcal{B}|}
    \sum_{i \in \mathcal{B}}
    \left\|
    \mathcal{G}_{\theta}(a^{(i)}) - u^{(i)}
    \right\|_{2}^{2},
\end{equation}
where $|\mathcal{B}|$ denotes the minibatch size and $\|\cdot\|_{2}$ denotes the Euclidean norm over the discretized solution degrees of freedom.

SC-NO augments this objective by matching the Jacobian of the neural operator to the Jacobian of the reference solver. The model Jacobian is
\begin{equation}
    J_{\theta}^{(i)}
    =
    \frac{\partial \mathcal{G}_{\theta}(a^{(i)})}{\partial p^{(i)}}.
\end{equation}

The sensitivity-constrained objective combines the state-prediction loss and the Jacobian-matching loss. For clarity, it can first be written in the fixed-weight form
\begin{equation}
    \mathcal{L}(\theta)
    =
    \mathcal{L}_{u}(\theta)
    +
    \lambda \mathcal{L}_{J}(\theta),
    \label{eq:scno_loss}
\end{equation}
where $\lambda > 0$ denotes the relative weight assigned to the Jacobian loss. In practice, however, we do not manually tune a fixed $\lambda$. Instead, we use learnable automatic loss weights inspired by uncertainty-based multi-task weighting~\citep{kendall2018multi}. For the loss components $\mathcal{L}_r$, with $r \in \{u,J\}$, the optimized objective is
\begin{equation}
    \mathcal{L}(\theta,\boldsymbol{\sigma})
    =
    \sum_{r \in \{u,J\}}
    \left[
    \frac{1}{2\sigma_r^2}\mathcal{L}_r(\theta)
    +
    \log(1+\sigma_r^2)
    \right],
    \label{eq:learnable_loss_weighting}
\end{equation}
where $\sigma_r$ are trainable scalar weighting parameters optimized jointly with the neural-operator parameters. The logarithmic term prevents the optimization from trivially increasing $\sigma_r$ to suppress a loss component, while the inverse-variance term adaptively balances the state and Jacobian losses during training.

The term $\mathcal{L}_{J}$ penalizes mismatch between selected entries of $J_{\theta}^{(i)}$ and $J^{(i)}$.

In the experiments, $p$ corresponds to the high-dimensional input field relevant to each task, such as a velocity field, initial condition, forcing field, or bed-topography/source-deformation field. The exact supervised quantity is specified for each benchmark in Section~\ref{sec:experimental_design}. A supporting analysis of this objective is provided in Appendix~\ref{sec_sup_method}. The analysis explains how matching solver-derived sensitivities can constrain the learned input--output response of the neural operator. We use this analysis as mechanistic support for the sensitivity-constrained objective, while empirical evidence for scaling behavior, robustness, and stability is evaluated in the results. In particular, the analysis shows that, under boundedness and uniform-convergence assumptions, matching solver-derived sensitivities constrain the learned input--output derivatives of the neural operator. This provides a mechanism for improved generalization, more stable long-horizon rollout, and more reliable gradient-based inversion. We use this result as theoretical support for the sensitivity-constrained objective, while empirical evidence for scaling behavior, robustness, and stability is evaluated separately in the results.

This objective encourages the learned operator to match not only the solution values produced by the numerical solver, but also the local input--output response of the solver. This distinction is important for inverse problems and robustness tests, where optimization or distribution shifts can repeatedly evaluate the surrogate away from the training inputs.

\begin{figure}
    \centering
    \includegraphics[width=0.98\textwidth, trim={0 0 0 0}, clip]{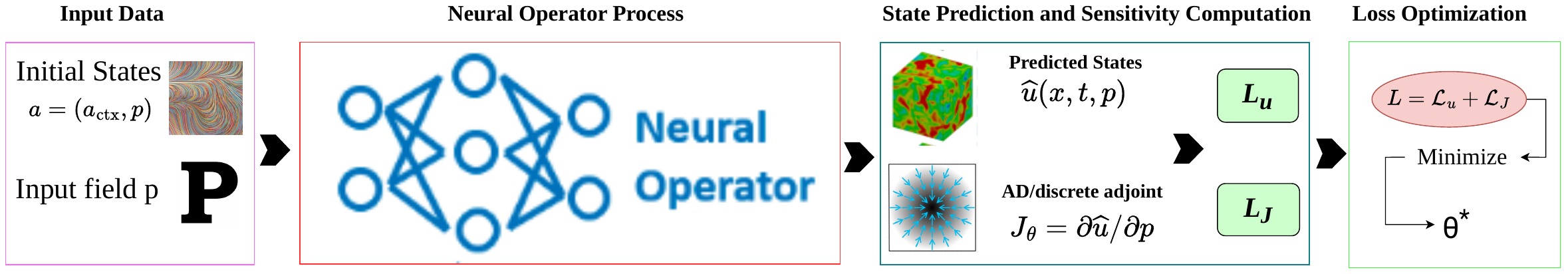}
    \captionsetup{width=0.98\textwidth}
    \caption{\scriptsize\textbf{Overview of the sensitivity-constrained neural-operator framework.}
    The input consists of contextual state information \(a_{\mathrm{ctx}}\) and a supervised input field or parameter \(p\). 
    A neural operator maps the input to predicted solution states \(\widehat{u}(x,t,p)\), where \(x\) and \(t\) denote space and time when applicable. 
    In addition to the state-prediction loss \(\mathcal{L}_u\), the framework computes model sensitivities \(J_\theta=\partial \widehat{u}/\partial p\) using automatic differentiation and matches them to solver-derived sensitivities obtained from automatic differentiation or discrete adjoints. 
    The resulting Jacobian loss \(\mathcal{L}_J\) is combined with the state loss using learnable loss weights, encouraging the surrogate to match both solution values and input--output sensitivities.}    \label{fig:sc_no_frame}
\end{figure}
Figure~\ref{fig:sc_no_frame} summarizes the SC-NO training pipeline. The method augments standard neural-operator training with Jacobian-level supervision from differentiable solvers or discrete adjoints. Rather than imposing the full Jacobian at every optimization step, SC-NO samples subsets of sensitivity entries during minibatch training. These subsets are resampled across iterations, allowing the model to receive sensitivity supervision over many input--output directions while keeping the per-step cost tractable. This design encourages the learned operator to match both the solver states and the local response of those states to high-dimensional input perturbations.

\subsection{Scalable Jacobian Sampling and Amortized Supervision}
\label{sec:jacobian_sampling}

For gridded PDE inputs, the full Jacobian can be extremely large. If the predicted solution has $N_u$ output degrees of freedom and the supervised input field has $N_p$ degrees of freedom, then the full Jacobian contains $N_u N_p$ entries. Directly imposing all entries at every training step is usually impractical in memory and computation.

We therefore use sampled Jacobian supervision. Let $\Omega_i \subset \{1,\ldots,N_u\}$ denote a randomly selected subset of output indices for sample $i$. In each minibatch, the sensitivity loss is formed only from the Jacobian rows associated with $\Omega_i$:
\begin{equation}
    \mathcal{L}_{J}(\theta)
    =
    \frac{1}{|\mathcal{B}|}
    \sum_{i \in \mathcal{B}}
    \frac{1}{|\Omega_i|}
    \left\|
    \mathcal{P}_{\Omega_i}
    \left(
    J_{\theta}^{(i)} - J^{(i)}
    \right)
    \right\|_{F}^{2},
    \label{eq:sampled_jacobian_loss}
\end{equation}
where $\mathcal{P}_{\Omega_i}$ extracts the sampled output rows and $\|\cdot\|_{F}$ denotes the Frobenius norm. The sampled rows are resampled across minibatches and epochs. Under uniform sampling and the normalization in Eq.~\eqref{eq:sampled_jacobian_loss}, this loss is an unbiased estimator of the corresponding full-row Jacobian loss.

This sampling strategy amortizes sensitivity supervision across training. Each individual update uses only a small subset of the full Jacobian, but different portions of the Jacobian are seen over the full optimization trajectory. As a result, the model receives gradient-level supervision over many input--output directions without requiring the complete Jacobian tensor to be loaded or differentiated through at every step. This mechanism is the main practical difference between SC-NO and a direct full-Jacobian training objective.

In the present experiments, the sampled sensitivities are evaluated at selected output states, typically at the final prediction time used for the sensitivity loss. This choice reduces the training burden and targets the accumulated response of the PDE solution to perturbations in the input field. A broader comparison of final-time, multi-time, and task-adaptive sensitivity sampling is left as a practical extension.

\subsection{Sensitivity Computation by Automatic Differentiation and Discrete Adjoints}
\label{sec:sensitivity_computation}

Two Jacobians are needed during SC-NO training: the model Jacobian $J_{\theta}$ and the solver-derived reference Jacobian $J$. The model Jacobian is computed by automatic differentiation through the neural operator. Since $\mathcal{G}_{\theta}$ is differentiable with respect to its inputs, vector--Jacobian products or batched gradient calls can be used to compute the sampled rows of $J_{\theta} = \partial \mathcal{G}_{\theta}(a)/\partial p$. This computation is performed only for the sampled output entries used in Eq.~\eqref{eq:sampled_jacobian_loss}, which keeps the per-update cost bounded by the sampling density.

The reference Jacobian $J$ is obtained from the numerical model. For simpler differentiable solvers, sensitivities can be computed by automatic differentiation through the discretized time-integration procedure. This is the approach used for the advection--diffusion benchmark. For larger or more complex solvers, differentiating through the full forward trajectory can become memory-intensive. In those cases, we use discrete adjoints derived from the time-discretized numerical scheme. This approach is used for the RANS benchmark with Spalart--Allmaras closure and for the finite-volume shallow-water tsunami solver.

Finite differences are not used for the high-dimensional sensitivity targets in this work. Although finite differences can be effective for a small number of scalar parameters, their cost scales with the number of input degrees of freedom and becomes impractical for gridded fields. Discrete adjoints avoid this dependence for each chosen output functional or block of output rows, making them more suitable for the high-dimensional settings considered here. The main text uses only this method-level description. PDE-specific solver equations, adjoint recursions, flux Jacobians, and implementation details are provided in the appendices.

\subsection{Computational Complexity}
\label{sec:computational_complexity}

The computational cost of SC-NO has three components: generating reference solution data, computing solver-derived sensitivities, and training the neural operator with the sampled Jacobian loss. Let $C_{\theta}$ denote the cost of one forward/backward training step for the neural operator using only the state loss, $N_u$ the number of output degrees of freedom, $N_p$ the number of supervised input degrees of freedom, and $m = |\Omega_i|$ the number of sampled output indices used for the Jacobian loss, with $m \ll N_u$. A direct full-Jacobian loss would require forming or accessing $O(N_u N_p)$ sensitivity entries per sample. In contrast, sampled Jacobian supervision uses only $O(mN_p)$ entries per sample in each update. The model-side differentiation cost similarly scales with the number of sampled output rows rather than the full output dimension. Thus, the sampling ratio $m/N_u$ controls the main training-time tradeoff between sensitivity coverage and computational overhead. Table~\ref{tab:symbolic_scno_cost} summarizes the symbolic cost scaling implied by the Jacobian dimensions and the sampled-row training objective. Here and in the table, $O(\cdot)$ denotes asymptotic scaling rather than measured runtime; empirical wall-clock costs are reported in Appendix~\ref{sec:cpu_time}.

\begin{table}[h!]
\centering
\caption{Symbolic cost comparison for state-only and sensitivity-constrained neural-operator training.}
\label{tab:symbolic_scno_cost}
\begin{tabular}{lll}
\toprule
\textbf{Component} & \textbf{Full Jacobian supervision} & \textbf{Sampled SC-NO supervision} \\
\midrule
State loss & $O(C_{\theta})$ & $O(C_{\theta})$ \\
Jacobian entries used per sample & $O(N_u N_p)$ & $O(mN_p)$ \\
Model Jacobian rows per update & $O(N_u)$ & $O(m)$ \\
Sensitivity storage accessed per update & $O(N_u N_p)$ & $O(mN_p)$ \\
Finite-difference solver sensitivities & $O(N_p)$ solver calls & Not used \\
Adjoint-based solver sensitivities & Output-functional/block dependent & Used for selected rows/blocks \\
\bottomrule
\end{tabular}
\end{table}

This complexity reduction does not make Jacobian supervision free. The sensitivity loss increases training time and memory relative to state-only training because it requires differentiating model outputs with respect to inputs and backpropagating the resulting loss through model parameters. In addition, solver-derived Jacobians must be generated or made accessible during data preparation. For this reason, all accuracy--cost comparisons in the paper account for total wall-clock cost, including data generation, Jacobian computation, and model training.

\section{Experimental Design}
\label{sec:experimental_design}

This section defines the benchmark problems, model comparisons, evaluation tasks, metrics, and cost accounting used to assess sensitivity-constrained neural operators. The purpose is to state the experimental protocol before interpreting the results. Controlled PDE benchmarks are used first to evaluate forward prediction, inverse reconstruction, data efficiency, and input-dimensionality effects under interpretable conditions. The large-scale shallow-water tsunami problem is then used later as a capstone application for sparse-observation source inversion and forecasting. Data generation and random-field sampling are described in Appendix~\ref{app:Data_Generation}; PDE-specific numerical solvers and sensitivity calculations are provided in Appendices~\ref{sec:AD}, \ref{sec:NSE-SA}, and \ref{sec:SWE-Tohoku}; metric definitions, hyperparameters, and computational cost breakdowns are reported in Appendices~\ref{app:metrics}, \ref{app:Hyperparameters}, and~\ref{sec:cpu_time}, respectively.

\subsection{Benchmark PDEs and High-Dimensional Inputs}
\label{sec:benchmark_pdes}

We evaluate the framework on three PDE classes with spatially distributed inputs. The first two benchmarks are controlled problems used to study high-dimensional gridded inputs, training sample size, and sensitivity supervision under interpretable conditions. The third benchmark applies the same framework to a larger Tohoku tsunami source-inversion and wave-propagation problem. Throughout this section, \(p\) denotes the input field or fields with respect to which the solver-derived Jacobian is matched in the SC-NO sensitivity loss. Full domain definitions, boundary and initial conditions, numerical solvers, random-field sampling procedures, and sensitivity calculations are provided in the benchmark appendices. Performance-metric definitions are provided in Appendix~\ref{app:metrics}. The supervised sensitivity targets are summarized in Table~\ref{tab:sensitivity_targets}; these targets define the field \(p\) used in the Jacobian-matching loss for each benchmark.

\begin{table}[h!]
\centering
\captionsetup{width=0.95\textwidth}
\caption{Sensitivity targets used for SC-NO training in each benchmark. Here \(p\) denotes the input field or fields with respect to which the solver-derived Jacobian is matched in the sensitivity loss.}
\label{tab:sensitivity_targets}
\small
\setlength{\tabcolsep}{5pt}
\renewcommand{\arraystretch}{1.15}
\begin{tabularx}{0.95\textwidth}{@{}lXX@{}}
\toprule
\textbf{Benchmark} 
& \textbf{Supervised input field(s) \(p\)} 
& \textbf{Jacobian source} \\
\midrule
PDE1 
& Initial concentration \(C_0(x)\) and velocity field \(\mathbf{u}(x)\) 
& Automatic differentiation through the differentiable advection--diffusion solver \\

PDE2 
& Initial vorticity \(\Omega_0(x)\) and forcing field \(f(x;\alpha,\beta)\) 
& Discrete adjoint of the time-discretized RANS--Spalart--Allmaras solver \\

PDE3
& Bed/source deformation field \(z_b(x)\), equivalently \(\Delta z_b(x)\) when \(z_{b0}(x)\) is fixed
& Discrete adjoint of the finite-volume shallow-water solver \\
\bottomrule
\end{tabularx}
\end{table}

\paragraph{PDE1: Advection--diffusion.}
The first benchmark models the transport of a concentration field \(C(x,t)\) under a spatially varying velocity field \(\mathbf{u}(x)\):
\[
    \frac{\partial C}{\partial t}
    +
    \nabla \cdot \bigl(\mathbf{u}(x)C\bigr)
    =
    D\nabla^2 C .
\]
Both the initial concentration \(C_0(x)\) and the velocity field \(\mathbf{u}(x)=[u_x(x),u_y(x)]\) are sampled as spatially coherent random fields. This provides a controlled high-dimensional setting in which the governing dynamics remain relatively interpretable. For the neural-operator task, the model receives the initial solution context together with \(C_0(x)\) and \(\mathbf{u}(x)\), and predicts the remaining concentration trajectory over the forecast interval. The sensitivity loss supervises the Jacobian with respect to both \(C_0(x)\) and \(\mathbf{u}(x)\). Full problem setup, solver details, and the PDE-specific operator-learning formulation are provided in Appendix~\ref{sec:AD}.

\paragraph{PDE2: RANS--Spalart--Allmaras.}
The second benchmark uses a vorticity--streamfunction formulation of the Navier--Stokes equations with Spalart--Allmaras closure. The governing system includes the Poisson relation
\[
    \nabla^2\Psi=-\Omega
\]
and the vorticity evolution equation
\[
    \frac{\partial \Omega}{\partial t}
    +
    \frac{\partial \Psi}{\partial x_2}
    \frac{\partial \Omega}{\partial x_1}
    -
    \frac{\partial \Psi}{\partial x_1}
    \frac{\partial \Omega}{\partial x_2}
    =
    \nabla\cdot\bigl[(\nu+\nu_t)\nabla\Omega\bigr]
    +
    f(x;\alpha,\beta).
\]
The high-dimensional inputs are the initial vorticity field \(\Omega_0(x)\) and the spatially distributed forcing field \(f(x;\alpha,\beta)\). For the multi-step prediction task, the neural operator receives a short context of the state trajectory together with the forcing field and predicts the remaining trajectory. For the scaling experiments, the intrinsic resolution of \(\Omega_0(x)\) is varied while the training grid is held fixed, allowing the effect of input dimensionality to be isolated from the numerical resolution of the learned solution field. For the rollout task, a one-step transition operator is trained and then applied autoregressively. The sensitivity loss supervises the Jacobian with respect to \(\Omega_0(x)\) and \(f(x;\alpha,\beta)\). Full problem setup, solver details, adjoint sensitivity formulation, and operator-learning settings are provided in Appendix~\ref{sec:NSE-SA}.

\paragraph{PDE3: Tohoku tsunami benchmark via the shallow water equations.}
The third benchmark models the 2011 Tohoku tsunami using the two-dimensional shallow water equations with bathymetry, earthquake-induced seafloor deformation, and friction:
\[
    \frac{\partial h}{\partial t}
    +
    \nabla\cdot(h\mathbf{u})
    =
    0,
\]
\[
    \frac{\partial(h\mathbf{u})}{\partial t}
    +
    \nabla\cdot
    \left(
    h\mathbf{u}\otimes\mathbf{u}
    +
    \frac{1}{2}gh^2 I
    \right)
    =
    -gh\nabla z_b
    +
    \mathrm{Friction}.
\]
Here, \(h\) denotes water depth, \(\mathbf{u}\) is the depth-averaged velocity, and \(z_b(x)\) is the bed elevation. The benchmark simulates tsunami wave propagation over the Tohoku bathymetry following earthquake-induced seafloor deformation. The high-dimensional source input is represented through the bed-topography/deformation field
\[
    z_b(x)=z_{b0}(x)+\Delta z_b(x),
\]
where \(z_{b0}(x)\) is the fixed reference bathymetry and \(\Delta z_b(x)\) is the earthquake-induced seafloor deformation. For the neural-operator task, the model receives the deformation field together with an initial water-stage context and predicts the subsequent water-stage evolution on the neural-operator grid. Since \(z_{b0}(x)\) is fixed, sensitivity with respect to \(z_b(x)\) is equivalent to sensitivity with respect to \(\Delta z_b(x)\). This benchmark is used as the capstone application for tsunami source inversion, forecasting, noise robustness, and spatial OOD evaluation. Full solver validation, Okada source construction, mesh-to-grid mapping, adjoint sensitivity calculation, and inversion details are provided in Appendix~\ref{sec:SWE-Tohoku}.

\subsection{Baselines and Sensitivity-Constrained Variants}
\label{sec:baselines_variants}

We compare standard neural operators against their sensitivity-constrained counterparts. The baseline families are Fourier Neural Operators (FNO)~\citep{li2021fourier}, Wavelet Neural Operators (WNO)~\citep{tripura2023wavelet}, and DeepONet~\citep{lu2021learning}. For each family, the sensitivity-constrained variant is trained with the same state-prediction objective as the corresponding baseline, augmented with a Jacobian-matching loss using solver-derived sensitivities. This gives the paired comparisons
\[
    \mathrm{FNO} \leftrightarrow \mathrm{SC\text{-}FNO}, 
    \qquad
    \mathrm{WNO} \leftrightarrow \mathrm{SC\text{-}WNO},
    \qquad
    \mathrm{DeepONet} \leftrightarrow \mathrm{SC\text{-}DeepONet}.
\]

The main text emphasizes SC-FNO because FNO is the primary operator architecture used throughout the central scaling, robustness, and tsunami experiments. The WNO and DeepONet variants are retained as architectural controls to test whether the effect of sensitivity supervision is specific to one model family or persists across different operator parameterizations. Their detailed quantitative comparisons are reported in the benchmark-specific appendices.

These comparisons do not introduce a new neural-operator architecture. The architectural components remain those of the underlying baseline models. The experimental question is whether adding sampled Jacobian supervision improves data efficiency, inverse stability, OOD behavior, and rollout stability relative to state-only training under the same benchmark conditions. Model configurations, training settings, and hyperparameters for all baseline and sensitivity-constrained variants are reported in Appendix~\ref{app:Hyperparameters}.

\subsection{Task Definitions and Evaluation Settings}
\label{sec:tasks}

We evaluate four task classes.

\paragraph{Forward prediction.}
In the forward setting, a model is trained to approximate the solution operator
\[
    \mathcal{G}_{\theta}: a(x,t,p) \mapsto u(x,t,p),
\]
where the input \(a(x,t,p)\) contains the available context states and the spatially distributed input field(s) \(p(x)\). The model predicts the solution over the remaining time interval. This task evaluates direct surrogate accuracy under the training and test distributions.

\paragraph{Inverse reconstruction.}
In the inverse setting, a pretrained neural operator is treated as a differentiable surrogate inside an optimization loop. To avoid overloading notation, we denote the unknown inversion variable by \(q\), which may be one of the supervised input fields or a lower-dimensional parameterization of that field. The inverse problem is written as
\[
    \min_{q}
    \left\|
    \mathcal{G}_{\theta}(a_{\mathrm{ctx}},q)
    -
    u_{\mathrm{target}}
    \right\|_2^2 .
\]
For the controlled benchmarks, this evaluates reconstruction of distributed input fields such as \(C_0(x)\) or \(\Omega_0(x)\). For the tsunami benchmark, the same principle is used in a sparse-observation setting to infer a seafloor deformation source from early gauge observations before forecasting the subsequent wave evolution. The full tsunami inversion workflow is described in Section~\ref{sec:tsunami}.

\paragraph{Out-of-distribution evaluation.}
OOD evaluation is used to test whether learned operators remain reliable when the test inputs differ from those seen during training. In the tsunami benchmark, OOD cases are defined by earthquake epicenter configurations drawn from spatial regions disjoint from the training epicenter region. This task is important because inverse optimization and forecasting workflows can evaluate the surrogate in parts of input space that are not well represented by the training data.

\paragraph{Long-horizon autoregressive rollout.}
To assess temporal stability, we also evaluate a one-step transition model in the RANS benchmark. The model is trained to advance the state by one time step using teacher-forced snapshot pairs and is then applied recursively to generate a longer trajectory. This setting tests whether sensitivity supervision reduces error growth when the model repeatedly consumes its own predictions.

\subsection{Computational Cost Accounting}
\label{sec:cost_accounting}

Because sensitivity supervision introduces additional preprocessing and training overhead, we account for computational cost explicitly. Unless otherwise stated, the reported total cost is
\begin{equation}
    \mathcal{C}_{\mathrm{total}}
    =
    \mathcal{C}_{\mathrm{data}}
    +
    \mathcal{C}_{J}
    +
    \mathcal{C}_{\mathrm{train}},
    \label{eq:total_cost}
\end{equation}
where \(\mathcal{C}_{\mathrm{data}}\) is the cost of generating reference solution data, \(\mathcal{C}_{J}\) is the cost of computing or preparing solver-derived Jacobian information, and \(\mathcal{C}_{\mathrm{train}}\) is the neural-operator training cost. For state-only baselines, \(\mathcal{C}_{J}=0\). For sensitivity-constrained models, \(\mathcal{C}_{J}\) is included in the reported wall-clock cost rather than treated as free information.

This accounting is used to interpret the accuracy--cost tradeoff. The cost breakdowns document the measured overhead of data generation, Jacobian preparation, and model training, while the corresponding results sections compare accuracy against total wall-clock cost. Equal-compute or same-wall-clock claims are made only where the results explicitly compare models at matched or interpolated total cost. Detailed computational cost breakdowns are provided in Appendix~\ref{sec:cpu_time}.

\section{Controlled PDE Benchmarks}
\label{sec:controlled_pde_benchmarks}

We first evaluate SC-NO on two controlled benchmarks, PDE1 and PDE2. These cases serve as preliminary testbeds because they involve interpretable high-dimensional inputs, clearly defined forward and inverse tasks, and solver-derived Jacobians that can be used directly for sensitivity supervision. The goal of this section is to examine how Jacobian supervision affects neural-operator behavior in controlled scientific computing settings. We evaluate three aspects: forward prediction of time-dependent PDE states, inverse reconstruction of distributed input fields, and long-horizon autoregressive rollout for the RANS benchmark. Together, these experiments isolate the effect of sensitivity supervision before moving to empirical scaling analysis and the larger SWE tsunami application.

\subsection{Forward Prediction with High-Dimensional Inputs}
\label{sec:controlled_forward_prediction}

We first consider the forward operator-learning problem, where each model maps distributed input fields to the corresponding time-dependent PDE solution.

\begin{figure}
    \centering
    \includegraphics[width=0.98\textwidth]{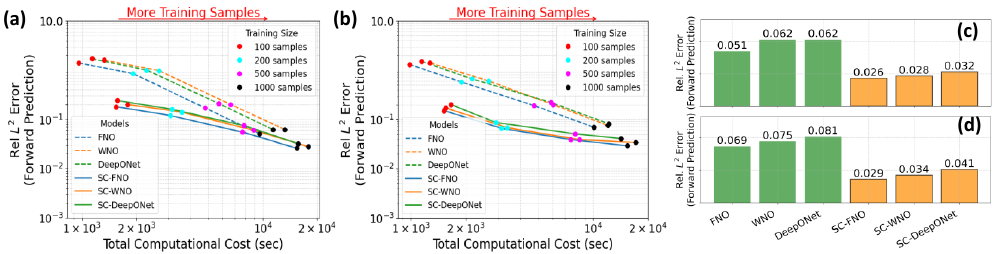}
    \captionsetup{width=0.98\textwidth}
    \caption{\scriptsize\textbf{Forward-prediction performance on the controlled PDE benchmarks.}
    (a,b) Relative \(L^2\) error versus total computation time as the number of training samples increases for PDE1 and PDE2, respectively.
    (c,d) Relative \(L^2\) error at 1000 training samples across FNO, WNO, DeepONet, and their sensitivity-constrained variants for PDE1 and PDE2, respectively.}
    \label{figs:pde1_pde2_forward}
\end{figure}

The forward relative \(L^2\) error is reported for PDE1 and PDE2 in Figure~\ref{figs:pde1_pde2_forward}a,b, respectively, as the training sample size and the corresponding computation time increase. In both PDE settings, the sensitivity-constrained models achieve lower prediction error than their state-only counterparts across the reported sample-size range. The improvement is most pronounced in the low-data regime and decreases as additional solution samples are added, which suggests that Jacobian supervision provides useful local response information when state supervision alone is insufficient.

\begin{figure}
    \centering
    \includegraphics[width=0.98\textwidth, trim={0 0 0 0}, clip]{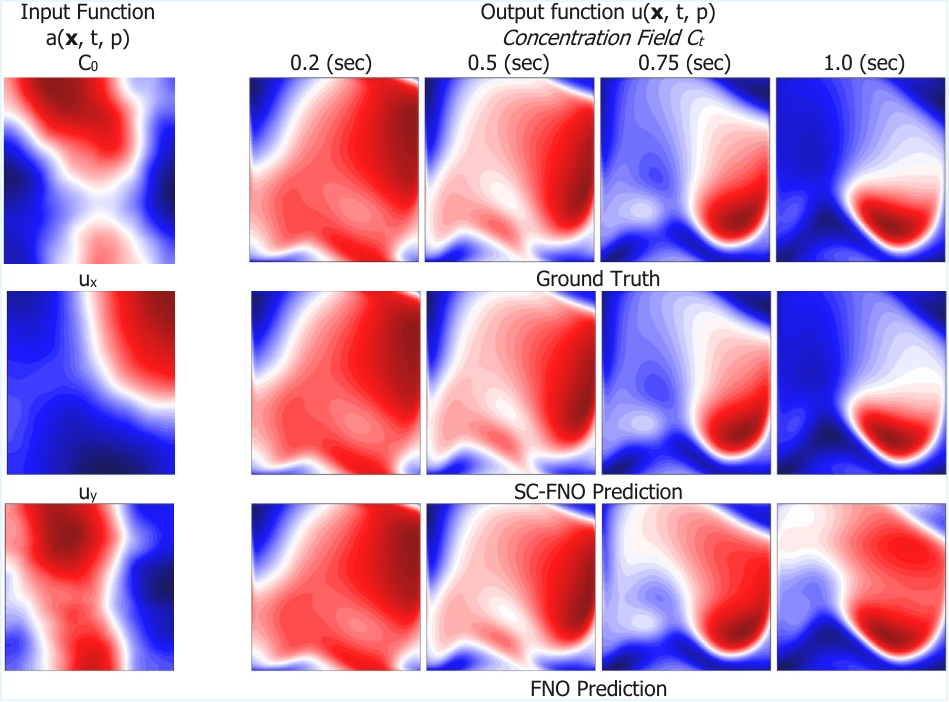}
    \captionsetup{width=0.98\textwidth}
    \caption{\scriptsize\textbf{PDE1 Sample Prediction.}  
    Representative forward prediction for PDE1 showing ground truth, SC-FNO, and FNO, trained on 1000 samples, at selected time snapshots.  
    SC-FNO remains closely aligned with the ground truth and avoids the dissipative errors observed in FNO.}
    \label{figs:pde1_sample_prediction}
\end{figure}

\begin{figure}
    \centering
    \includegraphics[width=0.98\textwidth, trim={0 0 0 0}, clip]{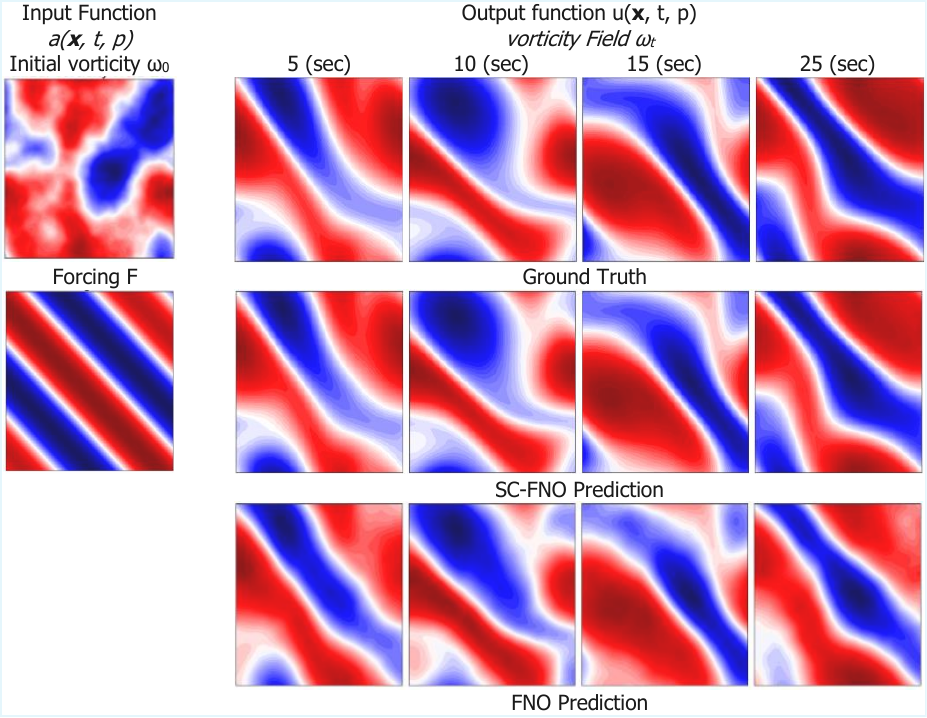}
    \captionsetup{width=0.98\textwidth}
    \caption{\scriptsize\textbf{PDE2 Sample Prediction.}  
    Representative forward predictions for PDE2 comparing the ground truth, SC-FNO, and FNO, trained on 200 samples, at selected time snapshots. SC-FNO maintains close agreement with the ground truth, while FNO exhibits dissipative drift and progressively smeared structures over time.}
    \label{figs:pde2_sample_prediction}
\end{figure}

Figure~\ref{figs:pde1_pde2_forward}c,d show that the benefit of sensitivity supervision is consistent across FNO, WNO, and DeepONet, indicating that the improvement is not limited to a single neural-operator architecture. Among these models, FNO shows the largest performance improvement after adding sensitivity supervision. This may be because FNO provides a strong global spectral representation for these gridded PDE benchmarks, making it particularly responsive to additional derivative information during training. Full numerical results across all models and sample sizes are provided in Appendix Tables~\ref{tab:pde1_forward_nostd} and~\ref{tab:pde2_forward_nostd}. These results show a bounded effect: sensitivity supervision improves forward accuracy and sample efficiency, especially with limited training data, but it does not replace state supervision. As the training set grows, unconstrained neural operators also improve, and the performance gap narrows. Thus, the main conclusion is that Jacobians provide useful additional training information for learning distributed-field solution operators in low-data regimes. Representative forward predictions are shown in Figure~\ref{figs:pde1_sample_prediction} for PDE1 and Figure~\ref{figs:pde2_sample_prediction} for PDE2. In the advection--diffusion case, SC-FNO better preserves the transported concentration structure, while FNO introduces stronger smoothing. In the RANS--Spalart--Allmaras case, SC-FNO better maintains coherent vorticity patterns, whereas FNO exhibits more dissipative drift and spatial smearing over time.

\subsection{Inverse Reconstruction of Distributed Fields}
\label{sec:controlled_inverse_reconstruction}

We next evaluate inverse reconstruction, where a trained neural operator is embedded inside a gradient-based optimization loop to infer an unknown distributed input field from observed solution states. This task is more sensitive to the learned derivative structure of the surrogate than forward prediction. A neural operator may produce acceptable state predictions while still providing inaccurate gradients with respect to the input field; such gradient errors can directly degrade inverse recovery.

\begin{figure}
    \centering
    \includegraphics[width=0.98\textwidth]{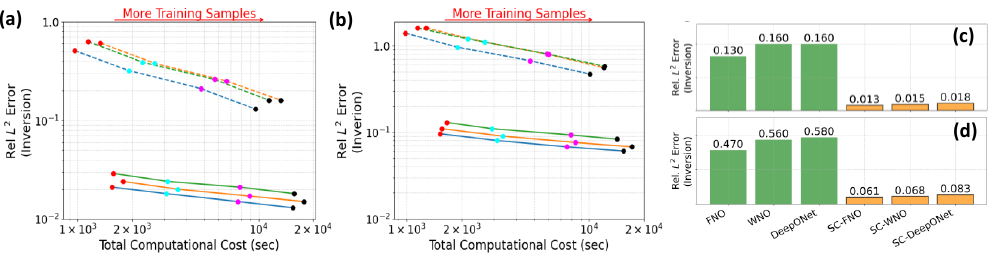}
    \captionsetup{width=0.98\textwidth}
    \caption{\scriptsize\textbf{Inverse-reconstruction performance on the controlled PDE benchmarks.}
    (a,b) Relative \(L^2\) error versus total computation time as the number of training samples increases for PDE1 and PDE2, respectively, where the reconstructed fields are \(C_0\) for PDE1 and \(\omega_0\) for PDE2.
    (c,d) Relative \(L^2\) error at 1000 training samples across FNO, WNO, DeepONet, and their sensitivity-constrained variants for PDE1 and PDE2, respectively.}
        
    \label{figs:pde1_pde2_inverse}
\end{figure}

Figure~\ref{figs:pde1_pde2_inverse}a reports the inverse relative \(L^2\) error for PDE1, where the unknown field is the initial concentration \(C_0\). Figure~\ref{figs:pde1_pde2_inverse}b reports the corresponding inverse error for PDE2, where the unknown field is the initial vorticity \(\omega_0\). In both benchmarks, the sensitivity-constrained models produce lower reconstruction errors than the state-only baselines across the reported training sample sizes and computational-cost range. The separation between constrained and unconstrained models is larger than in the forward task, which is consistent with the fact that inverse reconstruction depends directly on surrogate gradients.

\begin{figure}
    \centering
    \includegraphics[width=0.98\textwidth]{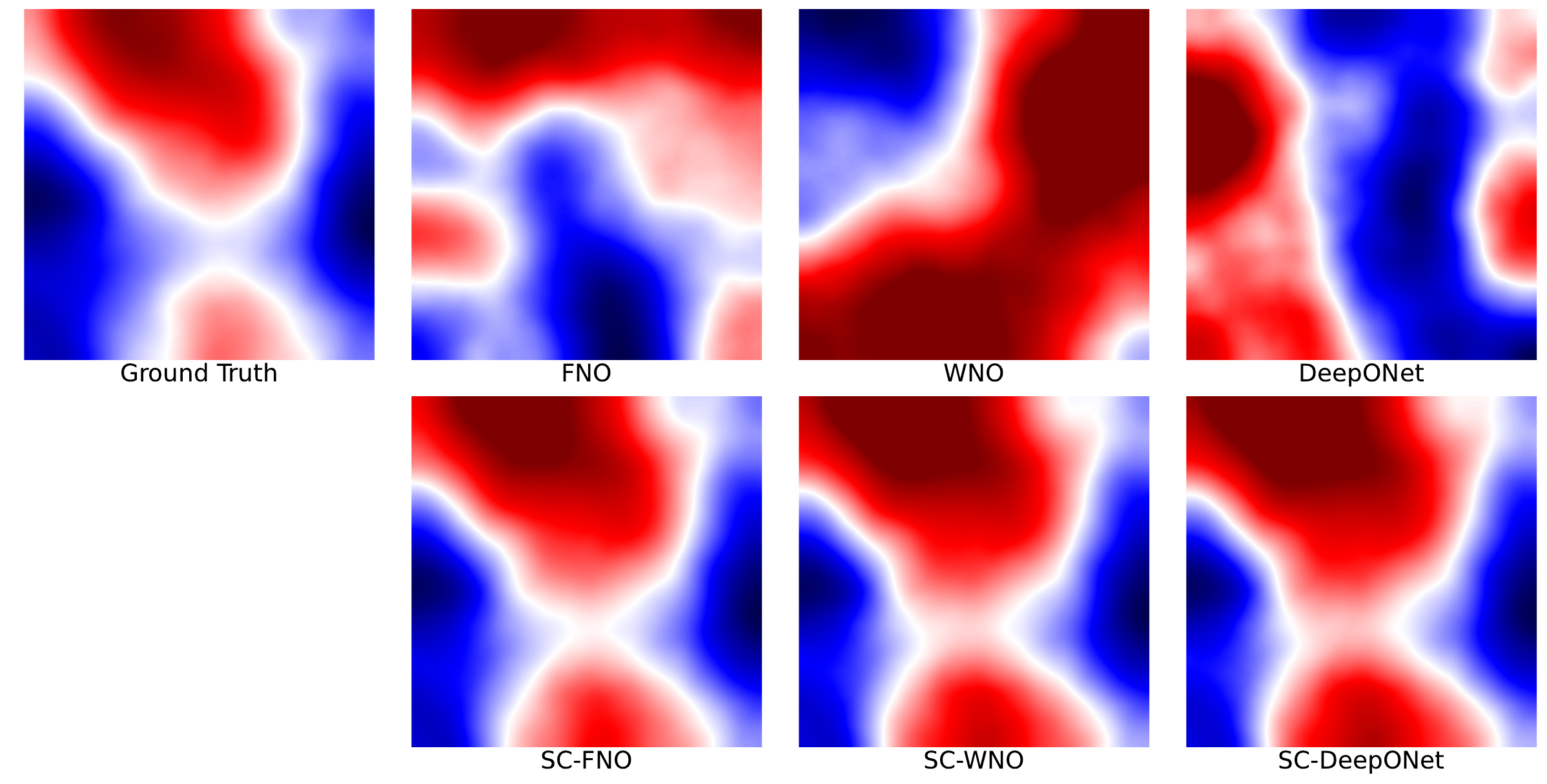}
    \captionsetup{width=0.98\textwidth}
    \caption{\scriptsize\textbf{Inversion of the initial concentration field for PDE1 (all models trained using 200 samples).} 
    Comparison of the true initial concentration field $C_0$ (top-left) with the inferred fields produced by FNO, WNO, DeepONet, and their sensitivity-constrained counterparts (SC-FNO, SC-WNO, SC-DeepONet).}    \label{fig:PDE1_C0_inversion}
    \end{figure}

\begin{figure}
    \centering
    \includegraphics[width=0.95\textwidth]{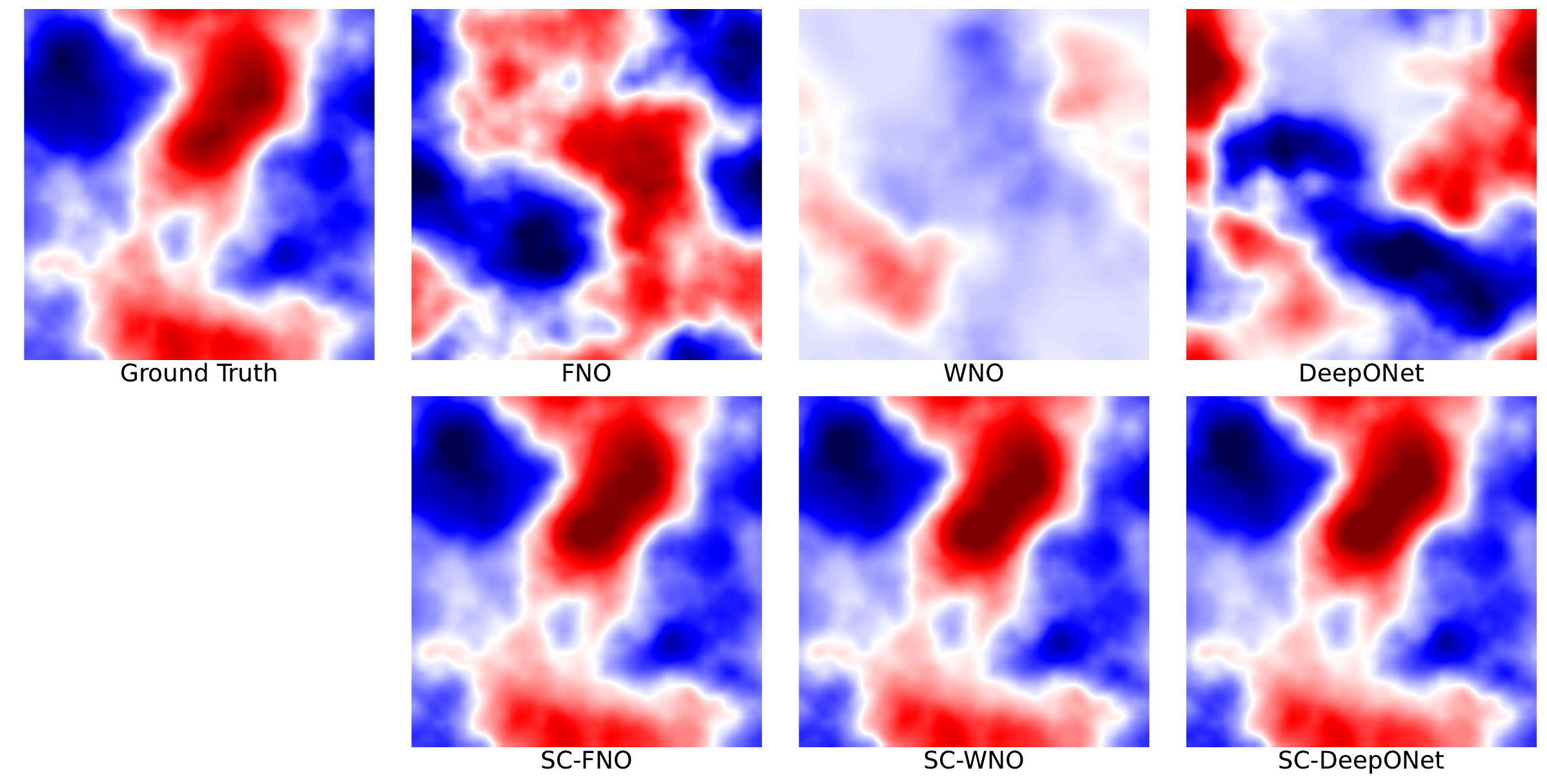}
    \caption{\scriptsize\textbf{Inversion of the Initial Vorticity Field for PDE2 (all models trained using 200 samples).}
    Comparison of the true initial vorticity field $\omega_0$ (top-left) with the inferred fields produced by FNO, WNO, DeepONet, and their sensitivity-constrained counterparts (SC-FNO, SC-WNO, SC-DeepONet). 
    }
    \label{fig:PDE2_w0_inversion}
\end{figure}

Figure~\ref{figs:pde1_pde2_inverse}c,d show that the inverse benefit of sensitivity supervision is consistent across FNO, WNO, and DeepONet. Among these architectures, FNO achieves the strongest inverse performance after adding sensitivity supervision. This result suggests that, for these gridded PDE benchmarks, FNO provides the most effective base architecture for exploiting the supervised sensitivity information. Full numerical results across all models and sample sizes are provided in Appendix Tables~\ref{tab:pde1_inverse_nostd} and~\ref{tab:pde2_inverse_nostd}. The inverse results support a practical conclusion, not a theoretical one. Jacobian supervision does not make the inverse problems well posed; instead, it improves the gradients supplied by the learned surrogate during optimization. For high-dimensional unknown fields such as \(C_0\) and \(\omega_0\), this leads to more accurate and more stable reconstructions in the reported experiments. Sample inverse reconstructions are shown in Figure~\ref{fig:PDE1_C0_inversion} for PDE1 and Figure~\ref{fig:PDE2_w0_inversion} for PDE2.

\subsection{Long-Horizon Autoregressive Rollout}
\label{sec:long_horizon_rollout}

We next evaluate temporal stability using the RANS--Spalart--Allmaras benchmark. Long-horizon autoregressive rollout is a standard stress test for neural PDE surrogates, because models trained under teacher forcing can suffer from compounding error, exposure bias, and closed-loop instability when deployed recursively~\citep{mccabe2023towards,lippe2023pde}. In this experiment, the neural operator is trained as a one-step transition model: given the current state, it predicts the next state. During inference, the same learned transition is applied in closed loop by feeding each predicted state back as the input for the next step, thereby generating the full trajectory over the target horizon. This autoregressive deployment is more demanding than direct multi-step prediction because the model no longer conditions only on ground-truth states; instead, it must remain stable under its own accumulated prediction errors.

\begin{figure}
    \centering
    \includegraphics[width=0.98\textwidth, trim={0 0 0 0}, clip]{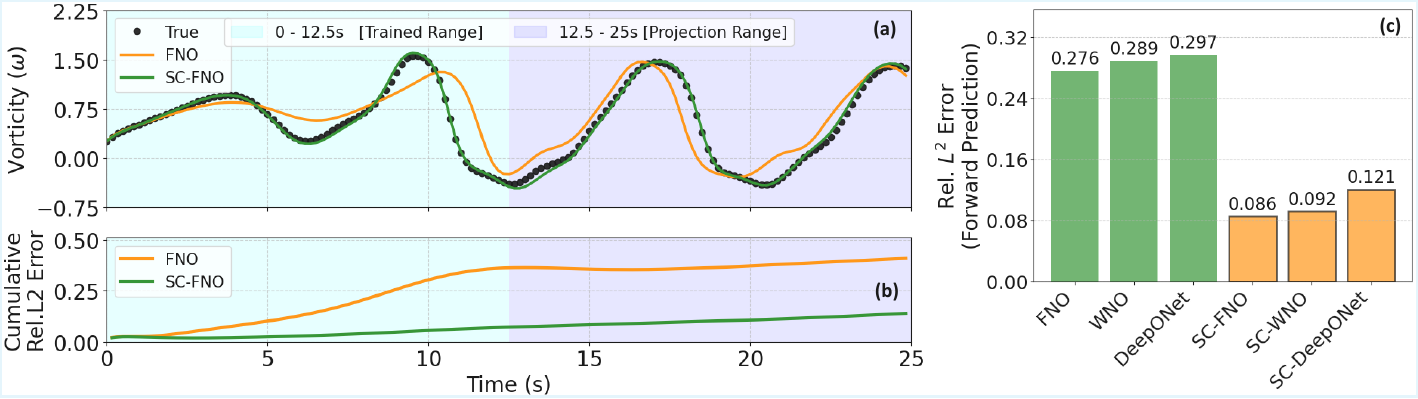}
    \captionsetup{width=0.98\textwidth}
    \caption{\scriptsize\textbf{Long-horizon autoregressive rollout on the RANS benchmark.}
    (a) Representative vorticity time series comparing ground truth, FNO, and SC-FNO during autoregressive rollout.
    (b) Cumulative relative \(L^2\) error over the rollout horizon.
    (c) Rollout error comparison across FNO, WNO, DeepONet, and their sensitivity-constrained variants, all trained with 1000 samples.}
    \label{fig:PDE2_rollout}
\end{figure}

Figure~\ref{fig:PDE2_rollout} shows that sensitivity supervision reduces error growth during autoregressive rollout. The one-step operator is trained only over the first half of the trajectory, \(0\text{--}12.5\,\mathrm{s}\), and is then applied recursively to predict the full horizon up to \(25\,\mathrm{s}\). In the trained range, both models follow the reference trajectory reasonably well, but their behavior separates in the projection range, \(12.5\text{--}25\,\mathrm{s}\). The standard FNO develops larger phase and amplitude errors as the rollout proceeds, whereas SC-FNO remains closer to the reference trajectory over the same horizon. The cumulative error curves in Figure~\ref{fig:PDE2_rollout}b show that neither model eliminates error accumulation, but SC-FNO slows its growth substantially. Figure~\ref{fig:PDE2_rollout}c summarizes the final rollout errors for models trained with 1000 samples. For FNO, sensitivity supervision reduces the final rollout error from \(0.276\) to \(0.086\), corresponding to an approximately \(3.2\times\) reduction. Similar reductions are observed for WNO and DeepONet when their sensitivity-constrained variants are used. These cross-architecture results indicate that the rollout-stability benefit is not specific to the FNO backbone, but is associated with the added sensitivity supervision. Overall, the results show that sensitivity supervision improves not only one-shot prediction accuracy but also the empirical stability of repeated closed-loop surrogate evaluations beyond the training horizon.

The reduced rollout error is consistent with improved local response behavior. In closed-loop prediction, each output becomes the next input, so the model is repeatedly evaluated on its own perturbed states rather than only on ground-truth trajectories. If the learned response to these perturbations is inaccurate, errors can accumulate as phase drift, amplitude bias, or excessive diffusion. Sensitivity supervision can reduce this effect by constraining selected surrogate derivatives in the supervised input directions. 

\section{Empirical Scaling with Input Dimension and Compute}
\label{sec:empirical_scaling}

The controlled benchmarks above show that sensitivity supervision improves both forward prediction and inverse reconstruction for distributed-field PDE inputs. The remaining question is whether this improvement is computationally worthwhile.
In high-dimensional PDE surrogate learning, there are two natural ways to improve model performance: generate more PDE solution samples, or extract more information from each existing sample using solver-derived sensitivities. The first strategy increases the number of training trajectories, whereas the second reflects the central paradigm developed in this work: augmenting each trajectory with solver-derived Jacobian information that encodes dense local input--output response relationships. Both strategies can improve forward prediction and inverse reconstruction, as shown in Figures~\ref{figs:pde1_pde2_forward} and~\ref{figs:pde1_pde2_inverse}; the purpose of this section is to quantify their relative effectiveness and computational efficiency. We therefore evaluate which strategy provides the better accuracy--cost tradeoff in the reported experiments. Specifically, we examine how error changes as the input degrees of freedom, training sample size, total compute budget, and Jacobian supervision density are varied. All costs follow the accounting in Eq.~\eqref{eq:total_cost}, including reference data generation, solver-derived Jacobian preparation, and neural-operator training, as reported in Appendix~\ref {sec:cpu_time}.

\subsection{Scaling with Input Degrees of Freedom}
\label{sec:scaling_input_dof}

We first isolate how the accuracy--cost behavior changes as the number of independent input degrees of freedom increases. Here, input degrees of freedom refer to the number of independent values used to generate the spatially varying input field, i.e., the intrinsic resolution of the initial vorticity field \(\Omega_0(x)\) in the RANS benchmark. To isolate this factor, the RANS solution grid is fixed at \(64\times64\), and only the intrinsic resolution of \(\Omega_0(x)\) is varied. Specifically, \(\Omega_0(x)\) is generated at \(4\times4\), \(16\times16\), and \(64\times64\) resolutions, corresponding to \(4^2\), \(16^2\), and \(64^2\) independent input degrees of freedom, and each realization is then upsampled to the common \(64\times64\) grid before training. Thus, the neural-operator architecture, solution grid, and output representation remain fixed, while the intrinsic dimensionality of the input field changes.

\begin{figure}
    \centering
    \includegraphics[width=0.98\textwidth]{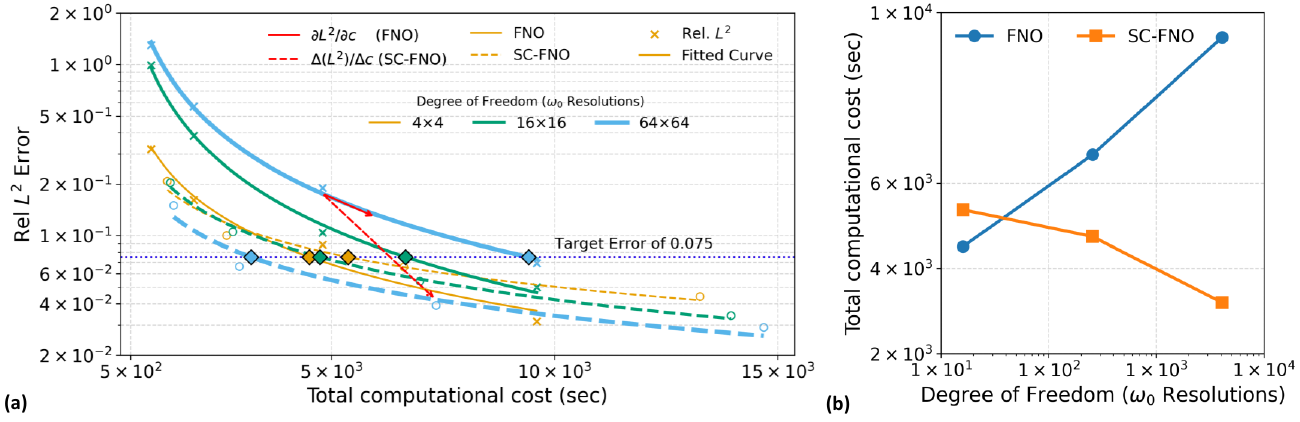}
    \captionsetup{width=0.98\textwidth}
    \caption{\scriptsize\textbf{Empirical accuracy--cost scaling with input degrees of freedom.}
    (a) Relative \(L^2\) error versus total computational cost for FNO and SC-FNO as the intrinsic resolution of \(\Omega_0\) is varied. 
    (b) Estimated total cost required to reach the target relative \(L^2\) error of \(0.075\), obtained from the fitted curves in panel (a). Total cost includes data generation, Jacobian preparation where applicable, and model training.}
    \label{fig:scaling_dof}    
\end{figure}

Figure~\ref{fig:scaling_dof}a reports the relative \(L^2\) error as a function of total computational cost for FNO and SC-FNO. For each DOF level, models are trained with 100, 200, 500, and 1000 trajectories, and each point represents one trained model at one dataset size. Total cost includes the wall-clock time required to generate the corresponding PDE solution data, prepare solver-derived Jacobians for SC-FNO, and train the neural operator. Thus, larger datasets increase the cost for both models, while SC-FNO also includes Jacobian-related overhead. The fitted curves summarize the empirical error--cost trend for each input resolution and directly compare two strategies: reducing error by adding more trajectories, or reducing error by extracting more sensitivity information from each trajectory.

For the standard FNO, adding more training trajectories improves accuracy, but the cost required to maintain a fixed error level increases as the intrinsic resolution of \(\Omega_0\) increases. This indicates that the data-only strategy becomes progressively more expensive as the input field contains more independent degrees of freedom. SC-FNO changes this tradeoff by augmenting each sample with solver-derived Jacobian information. These sensitivities provide local input--output response constraints, allowing the model to receive information about many perturbation directions of the distributed input field from each trajectory. Consequently, the SC-FNO error--cost curves remain lower and more stable across the tested input resolutions.

Figure~\ref{fig:scaling_dof}b summarizes panel~\ref{fig:scaling_dof}a by extracting the estimated total cost required to reach a fixed target error, here a relative \(L^2\) error of \(0.075\). This value is obtained by intersecting the fitted curves with the target-error line and reading the corresponding computational cost. For FNO, the required cost increases with the intrinsic resolution of \(\Omega_0\). For SC-FNO, the required cost remains lower over the tested range, indicating that the added cost of Jacobian supervision is outweighed by the accuracy gained from the additional sensitivity information.

These results support a practical conclusion: in the reported RANS setting, the sensitivity-supervision strategy is more cost-effective than the data-only strategy for maintaining accuracy as input dimensionality increases. This should be interpreted as a finite-range empirical scaling observation, not as a universal complexity law. For the tested resolutions and training budgets, the additional cost of SC-FNO is justified by the reduction in total cost required to reach the same target accuracy.

\subsection{Jacobian Sampling and State-Resolution Ablation}
\label{sec:jacobian_sampling_ablation}

The previous subsection shows that sensitivity supervision can reduce the total cost required to reach a fixed target error. We now test the mechanism behind this improvement by separating two information sources: state-value supervision and Jacobian supervision. The central question is whether sensitivity supervision is merely equivalent to providing denser solution data, or whether it supplies additional input--output information that state values alone do not contain.

\begin{figure}
    \centering
    \includegraphics[width=0.98\textwidth]{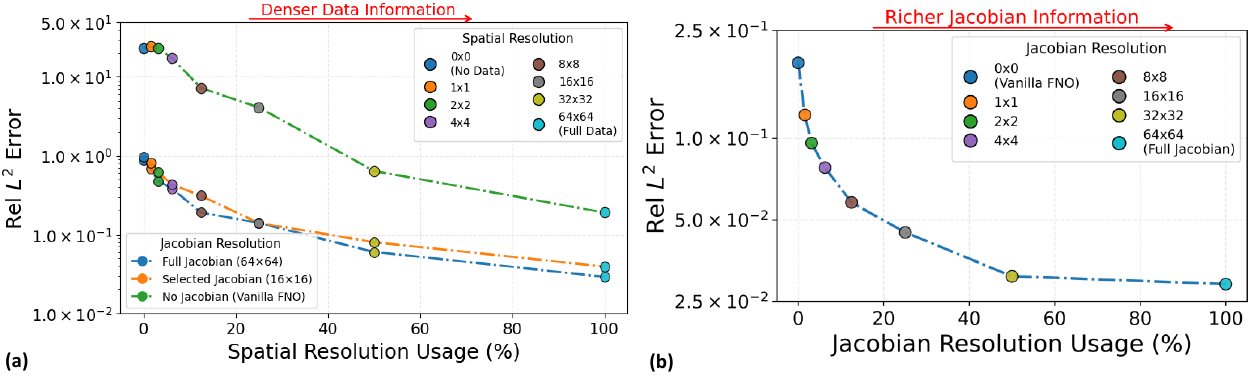}
    \captionsetup{width=0.98\textwidth}
    \caption{\scriptsize\textbf{Effect of Jacobian supervision density and state-resolution density.}
    (a) Relative \(L^2\) error versus spatial state-resolution usage under no, selected \((16\times16)\), and full \((64\times64)\) Jacobian supervision. 
    (b) Relative \(L^2\) error versus Jacobian resolution usage.}
    \label{fig:jacobian_state_ablation}
\end{figure}

We first examine this question in Figure~\ref{fig:jacobian_state_ablation}a. In this experiment, the spatial density of state supervision is varied from no state data to the full \(64\times64\) state field. For each state-resolution level, three training regimes are compared: no Jacobian supervision, selected Jacobian supervision using a \(16\times16\) subset of final-time Jacobian locations, and full final-time Jacobian supervision using the \(64\times64\) Jacobian locations. Thus, this panel is prepared by changing the amount of state information available to the model while separately controlling whether no, partial, or full sensitivity information is included.

Figure~\ref{fig:jacobian_state_ablation}a shows that denser state supervision improves all models, as expected. However, the models trained with Jacobian supervision remain below the no-Jacobian baseline across the tested state-resolution levels. This indicates that solver-derived sensitivities are not simply a substitute for more state observations. State data constrain the predicted solution values, whereas Jacobian supervision constrains how those solution values respond to perturbations in the distributed input field. The two sources of information are therefore complementary.

This also explains why SC-FNO can be worth its additional cost. Preparing and using Jacobians introduces overhead, but the sensitivity loss provides dense response information that would otherwise require many additional state-only trajectories to approximate indirectly. In this sense, the Jacobian acts as a compact physics-informed supervisory signal: it teaches the model local input--output behavior around each trajectory, rather than only the trajectory outcome itself. The improved error at comparable state-resolution levels supports the claim that the added Jacobian cost provides useful information beyond ordinary solution samples.

Figure~\ref{fig:jacobian_state_ablation}b then addresses the implementation question of how much Jacobian supervision is needed. In this experiment, the state-data setting is fixed, and only the number of sampled final-time Jacobian locations included in the sensitivity loss is varied. The \(0\times0\) point corresponds to the vanilla FNO with no sensitivity supervision. The remaining points progressively include denser Jacobian subsets, from \(1\times1\) to the full \(64\times64\) final-time Jacobian.

The error decreases sharply when moving from no Jacobian supervision to sparse or intermediate Jacobian coverage, and then the improvement begins to saturate as the sampled Jacobian density increases. This shows that full Jacobian enforcement is not required to obtain a benefit, although denser Jacobian coverage provides additional improvement. Together, the two panels support the sampling strategy used in this work: sensitivity supervision supplies information that is complementary to state data, its additional cost is justified by the resulting error reduction in the tested setting, and partial Jacobian sampling provides a practical compromise between accuracy and computational overhead.

\section{Near-Real-Time Inverse–Forward Inference for Tsunami Forecasting}
\label{sec:tsunami}

We now evaluate SC-NO in the Tohoku tsunami benchmark introduced in Section~\ref{sec:experimental_design}. This case serves as the capstone application because it combines high-dimensional gridded source inputs, sparse gauge observations, nonlinear wave propagation, spatial extrapolation, and repeated surrogate evaluations during inversion. The goal is not to propose an operational warning system, but to test whether sensitivity-constrained neural operators can support a near-real-time proof of concept for sparse-observation source reconstruction and subsequent tsunami forecasting.

The source input is the earthquake-induced seafloor deformation field \(\Delta z_b(\mathbf{x})\), generated from the Okada dislocation model using Tohoku-based fault-parameter ranges \citep{okada1992internal, grilli2013numerical}. Reference simulations are produced with a finite-volume shallow-water solver on an unstructured triangular mesh with \(308{,}001\) cells and \(154{,}598\) nodes over a \(1000\,\mathrm{km}\times1000\,\mathrm{km}\) domain initialized from ETOPO 2022 bathymetry. Each simulation spans \(3600\) seconds and is stored in \(51\) snapshots. The high-resolution solution fields and adjoint-derived sensitivities are coarsened to a fixed \(100\times100\) grid for neural-operator training and inference.

The neural operator receives \(\Delta z_b(\mathbf{x})\) and the available initial water-stage context, then predicts the subsequent tsunami evolution on the coarse grid. Sensitivity-constrained variants use the same state-prediction loss as their corresponding baseline operators, augmented with a sensitivity loss based on the Jacobian of the final water stage with respect to bed topography. Details of the case setup, solver validation, Okada source generation, mesh-to-grid mapping, and adjoint sensitivity calculation are provided in Appendix~\ref{sec:SWE-Tohoku}.

\subsection{Forward Operator Evaluation Under Known Source Inputs}
\label{sec:tsunami_forward}

We evaluate forward tsunami propagation under known seafloor deformation inputs. This experiment isolates the learned forward operator because the source deformation field is provided directly to the model, without introducing source-inversion error. The operators are trained using different numbers of tsunami scenarios, where each scenario corresponds to an earthquake-induced seafloor deformation field generated from the Okada fault model. Variability is introduced through both the fault parameters and the epicenter coordinates. We consider two forward-evaluation settings. In the in-distribution setting, the trained models are tested on unseen scenarios whose epicenter locations fall within the training epicenter region. In the out-of-distribution (OOD) setting, the models are tested on unseen scenarios whose epicenter locations lie outside the training region. This design separates interpolation over unseen source realizations from spatial extrapolation to source locations not represented during training. Appendix Figure~\ref{fig:nse_sampling_strategies} illustrates the epicenter sampling strategy used to define the in-distribution and spatial OOD tsunami scenarios. We perform this comparison across FNO, WNO, and DeepONet, each evaluated with and without sensitivity supervision.

\begin{figure}
    \centering
    \includegraphics[width=0.98\textwidth]{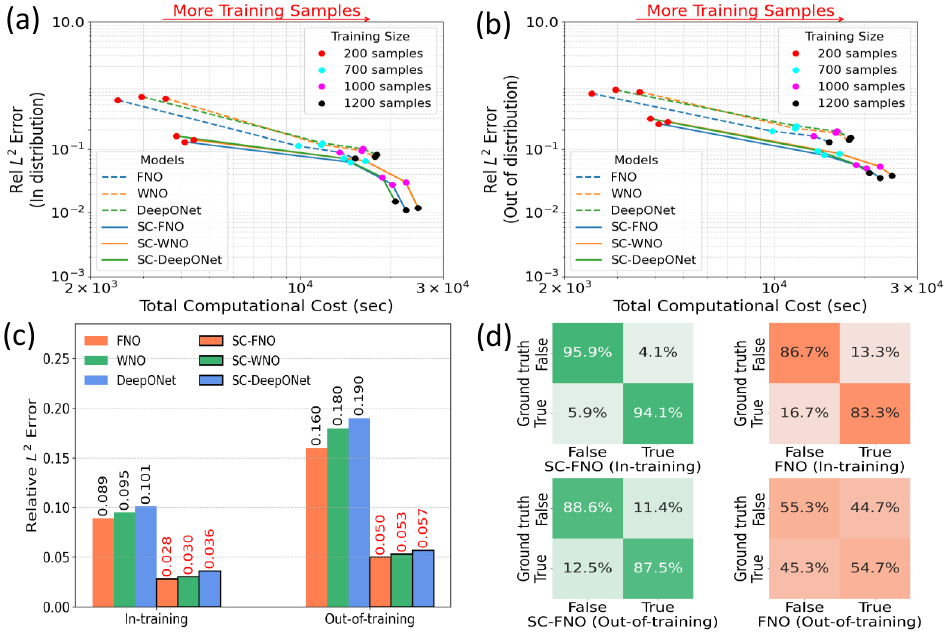}
    \captionsetup{width=0.98\textwidth}
    \caption{\scriptsize\textbf{Forward tsunami prediction across training-set sizes and spatial epicenter shifts.}
    (a) Relative \(L^2\) error versus total computational cost for in-distribution tsunami events, where epicenter locations fall within the training region.
    (b) Relative \(L^2\) error versus total computational cost for spatial out-of-distribution (OOD) events, where epicenter locations are shifted outside the training region.
    (c) Relative \(L^2\) error for FNO, WNO, DeepONet, and their sensitivity-constrained variants at \(1{,}000\) training samples, reported for both in-distribution and OOD events.
    (d) Inundation-classification confusion matrices for SC-FNO and FNO in both settings.}
    \label{fig:tsunami_forward_errors}
\end{figure}

Figures~\ref{fig:tsunami_forward_errors}a and~\ref{fig:tsunami_forward_errors}b report relative \(L^2\) error as a function of total computational cost for in-distribution and OOD tsunami events, respectively. Since larger training sets require greater data generation and training cost, movement to the right in these panels corresponds primarily to increasing the number of training simulations. Consistent with the controlled benchmarks, increasing the training set generally reduces error for all models. However, adding sensitivity supervision shifts the error--cost curves downward in both evaluation settings, showing that SC models achieve lower forward error at comparable computational cost.

Figure~\ref{fig:tsunami_forward_errors}c compares the operator architectures at \(1{,}000\) training samples. Sensitivity-constrained variants reduce error for FNO, WNO, and DeepONet, indicating that the improvement is not limited to one operator backbone. The advantage remains visible under the OOD epicenter split, although all models have higher errors than in the in-distribution case. Figure~\ref{fig:tsunami_forward_errors}d further evaluates inundation-region detection. SC-FNO reduces missed inundation relative to FNO in both settings, including the OOD case where the baseline model produces substantially more false negatives. Thus, sensitivity supervision improves known-source tsunami forecasting in both continuous wave-field prediction and threshold-based inundation capture. The OOD result should be interpreted specifically as improved robustness to the tested epicenter shift, not as unrestricted extrapolation to arbitrary rupture geometries.

\begin{figure}
    \centering
    \includegraphics[width=0.85\textwidth]{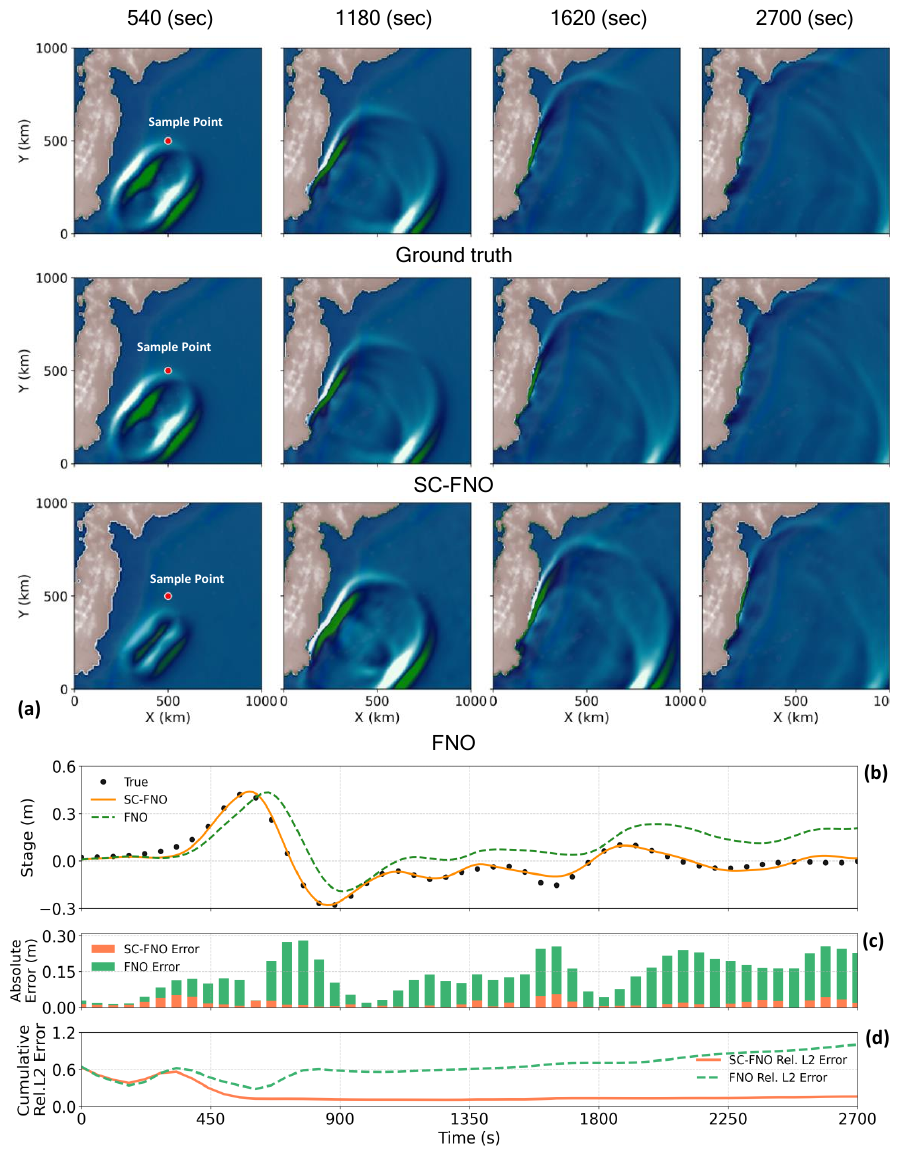}
    \captionsetup{width=0.85\textwidth}
    \caption{\scriptsize\textbf{Forward tsunami propagation under a known gridded seafloor deformation input.}
    (a) Reference shallow-water solution and neural-operator predictions at selected times for one tsunami event. 
    (b) Water-stage time series at the marked sample point. 
    (c) Absolute stage error at the same point. 
    (d) Cumulative relative \(L^2\) error over time.} 
    \label{fig:tsunami_forward_known_source}
\end{figure}

Figure~\ref{fig:tsunami_forward_known_source} provides a representative example of the forward predictions. In Figure~\ref{fig:tsunami_forward_known_source}a, SC-FNO remains closer to the reference solution across the reported time snapshots, whereas the baseline FNO exhibits stronger dissipative drift and loss of wave-field structure. The difference is also visible at the marked gauge location: FNO misrepresents the timing and amplitude of later wave arrivals, while SC-FNO more closely follows the reference stage signal (Figure~\ref{fig:tsunami_forward_known_source}b). The pointwise error comparison in Figure~\ref{fig:tsunami_forward_known_source}c and the cumulative relative \(L^2\) error in Figure~\ref{fig:tsunami_forward_known_source}d show the same trend, with sensitivity supervision reducing forward-propagation error over the forecast window.

\subsection{Gradient-Based Source Inference from Sparse Observations}
\label{sec:tsunami_inversion}

Forward tsunami prediction assumes that the source deformation is already known, but time-critical forecasting requires solving the inverse problem: the source must be inferred from sparse early observations before the full wave field is available. In this setting, the unknown is the earthquake-induced seafloor deformation field, and the observational constraint is the water-stage history recorded at a sparse set of gauge locations over the first \(30\,\mathrm{min}\) of the event. We therefore use the pretrained neural operators as differentiable surrogates inside a gradient-based source-inference framework that reconstructs the deformation field and its associated Okada fault parameters from early-stage gauge observations.

We use a three-stage inversion procedure to avoid relying only on unconstrained pixel-wise optimization of the deformation field. The procedure first performs a flexible field-space inversion, then projects the result into a compact Okada parameterization, and finally refines the Okada parameters through a differentiable decoder and neural-operator forecast model. Full implementation details and stage-wise optimization behavior are provided in Appendix~\ref{sec:inversion_framework}.

In Stage~1, the gridded seafloor deformation field is treated as the optimization variable. Starting from an initial guess, the deformation field is updated by backpropagating the mismatch between predicted and observed gauge histories through the pretrained neural operator. This field-space step provides a flexible source estimate, but the problem is underconstrained by sparse, short-window observations and can produce artifacts because the optimization is performed directly on grid values without an explicit fault-geometry constraint.

In Stage~2, the Stage~1 deformation estimate is mapped into a compact Okada source representation. A pretrained encoder maps the two-dimensional deformation field to a nine-dimensional parameter vector,
\[
    E_{\phi}: \Delta z_b \mapsto \mathbf{P}\in\mathbb{R}^{9},
\]
where \(\mathbf{P}\) contains the epicenter coordinates, depth, length, width, strike, dip, rake, and slip. This step converts the flexible field-space estimate into an interpretable fault-parameter representation.

In Stage~3, the Okada parameters are refined using differentiable feedback from the gauge misfit. A pretrained decoder maps \(\mathbf{P}\) back to a deformation field, which is then propagated through the neural operator to predict gauge time series. The mismatch between predicted and observed gauge histories is backpropagated through the decoder and the neural operator, allowing all nine Okada parameters to be updated by gradient descent. This final stage keeps the reconstruction within the Okada source representation while retaining gradient feedback from the tsunami response.

\begin{figure}
    \centering
    \includegraphics[width=0.98\textwidth, trim={0 0 0 0}, clip]{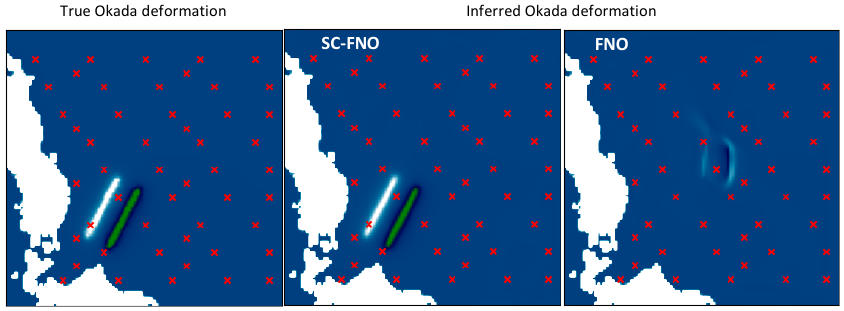}
    \captionsetup{width=0.98\textwidth}
    \caption{\scriptsize\textbf{Three-stage tsunami source inversion from sparse gauge observations.}
    Ground-truth Okada deformation and inferred deformation fields obtained from SC-FNO and FNO after Stage~3 inversion. Red crosses indicate gauge locations used as observational constraints.}
    \label{fig:PDE3_inversion_main_text}
\end{figure}

Figure~\ref{fig:PDE3_inversion_main_text} compares the inferred deformation fields after the three-stage inversion. The SC-FNO-based inversion recovers the main location, polarity, and large-scale geometry of the Okada deformation more accurately than the baseline FNO. The FNO inversion produces a weaker and more spatially diffuse source estimate, indicating that state-only training provides less reliable gradients for this source-inference problem. This behavior is consistent with the controlled inverse benchmarks: source inversion depends not only on forward prediction accuracy, but also on whether the learned surrogate provides useful derivatives with respect to high-dimensional inputs.

This difference is expected from the role of the surrogate inside the inversion loop. In forward prediction, the model is evaluated mainly through its state error, but in source inversion, the optimizer depends on derivatives of the predicted gauge histories with respect to the unknown deformation field. The gauge observations are sparse and restricted to the early window, so many source fields can produce similar short-time responses. In this underconstrained setting, inaccurate surrogate sensitivities can steer gradient descent toward diffuse or weak deformation patterns that partially reduce the gauge loss but do not recover the correct source structure. Sensitivity supervision directly constrains the local response of the learned operator to perturbations in the deformation field, producing gradients that are better aligned with the shallow-water solver. This explains why SC-FNO gives a sharper and more localized Okada reconstruction, whereas the state-only FNO produces a more spatially diffuse estimate. The result is consistent with the controlled inverse benchmarks, where the main advantage of SC-NO appeared in gradient-based reconstruction rather than only in one-shot forward accuracy.

\subsection{Forecast Reconstruction from Inferred Source Fields}
\label{sec:tsunami_forecast_reconstruction}

After the three-stage inversion estimates the Okada deformation field, we propagate the inferred source through the trained neural operator to reconstruct the subsequent tsunami response. This evaluates the coupled inverse--forward workflow: the model is no longer given the true deformation field, but must forecast the event from a source recovered using sparse early gauge observations. The reconstructed gauge histories therefore, test both the quality of the inferred deformation and the stability of the learned forward operator under an estimated, rather than prescribed, source input.

\begin{figure}
    \centering
    \includegraphics[width=0.98\textwidth, trim={0 0 0 0}, clip]{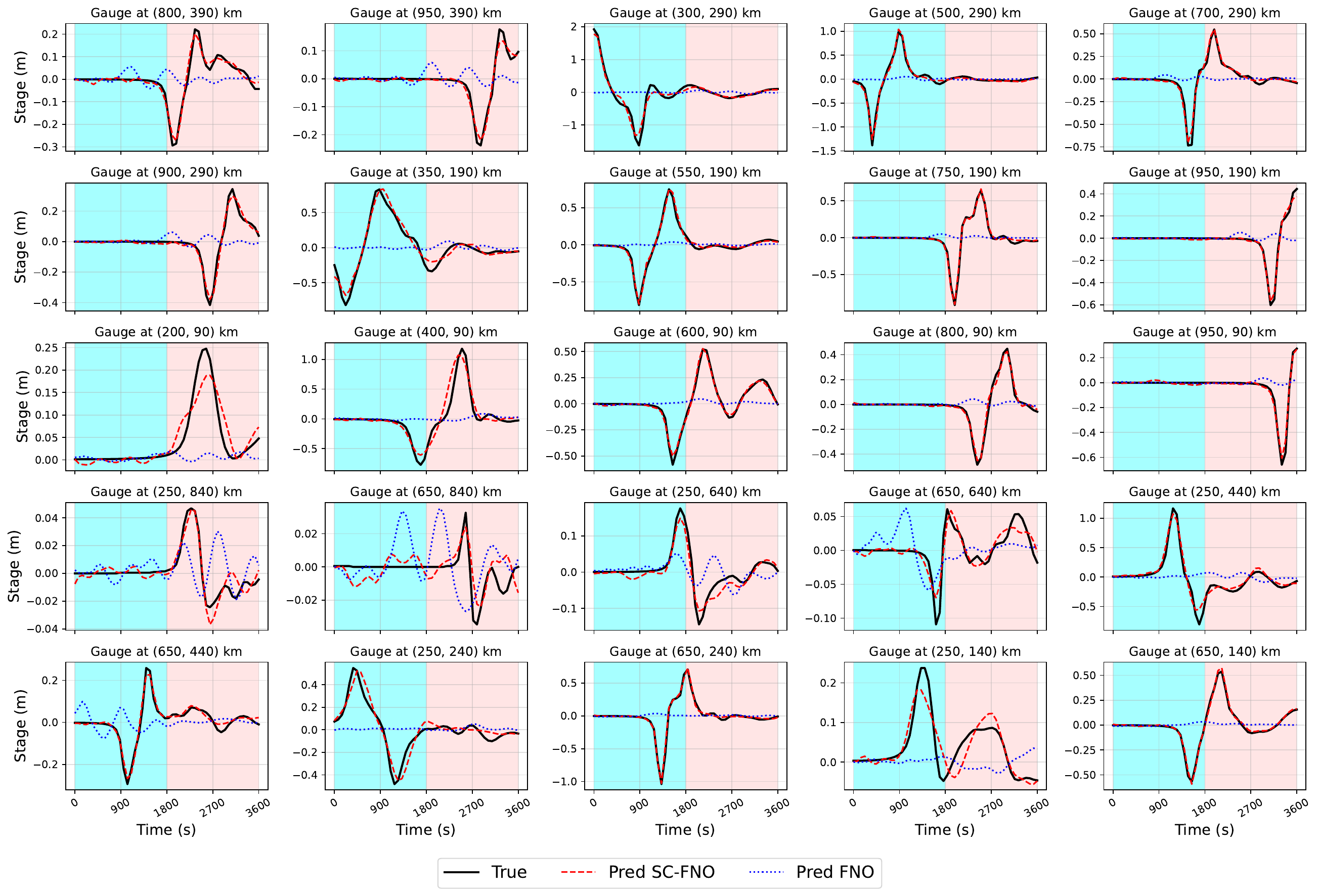}
    \captionsetup{width=0.98\textwidth}
    \caption{\scriptsize\textbf{Reconstructed tsunami gauge time series using the inferred Okada deformation.}
    Comparison between the reference gauge histories (black), SC-FNO predictions (red dashed), and FNO predictions (blue dotted) at multiple gauge locations.
    The aqua-shaded regions denote the early portion of the event (\(T_u = 30\,\mathrm{min}\)) used for inversion, while the light-red regions denote the subsequent forecast window predicted by the models.}
    \label{fig:gauge_timeseries}
\end{figure}

Figure~\ref{fig:gauge_timeseries} compares reconstructed gauge time series obtained from the SC-FNO- and FNO-inferred deformation fields. The aqua-shaded region denotes the first \(T_u=30\,\mathrm{min}\) used for inversion, while the light-red region denotes the subsequent forecast window. Across the reported gauges, SC-FNO more closely reproduces the reference phase and amplitude after the observation window, including negative leading waves and later positive arrivals. The improvement is visible across a range of wave amplitudes, from minor responses below \(0.3\,\mathrm{m}\) to larger responses exceeding \(1\,\mathrm{m}\). In contrast, the FNO-based reconstruction often produces weaker or phase-shifted responses, indicating that its inferred deformation field is less effective as a source for downstream propagation.

This result links the source-inversion behavior in Figure~\ref{fig:PDE3_inversion_main_text} to event-level reconstruction. The advantage of SC-FNO is not only that it recovers a sharper deformation field; the recovered source also produces more accurate gauge histories over the unobserved forecast window. This behavior is consistent with the controlled inverse benchmarks, where sensitivity supervision improved gradient-based recovery of high-dimensional inputs.

The difference between SC-FNO and FNO is expected because forecast reconstruction compounds two surrogate errors: the inverse-gradient error used to estimate the source and the forward-propagation error used to evolve that source. A state-only FNO may produce acceptable predictions when the true deformation is prescribed, but its learned Jacobian with respect to the deformation field is not explicitly constrained. During inversion, this can bias the optimization toward source fields that reduce the early gauge mismatch but do not preserve the correct spatial structure of the tsunami-generating deformation. Once propagated forward, these source errors appear as phase shifts, amplitude damping, or missing wave arrivals in the forecast window. Sensitivity supervision reduces this failure mode by aligning the neural-operator response to deformation perturbations with the adjoint-derived solver sensitivities, producing a source estimate that is both more localized and more dynamically consistent under subsequent propagation.

The full three-stage inversion and forward reconstruction require approximately \(4.5\) minutes on a single NVIDIA A100 GPU, involving thousands of neural-operator surrogate evaluations after offline training and data preparation. This timing reflects the inference-time cost of the proposed workflow, not the cost of generating the training simulations, computing adjoint sensitivities, or training the neural operators. The result, therefore, supports a near-real-time proof of concept for surrogate-assisted source inference and forecast reconstruction in this controlled Tohoku-based benchmark, but should not be interpreted as a complete operational warning system.

Current tsunami forecasting systems commonly rely on precomputed scenario databases and rapid scenario selection or superposition \citep{titov2005real, ishiwatari2012tsunami}. In contrast, the proposed workflow estimates an event-specific deformation field from early gauge observations and propagates the inferred source through a differentiable neural operator surrogate. The neural operators are evaluated on a fixed \(100\times100\) grid obtained by coarsening high-resolution finite-volume simulations and adjoint sensitivities, balancing storage and computational cost against the need to preserve dominant tsunami propagation dynamics. Coarse predictions can then be mapped back to fine-scale fields using the physically aligned upscaling procedure described in Appendix~\ref{sec:SWE-Tohoku}, which preserves consistency with the coarse mean depth while incorporating bed topography (Figure~\ref{fig:upsampling_comparison}).

\subsection{Robustness of Source Inversion to Noisy Observations}
\label{sec:tsunami_noise_robustness}

Sparse gauge observations are rarely noise-free in practical sensing environments. Instrument error, preprocessing uncertainty, timing mismatch, and local unresolved dynamics can all perturb the water-stage history used by the inversion algorithm. Because the source estimate is obtained by differentiating through the neural operator, observational noise can be amplified through the inverse problem and can lead to unstable or biased deformation reconstructions. We therefore evaluate the robustness of the three-stage inversion workflow by adding synthetic noise to the early-window gauge observations used in the inversion objective.

\begin{figure}
    \centering
    \includegraphics[width=0.98\textwidth, trim={0 0 0 0}, clip]{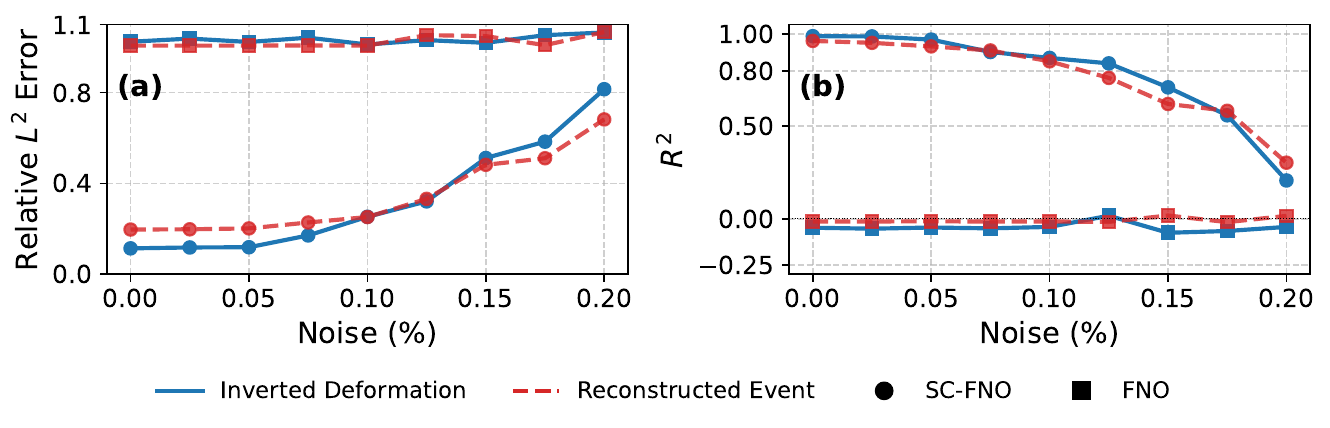}
    \captionsetup{width=0.98\textwidth}
    \caption{\scriptsize\textbf{Robustness of tsunami source inversion and event reconstruction to noisy gauge observations.}
    (a) Relative \(L^2\) error and (b) \(R^2\) score under increasing observation-noise levels.
    Solid lines denote metrics for the inverted source deformation, while dashed lines denote metrics for the reconstructed tsunami event.
    Circles indicate SC-FNO and squares indicate FNO.
    SC-FNO maintains lower source and event reconstruction error over the tested noise range, indicating improved robustness of the gradient-based inversion workflow to imperfect early-window gauge histories.}
    \label{fig:tsunami_noise_robustness}
\end{figure}

Figure~\ref{fig:tsunami_noise_robustness} reports the effect of increasing observation noise on both the inferred source deformation and the reconstructed tsunami event. For the baseline FNO, the source-reconstruction error remains high across all noise levels and the \(R^2\) score stays near zero, indicating that the inversion does not recover a useful deformation field even when the observation noise is small. In contrast, SC-FNO maintains substantially lower relative \(L^2\) error and high \(R^2\) over low-to-moderate noise levels. As the noise level increases, the SC-FNO source estimate also degrades, but it remains more informative than the FNO-based inversion across the tested range.

The same trend appears in the reconstructed-event metrics. Because the forecast is generated by propagating the inferred source through the learned operator, errors in the deformation estimate directly affect the downstream tsunami response. SC-FNO yields lower reconstructed-event error and higher \(R^2\), showing that its advantage is not limited to matching the source field visually. The learned sensitivities provide a more stable optimization landscape: perturbations in gauge observations are less likely to drive the inversion toward diffuse or dynamically inconsistent source fields. This is consistent with the role of sensitivity supervision as a Jacobian-level regularizer, aligning the neural-operator response to source perturbations with adjoint-derived solver sensitivities.

These results should be interpreted as robustness to the tested synthetic observation-noise perturbations, not as a full uncertainty-quantification study. A complete operational setting would also require uncertainty in gauge availability, rupture geometry, bathymetry, boundary conditions, and nearshore physics. Nevertheless, the experiment shows that sensitivity-constrained training improves the conditioning of the sparse-source inversion problem under imperfect gauge histories.

\section{Discussion}
\label{sec:discussion}

\subsection{What Sensitivity Supervision Changes}
\label{sec:what_sensitivity_changes}

The results suggest that sensitivity supervision changes the information available to the neural operator during training, rather than changing the underlying operator architecture. Standard neural operators are trained primarily from state pairs and must infer the local input--output response of the PDE solution operator indirectly from sampled trajectories. In contrast, SC-NO augments each trajectory with solver-derived Jacobian information, so that the learned surrogate is constrained not only to match solution values, but also to match selected directions of the local response of the numerical solver.

This distinction explains the pattern observed across the revised results. In the controlled PDE benchmarks, sensitivity-constrained variants improve forward prediction, especially in lower-data regimes, but the performance gap narrows as more state data are added. This indicates that Jacobian information complements state supervision rather than replacing it. The empirical scaling experiments make the same point from a cost perspective: for the tested RANS settings, adding solver-derived sensitivities provides a more favorable accuracy--cost tradeoff than relying only on additional solution trajectories. The Jacobian-density ablation further shows that the benefit is not simply due to using more state values; state supervision constrains the solution field, while Jacobian supervision constrains how that field changes under perturbations of the distributed input.

The practical effect is therefore best described as an improvement in the observed data--compute tradeoff for the high-dimensional PDE surrogate tasks studied here. The evidence does not establish a universal scaling law or a formal resolution of the curse of dimensionality. Instead, it shows that sampled Jacobian supervision can make each simulated trajectory more informative for learning gridded PDE solution operators, particularly when the downstream task depends on input perturbations, inverse optimization, or repeated surrogate evaluations.

\subsection{Why Sensitivity Supervision Matters More for Inversion}
\label{sec:why_inverse_benefits_more}

The controlled benchmarks show that sensitivity supervision improves both forward prediction and inverse reconstruction, but the gap is larger in the inverse setting. This behavior is expected because the two tasks use the learned surrogate in different ways. Forward prediction primarily evaluates the value of the learned operator,
\[
    \hat{u} = \mathcal{G}_{\theta}(a_{\mathrm{ctx}},p),
\]
where \(a_{\mathrm{ctx}}\) denotes the provided context and \(p\) denotes the supervised input field. In contrast, gradient-based inversion uses the surrogate as a differentiable map with respect to an unknown input variable. For an inverse variable \(q\), the optimization problem has the form
\[
    \min_q
    \left\|
    \mathcal{G}_{\theta}(a_{\mathrm{ctx}},q)
    -
    u_{\mathrm{target}}
    \right\|_2^2 .
\]
The update direction therefore depends on the surrogate derivative \(\partial \mathcal{G}_{\theta}/\partial q\), not only on the state error of \(\mathcal{G}_{\theta}\).

This distinction explains why a state-only neural operator can appear acceptable in forward prediction but perform poorly during inversion. A model may approximate solution fields reasonably well on held-out samples while still learning inaccurate derivatives with respect to the input field. During inverse reconstruction, the optimizer repeatedly evaluates the surrogate at intermediate estimates of \(q\), including fields that may not lie on the training distribution. If the learned derivatives are misaligned with the solver response, the optimizer can reduce the surrogate loss while moving toward input fields that are physically or structurally inconsistent with the true source.

The controlled inverse results support this interpretation. In PDE1 and PDE2, the reconstruction errors for \(C_0\) and \(\omega_0\) show a larger separation between sensitivity-constrained and state-only models than the corresponding forward prediction errors. The same mechanism appears in the tsunami source-inversion case. There, the unknown source is a gridded seafloor deformation field inferred from sparse early gauge histories. The inverse problem is underconstrained because many deformation fields can partially explain the short observation window. In this setting, inaccurate surrogate sensitivities can steer the optimization toward diffuse or weak source estimates that match the early gauges but do not recover the source structure needed for subsequent forecasting.

Sensitivity supervision targets this failure mode by aligning selected surrogate derivatives with solver-derived sensitivities. The resulting improvement should not be interpreted as proving that the inverse problems become well-posed. The evidence supports a narrower conclusion: for the reported high-dimensional field-reconstruction tasks, SC-NO provides more useful surrogate gradients for the tested optimization workflows, leading to more accurate reconstructions than state-only neural operators. This is why the benefit of sensitivity supervision is more pronounced in inverse reconstruction than in direct forward prediction.

\subsection{Scope of Sensitivity-Constrained Operator Learning}
\label{sec:scope_sensitivity_operator_learning}

The present results connect to a broader line of derivative-informed and physics-informed learning methods. Sobolev-type training \citep{czarnecki2017sobolev}, PDE-residual constraints in physics-informed neural operators \citep{li2024physics}, derivative-enhanced DeepONet \citep{qiu2024derivative}, derivative-informed neural operators \citep{cao2025derivative}, and sensitivity-constrained Fourier neural operators \citep{behroozi2025sensitivity} have shown that derivative information can improve generalization when state observations alone are insufficient. The focus of SC-NO is the high-dimensional gridded-input regime, where the supervised input may itself be a spatial field with thousands of degrees of freedom and where the Jacobian must be sampled and amortized to remain computationally practical.

The central distinction is that the sensitivity constraint is imposed at the operator level with respect to distributed input fields. In the benchmarks studied here, the supervised derivative is not a small set of scalar parameter gradients, but a sampled Jacobian relating perturbations in fields such as \(C_0(x)\), \(\omega_0(x)\), \(f(x)\), or \(\Delta z_b(x)\) to the predicted PDE response. This changes the role of derivative information from a local auxiliary regularizer to a scalable source of input--output response supervision. The sampled-Jacobian formulation makes this feasible by enforcing only subsets of the full sensitivity tensor at each update while resampling these constraints across training.

This distinction is also reflected in the experimental design. The controlled PDE benchmarks show that sensitivity supervision improves both forward prediction and inverse reconstruction for distributed-field inputs. The empirical scaling study further separates state-resolution information from Jacobian information and shows that the two are complementary. The tsunami case then evaluates the same mechanism in a sparse-observation inverse--forward workflow, where the surrogate is repeatedly differentiated with respect to an unknown source field. These results position SC-NO as a practical sensitivity-supervision strategy for neural PDE surrogates whose downstream use requires reliable gradients, rather than only accurate one-shot state prediction.

SC-NO is complementary to PDE-residual and physics-informed neural-operator losses. PDE-residual constraints evaluate whether a predicted state satisfies the governing equation, whereas the sensitivity loss constrains how the learned solution operator responds to perturbations of selected inputs. These two forms of physical information act on different aspects of the surrogate. In principle, they can be combined, but the present study isolates the effect of solver-derived Jacobian supervision on high-dimensional forward prediction, inverse reconstruction, rollout stability, and source-inference workflows.

\subsection{Limitations and Practical Requirements}
\label{sec:limitations_practical_requirements}

SC-NO requires access to solver-derived sensitivity information. In this work, these sensitivities are obtained through automatic differentiation for the advection--diffusion benchmark and through discrete adjoints for the RANS and shallow-water benchmarks. This requirement is reasonable in many scientific-computing settings where differentiable solvers or adjoint models are available, but it is still a practical constraint. Applying the method to a new PDE system requires either an existing sensitivity capability or additional effort to derive and implement one.

The cost of sensitivity supervision is also not negligible. Although sampled Jacobian supervision avoids enforcing the full sensitivity tensor at every training step, the method still requires sensitivity generation, storage, or access to sampled Jacobian entries, and model-side differentiation with respect to the supervised inputs. For this reason, the relevant comparison is not state error alone, but accuracy as a function of total computational cost. The empirical results in Sections~\ref{sec:controlled_pde_benchmarks}--\ref{sec:empirical_scaling} indicate that this tradeoff is favorable for the tested benchmarks, particularly in low-data, high-dimensional, and inverse settings. However, the optimal sampling density and the net benefit of sensitivity supervision may depend on the solver, the adjoint cost, the output dimension, and the downstream task.

The tsunami experiment should be viewed as a near-real-time inverse--forward proof of concept rather than a complete operational warning system. The workflow demonstrates that a trained sensitivity-constrained neural operator can support rapid source reconstruction from sparse early gauge observations and subsequent wave forecasting in a controlled Tohoku-based benchmark. Operational deployment would require additional components, including broader treatment of source uncertainty, sensor availability, bathymetric uncertainty, boundary-condition uncertainty, and integration with existing warning-system protocols. These requirements do not weaken the computational result, but they define the boundary between the present surrogate-inference study and a deployed forecasting system.

Finally, the evidence in this paper is empirical over the tested PDE classes, resolutions, and training budgets. The results show that sampled Jacobian supervision improves the observed data--compute tradeoff and inverse behavior in these settings, but they should not be interpreted as a universal complexity result. Establishing when sensitivity supervision remains beneficial across broader PDE classes, longer time horizons, and different adjoint implementations remains an important direction for future work.

\subsection{Implications for Time-Critical PDE Inference}
\label{sec:implications_time_critical}

The broader implication of these results is that derivative-aligned neural operators can be useful when a PDE surrogate is not only queried once, but repeatedly used inside an inference or forecasting loop. This occurs in source inversion, data assimilation, uncertainty exploration, design optimization, and long-horizon prediction. In these settings, state accuracy alone is often insufficient because the surrogate is evaluated under perturbations, intermediate optimization states, or autoregressive feedback.

The tsunami case illustrates this point in a time-critical geophysical setting. The full inverse--forward workflow uses a trained neural operator to estimate a source deformation from early gauge observations and then forecast the later wave response. The reported runtime supports the feasibility of minute-scale surrogate-assisted inference after offline training and sensitivity generation. This should be interpreted as a pathway toward time-critical PDE inference rather than as a complete operational system.

More generally, the results suggest that sensitivity supervision may be valuable in applications where high-fidelity solvers are too expensive for repeated online use, but where differentiable or adjoint information can be generated offline. Potential examples include flood inundation, storm surge, plume transport, subsurface flow, volcanic hazard modeling, and other PDE-governed systems requiring rapid inverse or uncertainty-aware prediction. The most defensible conclusion is that sampled Jacobian supervision improves the reliability of neural PDE surrogates in the tested high-dimensional settings, especially when the surrogate is used for gradient-based inversion or repeated closed-loop evaluation.

\section{Conclusion}
\label{sec:conclusion}

This work studied sensitivity-constrained neural operators for high-dimensional forward and inverse PDE inference. The proposed framework augments standard neural-operator training with sampled solver-derived Jacobian supervision, allowing the learned surrogate to match both solution values and selected input--output sensitivities. The method does not introduce a new neural-operator architecture; instead, it provides a training strategy for improving the reliability of existing differentiable operator models when the input is a distributed spatial field.

Across the controlled advection--diffusion and RANS benchmarks, sensitivity supervision improved forward prediction and produced larger gains in gradient-based inverse reconstruction. The empirical scaling experiments further showed that, in the tested high-dimensional gridded-input setting, Jacobian supervision improved the observed accuracy--cost tradeoff relative to relying only on additional state trajectories. The rollout experiment and the tsunami source-inference workflow also indicate that sensitivity supervision is especially useful when the surrogate is used repeatedly, either in closed-loop prediction or inside an inverse optimization loop.

The tsunami benchmark demonstrates a near-real-time inverse--forward proof of concept: a trained sensitivity-constrained neural operator can reconstruct a gridded seafloor deformation source from sparse early gauge observations and forecast the subsequent wave response within a minute-scale inference workflow. This result should not be interpreted as a complete operational warning system, since operational deployment would require broader uncertainty treatment, sensor robustness, and validation across additional events. Nevertheless, the results suggest that sampled Jacobian supervision is a practical mechanism for improving neural PDE surrogates in high-dimensional settings where forward accuracy, inverse stability, robustness, and computational cost must be considered together.

\clearpage

\section*{Supplementary Materials}
\addcontentsline{toc}{section}{Supplementary Materials}

\section*{Appendices}

The appendices provide supporting theoretical, methodological, numerical, and experimental details for the main manuscript. They are organized as follows:

\begin{enumerate}[label=\Alph*.]
    \item \textbf{Theoretical justification for sensitivity-constrained training}
    \begin{itemize}
        \item Neural operators
        \item Sensitivity-constrained neural operators (SC-NO)
        \item Assumptions, notation, cost functionals, improved generalization, long-term prediction stability, and inversion accuracy
    \end{itemize}

    \item \textbf{Data generation}
    \begin{itemize}
        \item Input-function structure
        \item Dataset structure with true sensitivities
        \item Scaled Gaussian random fields for parameter generation
    \end{itemize}

    \item \textbf{PDE1: Advection--diffusion equation with spatially distributed velocity field}
    \begin{itemize}
        \item Problem setup
        \item Numerical solver
        \item Neural-operator learning setup
        \item Additional results
    \end{itemize}

    \item \textbf{PDE2: Turbulent Navier--Stokes equations with Spalart--Allmaras closure}
    \begin{itemize}
        \item Problem setup
        \item Numerical solver
        \item Neural-operator learning setup
        \item Adjoint-based sensitivities
        \item Additional results
    \end{itemize}

    \item \textbf{PDE3: 2D shallow water equations for the 2011 Tohoku tsunami}
    \begin{itemize}
        \item Problem setup
        \item Okada deformation model
        \item Finite-volume solver
        \item Adjoint-based sensitivities
        \item Neural-operator learning
        \item Fine-to-coarse grid mapping
        \item Three-stage source inversion
        \item Additional results
    \end{itemize}

    \item \textbf{Performance metrics}

    \item \textbf{Hyperparameters and settings}

    \item \textbf{Computational cost analysis}

\end{enumerate}

\clearpage
\appendix

\section{Supporting Analysis: Mechanisms of Sensitivity-Constrained Training}
\label{sec_sup_method}

\renewcommand{\thefigure}{A.\arabic{figure}}  
\renewcommand{\thetable}{A.\arabic{table}}  
\setcounter{figure}{0}
\setcounter{table}{0}

Neural Operators (NOs) provide a framework for learning mappings between infinite-dimensional function spaces, enabling resolution-invariant PDE approximations \citep{lu2021learning, kovachki2023neural}. Unlike traditional solvers that discretize PDEs explicitly, NOs approximate a solution operator \( \mathcal{G}: \mathcal{U} \to \mathcal{V} \), where \( \mathcal{U} \) represents the space of input functions (e.g., initial conditions, boundary conditions, parameters), and \( \mathcal{V} \) denotes the space of solutions. Given an input function \( a(x, t, p) \in \mathcal{U} \), which encapsulates problem-specific information, and an output solution \( u(x, t, p) \in \mathcal{V} \), the operator mapping is:

\begin{equation}
u(x, t, p) = \mathcal{G}(a)(x, t, p),
\end{equation}

where \( x \in \mathcal{X} \) are spatial coordinates, \( t \in [0, T] \) represents time, and \( p \in \mathbb{R}^m \) denotes system parameters. Neural Operators parameterize \( \mathcal{G} \) with learnable weights \( \theta \), forming \( \mathcal{G}_\theta \):

\begin{equation}
u(x, t, p) = \mathcal{G}_\theta(a)(x, t, p).
\end{equation}

Unlike conventional neural networks, NOs operate on function spaces, allowing solutions to be transferred across different resolutions.

\subsection*{Neural Operators}
The architecture of a Neural Operator consists of iterative layers that update the latent representation \( v_t(x, t, p) \) via:

\begin{equation}
v_{t+1}(x, t, p) = \sigma \left( \mathcal{K}_\theta(v_t)(x, t, p) + \mathcal{L}_\theta(v_t)(x, t, p) \right),
\end{equation}

where \( \mathcal{K}_\theta \) is a nonlocal integral operator, \( \mathcal{L}_\theta \) is a local transformation, and \( \sigma \) is a nonlinear activation function. The nonlocal operator is defined as:

\begin{equation}
\mathcal{K}_\theta(v_t)(x, t, p) = \int_{\mathcal{X}} \int_0^T \int_{\mathbb{R}^m} K_\theta(x, t, p; y, s, q) v_t(y, s, q) \, dy \, ds \, dq,
\end{equation}

where \( K_\theta \) is a learnable kernel encoding interactions across space, time, and parameters. The local transformation follows:

\begin{equation}
\mathcal{L}_\theta(v_t)(x, t, p) = W_\theta v_t(x, t, p) + b_\theta(x, t, p).
\end{equation}

Different parameterization strategies yield distinct NO architectures:

- \textbf{Fourier Neural Operators (FNOs)} employ spectral transforms for efficiency, defining the integral operator in Fourier space:

\begin{equation}
\mathcal{F}(\mathcal{K}_\theta(v_t))(k, \omega) = R_\theta(k, \omega) \cdot \mathcal{F}(v_t)(k, \omega),
\end{equation}

where \( \mathcal{F} \) denotes the Fourier transform, and \( R_\theta(k, \omega) \) is a learnable kernel truncated for high frequencies:

\begin{equation}
R_\theta(k, \omega) = 0 \quad \text{for} \quad |k|, |\omega| > k_{\text{max}}.
\end{equation}

- \textbf{Wavelet Neural Operators (WNOs)} replace Fourier transforms with wavelet decompositions to capture localized structures across scales \citep{tripura2023wavelet}.

- \textbf{DeepONets} decompose the mapping into a trunk network (spatial-temporal encoding) and a branch network (parametric encoding) \citep{lu2021learning}:

\begin{equation}
\mathcal{K}_\theta(v_t)(x, t, p) = \sum_{i=1}^m b_i(p) \tau_i(x, t),
\end{equation}

where \( b_i(p) \) and \( \tau_i(x, t) \) are learned functions.


After \( T \) iterations, the final output is projected onto the solution space:

\begin{equation}
u(x, t, p) = Q_\theta(v_T)(x, t, p).
\end{equation}

\subsection*{Sensitivity-Constrained Neural Operators (SC-NO)}
Standard NOs minimize a solution-based loss:

\begin{equation}
L_u = \ell_u(\hat{u}(x, t; p), u(x, t; p)),
\end{equation}

where \( \ell_u \) measures prediction error. However, they do not explicitly account for how solutions change with respect to parameters \( p \), which is crucial for generalization and inverse problems. To address this, we introduce a sensitivity regularization term:

\begin{equation}
L_S = \ell_s\left( \frac{\partial \hat{u}(x, t; p)}{\partial p}, \frac{\partial u(x, t; p)}{\partial p} \right),
\end{equation}

where \( \frac{\partial \hat{u}}{\partial p} \) is computed via automatic differentiation, and \( \frac{\partial u}{\partial p} \) is derived from a high-fidelity solver. The final training objective balances both terms:

\begin{equation}
L = L_u + \lambda L_S,
\end{equation}

where \( \lambda \) controls sensitivity enforcement.

\subsection*{Theoretical Justification} \label{prove}

We provide a mathematical basis for why Sensitivity-Constrained Neural Operators (SC-NOs) outperform standard training by incorporating sensitivity loss, improving generalization, long-term stability, and inversion accuracy. We define the operator $\mathcal{G}_\theta: \mathcal{U} \to \mathcal{V}$, which maps an input function $a(x) = (u_0(x), p(x))$ to $\hat{u}(x,t) = \mathcal{G}_\theta(a(x))$, approximating the true solution $u(x,t)$.

\subsubsection*{Assumptions and Notation}

We work in bounded domains: the parameter space $\mathcal{P}_{\text{param}} \subset \mathbb{R}^m$ and the spatio-temporal domain $\mathcal{X} \times [0,T]$. We assume $\hat{u}$ is differentiable with respect to $p$. We use the Euclidean norm $|\cdot|$ on $\mathbb{R}^m$ and the corresponding induced operator norms. We assume the true solution $u$ and its sensitivity $\frac{\partial u}{\partial p}$ are bounded by $M$ and $M_p$ respectively, for all $(x,t,p)$.
\begin{equation}
|u(x,t;p)| \leq M, \quad \left\| \frac{\partial u}{\partial p}(x,t;p) \right\| \leq M_p.
\end{equation}

Our core argument relies on standard (though strong) assumptions from learning theory: for an overparameterized, well-trained model $\theta_s^*$, a small empirical loss implies a small uniform error bound.

\paragraph{Assumption 1 (Uniform Convergence of Gradients):}
We assume that a model $\theta_s^*$ trained to $L_s(\theta_s^*)$ on $n$ samples satisfies:
\begin{equation}
\sup_{x,t,p} \left\| \frac{\partial \hat{u}}{\partial p}(x,t;p;\theta_s^*) - \frac{\partial u}{\partial p}(x,t;p) \right\| \le \varepsilon_s
\end{equation}
where $\varepsilon_s \to 0$ as $L_s \to 0$ and $n \to \infty$.

\paragraph{Assumption 2 (Uniform Convergence of Solutions):}
We assume a similar uniform bound $\varepsilon_u$ for the solution itself:
\begin{equation}
\sup_{x,t,p} |\hat{u}(x,t;p;\theta_s^*) - u(x,t;p)| \le \varepsilon_u
\end{equation}
where $\varepsilon_u \to 0$ as $L_u \to 0$ and $n \to \infty$.

\subsubsection*{Cost Functionals}

Training minimizes the following losses, where $\frac{\partial u}{\partial p}$ is computed on-the-fly via a differentiable solver or adjoint model:
\begin{equation}
L_u(\theta) = \frac{1}{n} \sum_{i=1}^n [\hat{u}(x_i, t_i; p_i;\theta) - u(x_i, t_i; p_i)]^2,
\end{equation}
\begin{equation}
L_s(\theta) = \frac{1}{n} \sum_{i=1}^n \left\| \frac{\partial \hat{u}}{\partial p}(x_i, t_i; p_i;\theta) - \frac{\partial u}{\partial p}(x_i, t_i; p_i) \right\|^2,
\end{equation}
\begin{equation}
L(\theta) = L_u(\theta) + \lambda L_s(\theta), \quad \lambda > 0.
\end{equation}

\subsubsection*{Improved Generalization}

Training with $L(\theta)$ acts as a powerful regularizer by explicitly constraining the model's Lipschitz constant with respect to parameters. We define this constant as:
\begin{equation}
L_{\text{lip}}(\theta) = \sup_{x,t,p} \left\| \frac{\partial \hat{u}}{\partial p}(x,t;p;\theta) \right\|.
\end{equation}
For the sensitivity-constrained model $\theta_s^*$, we can rigorously bound its Lipschitz constant using Assumption 1 and the triangle inequality:
\begin{align}
L_{\text{lip}}(\theta_s^*) &= \sup \left\| \frac{\partial u}{\partial p} + \left( \frac{\partial \hat{u}}{\partial p} - \frac{\partial u}{\partial p} \right) \right\| \notag \\
&\le \sup \left\| \frac{\partial u}{\partial p} \right\| + \sup \left\| \frac{\partial \hat{u}}{\partial p} - \frac{\partial u}{\partial p} \right\| \notag \\
&\le M_p + \varepsilon_s.
\end{align}
By driving $\varepsilon_s \to 0$, we force the model to inherit the (bounded) Lipschitz constant of the true physical system. This has a direct impact on generalization. For fixed $(x,t)$, we define the hypothesis class $\mathcal{F}_{\text{SC}} = \{p \mapsto \hat{u}(x,t;p;\theta): L_{\text{lip}}(\theta) \le M_p+\varepsilon_s\}$. On a bounded parameter domain, standard results imply that the empirical Rademacher complexity $\hat{R}_n(\mathcal{F}_{\text{SC}})$ of this class scales linearly with its Lipschitz constant, $\hat{R}_n(\mathcal{F}_{\text{SC}}) \lesssim (M_p + \varepsilon_s)$. This complexity is strictly lower than that of an unconstrained class $\mathcal{F}_u$ that admits arbitrarily large Lipschitz constants. The standard generalization bound,
\begin{equation}
\mathbb{E}[(\hat{u} - u)^2] \leq L_u(\theta) + 2 \hat{R}_n(\mathcal{F}) + \text{ComplexityPenalty}(\delta, n),
\end{equation}
is therefore tighter for $\mathcal{F}_{\text{SC}}$, implying better performance on unseen parameters.

\subsubsection*{Long-Term Prediction Stability}

This argument is most clearly made by setting the parameter $p$ to be the initial condition $u_0$. The sensitivity $\frac{\partial u(t)}{\partial u_0}$ is the \textbf{fundamental solution operator}, and its norm governs the growth of perturbations (i.e., the Lyapunov exponents).

Assume the true physical system's growth is bounded:
\begin{equation}
\left\| \frac{\partial u(t)}{\partial u_0} \right\| \le C e^{\alpha t}
\end{equation}
for some physical exponent $\alpha$ (which could be positive for a chaotic system, or negative for a stable one). A standard model $\theta_u^*$, trained only on $L_u$, may learn an incorrect operator $\frac{\partial \hat{u}}{\partial u_0}$ with a much larger exponent $\hat{\alpha} \gg \alpha$, leading to spurious amplification of small errors. The SC-NO, by minimizing $L_s$ (with $p=u_0$), uses Assumption 1 to force its learned operator to be close to the true one:
\begin{align}
\left\| \frac{\partial \hat{u}(t)}{\partial u_0} \right\| &\le \left\| \frac{\partial u(t)}{\partial u_0} \right\| + \sup \left\| \frac{\partial \hat{u}}{\partial u_0} - \frac{\partial u}{\partial u_0} \right\| \notag \\
&\le C e^{\alpha t} + \varepsilon_s.
\end{align}
This shows that the SC-NO reproduces the physical growth rates of perturbations up to $O(\varepsilon_s)$ and does not introduce spurious instabilities beyond those present in the governing system.

\subsubsection*{Inversion Accuracy}

In parameter inversion, we seek to find $p^*$ by minimizing a loss $\ell$ between our model $\hat{u}$ and an observation $u_{\text{obs}}$. Let $J(p) = \ell(u(p), u_{\text{obs}})$ be the "true" (but intractable) loss landscape, and $J_\theta(p) = \ell(\hat{u}_\theta(p), u_{\text{obs}})$ be the surrogate landscape our optimizer actually sees.

Gradient descent updates use the gradient of $J_\theta$:
\begin{equation}
p_{k+1} = p_k - \eta \nabla_p J_\theta(p_k), \quad \text{where} \quad \nabla_p J_\theta = \frac{\partial \ell}{\partial \hat{u}} \cdot \frac{\partial \hat{u}}{\partial p}.
\end{equation}
The success of the inversion depends on the \textbf{gradient error} $\|\nabla J_\theta(p) - \nabla J(p)\|$. If $\ell$ is smooth, this error is bounded by our assumed uniform errors:
\begin{equation}
\|\nabla J_\theta(p) - \nabla J(p)\| \le C_1 \varepsilon_u + C_2 \varepsilon_s
\end{equation}
for some constants $C_1, C_2$.

A standard model $\theta_u^*$ is only trained to minimize $\varepsilon_u$, leaving a large, uncontrolled gradient error $C_2 \varepsilon_s$. The optimizer is fed an inaccurate gradient. An SC-NO is explicitly trained to drive both $\varepsilon_u \to 0$ and $\varepsilon_s \to 0$. We further assume that, in a neighborhood of $p^*$, the true objective $J(p)$ is $\mu$-strongly convex. Under this condition, standard perturbation results for gradient descent imply that optimization on $J_\theta$ converges to an $O(\varepsilon_u + \varepsilon_s)$ neighborhood of the true optimum $p^*$. By minimizing $\varepsilon_s$, SC-NOs guarantee a high-fidelity gradient, ensuring a much more accurate and stable convergence.

\clearpage
\section{Data Generation}
\label{app:Data_Generation}
\renewcommand{\thefigure}{B.\arabic{figure}}  
\renewcommand{\thetable}{B.\arabic{table}}  
\setcounter{figure}{0}
\setcounter{table}{0}

For a PDE defined over a spatial domain \( \mathcal{X} = \mathbb{R}^2 \) with \( \mathbf{x} = (x_1, x_2) \in \mathcal{X} \) and time \( t \in [0, T] \), including \( n_t \) discrete time steps of the solution over the interval \( [0, T] \), we generate training data for the Sensitivity-Constrained Neural Operator (SC-NO) to learn the mapping \( \mathcal{G}_\theta: \mathcal{U} \to \mathcal{V} \), where \(\mathcal{U}\) is the space of input functions and \(\mathcal{V}\) is the space of solutions, as described in the Methods section. Each data sample comprises an input function \( a(\textbf{x}, t, p) \in \mathcal{U} \), the corresponding solution \( u(\textbf{x}, t, p) = \mathcal{G}(a)(\textbf{x}, t, p) \in \mathcal{V} \), and the true Jacobian \( \frac{\partial u}{\partial p} \), computed from a high-fidelity solver, to support sensitivity-aware training.

\subsubsection*{Structure of the Input Function}

The input function is defined as:
\begin{equation}
a(\textbf{x}, t, p) = \big( a_{\text{init}}(\textbf{x}, t), p(\textbf{x}) \big),
\end{equation}
where:
\begin{itemize}
    \item \( a_{\text{init}}(\textbf{x}, t) \) contains the initial time steps of the solution:
    \begin{equation}
    a_{\text{init}}(\textbf{x}, t) = \big( u(\textbf{x}, t_1, p), u(\textbf{x}, t_2, p), \dots, u(\textbf{x}, t_{t_u}, p) \big), \quad t \in [0, t_u],
    \end{equation}
    representing \( t_u \) steps of the state \( u(\textbf{x}, t, p) \).

    \item \( p(\textbf{x}) \) is the spatially varying parameter field.

\end{itemize}
The total simulation time is:
\begin{equation}
n_t = t_u + t_G,
\end{equation}
where \( t_u \) is the number of initial time steps provided in \( a_{\text{init}} \), and \( t_G \) is the additional time steps predicted by \( \mathcal{G}_\theta \).

\subsubsection*{Dataset Structure with True Sensitivities}

Each dataset sample is structured as:
\begin{equation}
\big[ a(\textbf{x}, t, p),\; \textbf{x},\; \mathcal{G}(a)(\textbf{x}, t, p),\; \frac{\partial \mathcal{G}(a)}{\partial p}(\textbf{x}, t, p) \big],
\end{equation}
where:
\begin{itemize}
    \item \( a(\textbf{x}, t, p) = (a_{\text{init}}(x, t), p(\textbf{x})) \) is the input function in \(\mathcal{U}\).
    \item \( \textbf{x} = (x_1, x_2), \dots, (x_{n_x}, x_{n_y}) \) are spatial grid points over \(\mathcal{X}\).
    \item \( \mathcal{G}(a)(\textbf{x}, t, p) = u(\textbf{x}, t, p) \) is the true PDE solution over \( t \in [t_u, n_t] \).
    \item \( \frac{\partial \mathcal{G}(a)}{\partial p}(\textbf{x}, t, p) \) is the true Jacobian with respect to \( p(\textbf{x}) \), obtained from a high-fidelity solver.
\end{itemize}

For each realization \( i \), the input is:
\begin{equation}
a^{(i)}(\textbf{x}, t, p) =
\begin{bmatrix}
a_{\text{init}}^{(i)}(x_1, x_2, t) & a_{\text{init}}^{(i)}(x_2, x_3, t) & \cdots & a_{\text{init}}^{(i)}(x_{n_x}, x_{n_y}, t) \\
p^{(i)}(x_1, x_2) & p^{(i)}(x_2, x_3) & \cdots & p^{(i)}(x_{n_x}, x_{n_y})
\end{bmatrix}, \quad n_x \times n_y \times t_u,
\end{equation}
and the output solution is:
\begin{equation}
\mathcal{G}(a^{(i)})(\textbf{x}, t, p) =
\begin{bmatrix}
u^{(i)}(x_1, x_2, t, p(x_1, x_2)) \\
u^{(i)}(x_2, x_3, t, p(x_2, x_3)) \\
\vdots \\
u^{(i)}(x_{n_x}, x_{n_y}, t, p(x_{n_x}, x_{n_y}))
\end{bmatrix}, \quad n_x \times n_y \times t_G.
\end{equation}

Since \( u(\textbf{x}, t, p) \) evolves over space and time, its sensitivity to the spatially varying parameter \( p(\textbf{x}) \) is a matrix-valued function. To capture the cumulative effect of \( p(\textbf{x}) \) on the solution, we compute the Jacobian at the final time step \( t = T \):
\begin{equation}
\frac{\partial \mathcal{G}(a)}{\partial p}(\textbf{x}, T, p) =
\begin{bmatrix}
\frac{\partial u}{\partial p}(x_1, x_2; x_1, x_2) & \frac{\partial u}{\partial p}(x_1, x_2; x_2, x_3) & \cdots & \frac{\partial u}{\partial p}(x_1, x_2; x_{n_x}, x_{n_y}) \\[6pt]
\frac{\partial u}{\partial p}(x_2, x_3; x_1, x_2) & \frac{\partial u}{\partial p}(x_2, x_3; x_2, x_3) & \cdots & \frac{\partial u}{\partial p}(x_2, x_3; x_{n_x}, x_{n_y}) \\[6pt]
\vdots & \vdots & \ddots & \vdots \\[6pt]
\frac{\partial u}{\partial p}(x_{n_x}, x_{n_y}; x_1, x_2) & \frac{\partial u}{\partial p}(x_{n_x}, x_{n_y}; x_2, x_3) & \cdots & \frac{\partial u}{\partial p}(x_{n_x}, x_{n_y}; x_{n_x}, x_{n_y})
\end{bmatrix}, \quad (n_x \times n_y) \times (n_x \times n_y).
\end{equation}
Each entry \( \frac{\partial u}{\partial p}(x_i, x_j; x_k, x_l) \) quantifies the response of the solution at \( (x_i, x_j) \) at \( t = T \) to changes in \( p(x_k, x_l) \), reflecting both local and nonlocal PDE-induced dependencies. By evaluating the Jacobian at \( t = T \), we encapsulate the temporal propagation of sensitivity from \( t = 0 \) to \( T \), as the final state \( u(x, T, p) \) integrates the system’s evolution. This approach aligns with SC-NO training, where \( L_s = \ell_s\left( \frac{\partial \hat{u}}{\partial p}, \frac{\partial u}{\partial p} \right) \) enforces sensitivity accuracy, reducing computational cost by focusing on the final time step while preserving critical spatial-parametric relationships for optimization.

\subsection*{Scaled Gaussian Random Field for Parameter Generation}

To generate spatially varying random parameters \( p(\mathbf{x}) \) across the computational domain, we define a scaled and bounded Gaussian random field (GRF) as:

\begin{equation}
p(\mathbf{x}) = \mathcal{T} \big( Z(\mathbf{x}); \text{p}_{\text{min}}, \text{p}_{\text{max}}, \sigma, \mathcal{X} \big),
\end{equation}

where \( \mathcal{T} \) is a transformation ensuring that \( p(\mathbf{x}) \) is confined within the predefined range \( [\text{p}_{\text{min}}, \text{p}_{\text{max}}] \). Specifically, the transformation is defined as:

\begin{equation}
\mathcal{T} \big( Z(\mathbf{x}) \big) = \text{p}_{\text{min}} + (\text{p}_{\text{max}} - \text{p}_{\text{min}}) \cdot 
\frac{\tanh \left( \sigma Z(\mathbf{x}) \right) - \min\limits_{\mathbf{y} \in \mathcal{X}} \tanh \left( \sigma Z(\mathbf{y}) \right)}
{\max\limits_{\mathbf{y} \in \mathcal{X}} \tanh \left( \sigma Z(\mathbf{y}) \right) - \min\limits_{\mathbf{y} \in \mathcal{X}} \tanh \left( \sigma Z(\mathbf{y}) \right)}.
\end{equation}

Here, \( Z(\mathbf{x}) \) is a zero-mean Gaussian random field sampled as:

\begin{equation}
Z(\mathbf{x}) \sim \mathcal{N} \left( 0, (-\Delta + 9I)^{-2} \right),
\end{equation}

where \( \Delta \) is the Laplacian operator, enforcing spatial smoothness and local correlation in \( p(\mathbf{x}) \). The parameter \( \sigma \) is a scaling factor that modulates the range of fluctuations in the transformed field. The transformation \( \mathcal{T} \) normalizes the GRF realization using the hyperbolic tangent function \( \tanh(\cdot) \), ensuring smooth transitions and bounded outputs. The resulting field \( p(\mathbf{x}) \) maintains spatial coherence while adhering to the predefined range, making it well-suited for parameterizing complex PDE systems. By employing this approach, we achieve a structured yet stochastic representation of spatially varying parameters, facilitating robust generalization in neural operator training.

\clearpage
\section{PDE1: Advection-Diffusion Equation with Spatially Distributed Velocity Field}
\label{sec:AD}

\renewcommand{\thefigure}{C.\arabic{figure}}  
\renewcommand{\thetable}{C.\arabic{table}}  
\setcounter{figure}{0}
\setcounter{table}{0}

\subsection*{Problem setup}
We model the transport of a concentration field \( C(\textbf{x}, t) \) in a two-dimensional spatial domain \(\mathcal{X} = [0, 1.0] \times [0, 1.0]\) over time \( t \in [0, 1.0] \), governed by the advection-diffusion equation with a spatially varying velocity field. Defining \( \textbf{x} = (x_1, x_2) \in \mathcal{X} \), the system is:
\begin{equation}
\frac{\partial C}{\partial t} + \nabla \cdot \bigl( \mathbf{u}(\textbf{x}) C \bigr) = D \nabla^2 C,
\label{eq:advection-diffusion}
\end{equation}
where \( C(\textbf{x}, t) \) is the concentration, \(\mathbf{u}(\textbf{x}) = [u_x(\textbf{x}), u_y(\textbf{x})] \) is the velocity field with components \( u_x(\textbf{x}) = u_x(x_1, x_2) \) and \( u_y(x) = u_y(x_1, x_2) \), and \( D = 0.005 \) is the constant diffusion coefficient. Periodic boundary conditions are applied in both \( x_1 \) and \( x_2 \) directions. The initial condition \( C_0(\mathbf{x}) = C(\mathbf{x}, 0) \) and velocity components \( u_x(\mathbf{x}) \) and \( u_y(\mathbf{x}) \) are sampled from Gaussian random fields (GRFs) with specified parameter ranges:

\begin{equation}
C_0(\mathbf{x}) \sim \text{GRF}(p_{\min} = -1.0, p_{\max} = 2.0, \sigma = 2, \mathcal{X}),
\end{equation}

\begin{equation}
u_x(\mathbf{x}), u_y(\mathbf{x}) \sim \text{GRF}(p_{\min} = -1.0, p_{\max} = 1.0, \sigma = 2, \mathcal{X}).
\end{equation}

\subsection*{Numerical Solver for the Advection-Diffusion Equation}

To generate training data for neural operator learning, we implemented a differentiable numerical solver for the advection-diffusion equation using \texttt{torchdiffeq}, reformulating the system as an ordinary differential equation (ODE):
\begin{equation}
\frac{dU}{dt} = \text{RHS}(\mathbf{x}, t, p),
\end{equation}
where \( U = [C] \) denotes the concentration field, and \( \text{RHS}(\mathbf{x}, t, p) \) encapsulates the advection \( \nabla \cdot (\mathbf{u}C) \) and diffusion \( \nabla^2 C \) terms with \( p = \mathbf{u} = [u_x, u_y] \) as the velocity field. Spatial discretization employs a finite difference method (FDM) on a \( 150 \times 150 \) grid within \( \mathcal{X} \), downsampled to \( S_x = S_y = 50 \) divisions, using a second-order central difference scheme for the Laplacian and an upwind scheme for the advection term to ensure stability. Periodic boundary conditions are enforced via ghost cells to maintain flux continuity. Time integration leverages an explicit fourth-order Runge–Kutta (RK4) scheme with a step size of \( 10^{-4} \), recording solutions at \( t = 0.01 \) intervals over \( [0, T = 1] \), yielding \( n_t = 100 \) time steps. Implemented with gradient tracking in \texttt{torchdiffeq}, the solver computes the Jacobian \( \frac{\partial U}{\partial p} \) with respect to the velocity parameters, supporting sensitivity-aware training in the SC-NO framework.

\subsection*{Neural Operator Learning for the Advection-Diffusion Equation}

We apply neural operator learning to the advection-diffusion equation defined over a spatial domain \( \mathcal{X} = [0, S_x] \times [0, S_y] \subset \mathbb{R}^2 \) with \( \mathbf{x} = (x_1, x_2) \in \mathcal{X} \) and time \( t \in [0, T=1.0] \), discretized into \( n_t \) time steps. The spatially varying velocity field \( \mathbf{u}(\mathbf{x}) = [u_x(\mathbf{x}), u_y(\mathbf{x})] \) serves as the parameter \( p(\mathbf{x}) \) within the input function \( a(\mathbf{x}, t, p) \in \mathcal{U} \), which also includes the first \( t_u \) time steps of the solution \( u(\mathbf{x}, t, p) \). We train a neural operator \( \mathcal{G}_\theta \) in the SC-NO framework to predict the subsequent \( t_G = n_t - t_u \) time steps, guided by the Jacobian \( \frac{\partial u}{\partial p} \), using the mapping:
\begin{equation}
\mathcal{G}_\theta: \left( \mathcal{X} \times [0, t_u], p(\mathbf{x}) \right) \to u(\mathbf{x}, t, p), \quad \text{for } (\mathbf{x}, t) \in \mathcal{X} \times [t_u, T].
\end{equation}
This approach unifies the spatial-temporal input and parameter field to learn a solution operator spanning \( [t_u, T] \) across all PDE instances.

\subsection*{Additional results for PDE1}

\begin{table}[ht!]
\centering
\caption{\scriptsize\textbf{Comparison of Error Metrics Across Models Trained with Different Sample Sizes (Forward)}}
\label{tab:pde1_forward_nostd}
\scriptsize
\setlength{\tabcolsep}{3pt}
\begin{tabular}{lcccccccc}
\hline
\textbf{Model} & \multicolumn{2}{c}{\textbf{1000 Samples}} & \multicolumn{2}{c}{\textbf{500 Samples}} & \multicolumn{2}{c}{\textbf{200 Samples}} & \multicolumn{2}{c}{\textbf{100 Samples}} \\
 & \textbf{Rel. \(L^2\)} & \textbf{MAE} & \textbf{Rel. \(L^2\)} & \textbf{MAE} & \textbf{Rel. \(L^2\)} & \textbf{MAE} & \textbf{Rel. \(L^2\)} & \textbf{MAE} \\
\hline
FNO        & \(5.1 \times 10^{-2}\) & \(5.6 \times 10^{-3}\) & \(1.7 \times 10^{-1}\) & \(1.9 \times 10^{-2}\) & \(8.5 \times 10^{-1}\) & \(9.4 \times 10^{-2}\) & \(1.4 \times 10^{0}\) & \(1.5 \times 10^{-1}\) \\
WNO        & \(6.2 \times 10^{-2}\) & \(6.6 \times 10^{-3}\) & \(2.0 \times 10^{-1}\) & \(2.2 \times 10^{-2}\) & \(9.8 \times 10^{-1}\) & \(1.1 \times 10^{-1}\) & \(1.6 \times 10^{0}\) & \(1.8 \times 10^{-1}\) \\
DeepONet   & \(6.2 \times 10^{-2}\) & \(6.9 \times 10^{-3}\) & \(2.1 \times 10^{-1}\) & \(2.3 \times 10^{-2}\) & \(1.0 \times 10^{0}\) & \(1.1 \times 10^{-1}\) & \(1.7 \times 10^{0}\) & \(1.8 \times 10^{-1}\) \\
SC-FNO     & \(\mathbf{2.6 \times 10^{-2}}\) & \(\mathbf{1.9 \times 10^{-3}}\) & \(\mathbf{5.5 \times 10^{-2}}\) & \(\mathbf{6.2 \times 10^{-3}}\) & \(\mathbf{1.2 \times 10^{-1}}\) & \(\mathbf{1.4 \times 10^{-2}}\) & \(\mathbf{1.8 \times 10^{-1}}\) & \(\mathbf{2.0 \times 10^{-2}}\) \\
SC-WNO     & \(2.8 \times 10^{-2}\) & \(2.1 \times 10^{-3}\) & \(6.0 \times 10^{-2}\) & \(6.9 \times 10^{-3}\) & \(1.4 \times 10^{-1}\) & \(1.5 \times 10^{-2}\) & \(2.0 \times 10^{-1}\) & \(2.2 \times 10^{-2}\) \\
SC-DeepONet& \(3.2 \times 10^{-2}\) & \(2.5 \times 10^{-3}\) & \(7.5 \times 10^{-2}\) & \(8.4 \times 10^{-3}\) & \(1.6 \times 10^{-1}\) & \(1.9 \times 10^{-2}\) & \(2.4 \times 10^{-1}\) & \(2.7 \times 10^{-2}\) \\
\hline
\end{tabular}
\end{table}

\begin{table}[ht!]
\centering
\caption{\scriptsize\textbf{Comparison of Error Metrics Across Models Trained with Different Sample Sizes (Inverse)}}
\label{tab:pde1_inverse_nostd} 
\scriptsize
\setlength{\tabcolsep}{3pt}
\begin{tabular}{lcccccccc}
\hline
\textbf{Model} & \multicolumn{2}{c}{\textbf{1000 Samples}} & \multicolumn{2}{c}{\textbf{500 Samples}} & \multicolumn{2}{c}{\textbf{200 Samples}} & \multicolumn{2}{c}{\textbf{100 Samples}} \\
 & \textbf{Rel. \(L^2\)} & \textbf{MAE} & \textbf{Rel. \(L^2\)} & \textbf{MAE} & \textbf{Rel. \(L^2\)} & \textbf{MAE} & \textbf{Rel. \(L^2\)} & \textbf{MAE} \\
\hline
FNO & \(1.3 \times 10^{-1}\) & \(6.2 \times 10^{-2}\) & \(2.1 \times 10^{-1}\) & \(1.0 \times 10^{-1}\) & \(3.2 \times 10^{-1}\) & \(1.5 \times 10^{-1}\) & \(5.1 \times 10^{-1}\) & \(2.4 \times 10^{-1}\) \\
WNO & \(1.6 \times 10^{-1}\) & \(7.4 \times 10^{-2}\) & \(2.5 \times 10^{-1}\) & \(1.2 \times 10^{-1}\) & \(3.8 \times 10^{-1}\) & \(1.8 \times 10^{-1}\) & \(6.1 \times 10^{-1}\) & \(2.8 \times 10^{-1}\) \\
DeepONet & \(1.6 \times 10^{-1}\) & \(7.6 \times 10^{-2}\) & \(2.6 \times 10^{-1}\) & \(1.2 \times 10^{-1}\) & \(3.9 \times 10^{-1}\) & \(1.8 \times 10^{-1}\) & \(6.3 \times 10^{-1}\) & \(2.9 \times 10^{-1}\) \\
SC-FNO & \(\mathbf{1.3 \times 10^{-2}}\) & \(\mathbf{6.2 \times 10^{-3}}\) & \(\mathbf{1.5 \times 10^{-2}}\) & \(\mathbf{7.1 \times 10^{-3}}\) & \(\mathbf{1.8 \times 10^{-2}}\) & \(\mathbf{8.2 \times 10^{-3}}\) & \(\mathbf{2.1 \times 10^{-2}}\) & \(\mathbf{1.0 \times 10^{-2}}\) \\
SC-WNO & \(1.5 \times 10^{-2}\) & \(6.9 \times 10^{-3}\) & \(1.7 \times 10^{-2}\) & \(8.0 \times 10^{-3}\) & \(2.0 \times 10^{-2}\) & \(9.2 \times 10^{-3}\) & \(2.4 \times 10^{-2}\) & \(1.1 \times 10^{-2}\) \\
SC-DeepONet & \(1.8 \times 10^{-2}\) & \(8.4 \times 10^{-3}\) & \(2.1 \times 10^{-2}\) & \(9.7 \times 10^{-3}\) & \(2.4 \times 10^{-2}\) & \(1.1 \times 10^{-2}\) & \(2.9 \times 10^{-2}\) & \(1.4 \times 10^{-2}\) \\
\hline
\end{tabular}
\end{table}



\clearpage
\section{PDE2: Navier--Stokes Equations with Spalart--Allmaras Closure}
\label{sec:NSE-SA}

\renewcommand{\thefigure}{D.\arabic{figure}}  
\renewcommand{\thetable}{D.\arabic{table}}  
\setcounter{figure}{0}
\setcounter{table}{0}

\subsection*{Problem setup}
We simulate two-dimensional incompressible flows in the unit square domain \( \mathcal{X} = [0, 1] \times [0, 1] \) over the time interval \( t \in [0, T] \) with \( T = 25 \). The governing equations are expressed in vorticity--stream function form, coupled with the one-equation Spalart--Allmaras (S-A) turbulence model. Let \( \mathbf{x} = (x_1, x_2) \in \mathcal{X} \). The system reads:
\begin{align}
\nabla^2 \Psi &= -\Omega, \label{eq:streamfunction} \\[4pt]
\frac{\partial \Omega}{\partial t}
+ \frac{\partial \Psi}{\partial x_2} \frac{\partial \Omega}{\partial x_1}
- \frac{\partial \Psi}{\partial x_1} \frac{\partial \Omega}{\partial x_2}
&= \nabla \cdot \bigl[ (\nu + \nu_t) \nabla \Omega \bigr] + f(\mathbf{x}; \alpha, \beta), \label{eq:vorticity} \\[6pt]
\frac{\partial \tilde{\nu}}{\partial t}
+ \frac{\partial \Psi}{\partial x_2} \frac{\partial \tilde{\nu}}{\partial x_1}
- \frac{\partial \Psi}{\partial x_1} \frac{\partial \tilde{\nu}}{\partial x_2}
&= c_{b1} \tilde{S} \tilde{\nu}
+ \frac{1}{\sigma} \left[ \nabla \cdot \bigl( (\nu + \tilde{\nu}) \nabla \tilde{\nu} \bigr) + c_{b2} (\nabla \tilde{\nu})^2 \right]
- c_{w1} f_w \left( \frac{\tilde{\nu}}{d} \right)^2, \label{eq:SA}
\end{align}
where $\Psi$ denotes the stream function, $\Omega$ the vorticity, and the molecular viscosity is taken as $\nu \in [0.001,\,0.002]$, which corresponds to Reynolds numbers in the range $500$--$1000$. The turbulent eddy viscosity is modeled as $\nu_t = \tilde{\nu}\, f_{v1}$.
\[
f_{v1} = \frac{\chi^3}{\chi^3 + c_{v1}^3}, \quad \chi = \frac{\tilde{\nu}}{\nu}.
\]
The term \( f(\mathbf{x}; \alpha, \beta) \) is a parametric body force, and \( d \) is the distance to the nearest wall.

Physical boundary conditions are no-slip walls (\( \mathbf{u} = 0 \)), implying \( \Psi = 0 \) and \( \partial \Psi / \partial n = 0 \) on \( \partial \mathcal{X} \). Vorticity vanishes on walls via the initial mask \( m(x_1, x_2) = \sin(\pi x_1) \sin(\pi x_2) \).

The numerical solver uses zero wall boundary conditions in both directions for its finite difference discretization and fast Poisson solving. Physical wall conditions are enforced weakly through:
\begin{itemize}
    \item Initial vorticity masking to ensure \( \Omega_0 = 0 \) on wall-adjacent cells,
    \item Spectral projection of \( \Psi \) after each Poisson solve to satisfy \( \Psi = 0 \) on \( \partial \mathcal{X} \),
    \item Thom’s formula at each time step to enforce \( \mathbf{u} = 0 \) on walls.
\end{itemize}

The forward simulation uses a uniform fine grid of resolution \( 64 \times 64 \). The initial modified eddy viscosity is sampled from a Gaussian random field (GRF):
\[
\tilde{\nu}_0(\mathbf{x}) \sim \text{GRF}\bigl(p_{\min} = -2.5 \times 10^{-4},\; p_{\max} = 2.5 \times 10^{-4},\; \sigma = 1,\; \mathcal{X}\bigr).
\]

The initial vorticity \( \Omega_0 \) is generated on a coarse grid of resolution \( N_c \times N_c \), where \( N_c = 2^n \) and \( n \in \{2, 3, 4, 5, 6\} \) (corresponding to \( N_c \times N_c \in \{4\times4, 8\times8, 16\times16, 32\times32, 64\times64\} \)). For a given \( n \), the coarse field is
\[
\Omega_0^c(\mathbf{x}) = m(x_1, x_2) \cdot \text{GRF}\bigl(p_{\min} = -2.5,\; p_{\max} = 2.5,\; \sigma = 2,\; \mathcal{X}_c^n\bigr),
\]
upsampled to \( 64 \times 64 \) via pixel-wise repetition by factor \( 2^{6-n} \). This ensures zero vorticity on wall-adjacent fine-grid cells and introduces multiscale initial perturbations.

The forcing is
\[
f(\mathbf{x}; \alpha, \beta) = 0.1 \left[ \sin\!\bigl(\alpha (x_1 + x_2)\bigr) + \cos\!\bigl(\beta (x_1 + x_2)\bigr) \right],
\]
with \( \alpha, \beta \in [0, 3\pi] \) sampled uniformly.

\subsection*{Numerical Solver for the Turbulent Navier--Stokes Equations}

The system is recast as an ODE:
\[
\frac{d\mathbf{U}}{dt} = \text{RHS}(\mathbf{x}, t; p), \quad \mathbf{U} = [\Omega, \tilde{\nu}],
\]
with \( \Psi \) recovered via a fast Poisson solver (e.g., a Sine Transform) under zero wall computational boundary conditions. Spatial discretization uses second-order central finite differences on uniform grids. Time integration uses the Fourth Order Runge-Kutta solver from \texttt{torchdiffeq}, storing solutions at \( \Delta t = 0.1 \) over \( [0, 25.0] \), yielding \( n_t = 250 \) snapshots. The solver is fully differentiable, enabling adjoint-based gradients with respect to \( \Omega_0 \) and the forcing term.

\subsection*{Neural Operator Learning for the Turbulent Navier--Stokes Equations}

We employ neural operator learning to tackle the turbulent Navier--Stokes equations with the Spalart--Allmaras model over a spatial domain $\mathcal{X}$ and time $t \in [0, 25.0]$, discretized into $n_t$ snapshots (here, $n_t=250$ at $\Delta t=0.1\,\mathrm{s}$). The parametric, spatially varying forcing $f(\mathbf{x}; \alpha,\beta)$ acts as the parameter field $p(\mathbf{x})$ within the input function $a(\mathbf{x}, t, p)\in\mathcal{U}$. The solution state is $u(\mathbf{x}, t, p) = [\Psi(\mathbf{x}, t),\, \Omega(\mathbf{x}, t),\, \tilde{\nu}(\mathbf{x}, t)]$.

\paragraph{(I) Sequential multi-step forecasting (context $\rightarrow$ full-horizon prediction).}
We train a neural operator $\mathcal{G}_\theta$ within the SC-NO framework that consumes a short context of the trajectory together with the forcing field and predicts the remainder of the sequence in one shot. Concretely, the input consists of the initial segment of the solution field \( u(\mathbf{x}, t, p) \) over \( t \in [0, t_u] \) and the parameter field \( p(\mathbf{x}) = f(\mathbf{x}; \alpha, \beta) \):
\[
\mathcal{G}_\theta:\; \big(u(\mathbf{x}, t, p)\big)_{t \in [0, t_u]},\, p(\mathbf{x})\;\longrightarrow\; u(\mathbf{x}, t, p),
\quad (\mathbf{x}, t) \in \mathcal{X} \times [t_u,\,25.0].
\]
The training loss aggregates the state error over all predicted times together with a sensitivity regularizer. This regime corresponds to the “sequential prediction” setting used in the main text.

Additionally, to evaluate the model’s sensitivity to spatial resolution, we construct a family of simulations with multiresolution initial vorticity fields. Specifically, the base vorticity field \( \Omega_0^c(\mathbf{x}) \) is generated on coarse grids \( \mathcal{X}_c^n \) with \( N_c = 2^n \), \( n \in \{2,3,4,5,6\} \), and then upsampled to the working grid \( \mathcal{X} \) for training. Each resolution level thus defines a distinct realization of the input sequence \( \big(u(\mathbf{x}, t, p)\big)_{t \in [0, t_u]} \), differing only in the spatial smoothness and detail of the initial vorticity \( \Omega_0(\mathbf{x}) \). The operator learns the mapping  
\[
\mathcal{G}_\theta:\; \big(u(\mathbf{x}, t, p)\big)_{t \in [0, t_u]},\, p(\mathbf{x})\;\longrightarrow\; u(\mathbf{x}, t, p),
\quad (\mathbf{x}, t) \in \mathcal{X} \times [t_u,\,25.0],
\]
allowing us to systematically assess how prediction accuracy and learned sensitivities vary across different resolutions of the initial vorticity field.

\paragraph{(II) One-step transition with autoregressive rollout (rollout model).}  
In this regime, we train a \emph{transition operator} that advances the state by a single time step given the current state and the forcing field. During training, the operator is supervised on consecutive snapshot pairs using teacher forcing. Specifically, it learns the mapping
\[
\mathcal{F}_\theta:\; \big(u(\cdot,t,p),\, p(\mathbf{x})\big)\;\longrightarrow\; u(\cdot,t+\Delta t,p),
\]
with training pairs $\big(u(\cdot,t,p),\, u(\cdot,t+\Delta t,p)\big)$ drawn from the first 125 snapshots, corresponding to $t \in [0,12.5]$\,s. To improve robustness, the predicted states used during training are perturbed with controlled noise before being fed back as inputs, preventing the model from overfitting to perfectly clean trajectories.  

At inference time, the learned operator is applied recursively in an autoregressive manner to generate full trajectories. Starting from the initial condition $u(\cdot,0,p)$, we roll the model forward for 250 steps to cover the entire horizon $[0,25.0]$\,s:  
\[
\hat{u}(\cdot,t+\Delta t,p)=\mathcal{F}_\theta\!\big(\hat{u}(\cdot,t,p),\,p\big), \quad t=0,\Delta t,2\Delta t,\ldots,25.0-\Delta t.
\]
This formulation explicitly probes long-horizon stability and error accumulation: the model is trained only on short-range one-step transitions ($0$–$12.5$\,s) but evaluated on the full $0$–$25$\,s trajectory (250 snapshots).  

\subsection*{Adjoint-Based Sensitivity Computation for the Turbulent Navier--Stokes Equations}
\label{sec:adjoint_nse}

To compute sensitivities of the final-time vorticity field $\Omega(\mathbf{x}, T)$ with respect to the initial vorticity $\Omega_0(\mathbf{x})$ and the forcing $f(\mathbf{x}, t; \alpha, \beta)$, we derive the continuous adjoint system. The adjoint method is a powerful technique for efficiently computing the sensitivity of a scalar objective functional, $J$, to a large number of input parameters or fields (such as initial conditions $\Omega_0$ or forcing $f$). Its primary advantage is that the computational cost of obtaining the gradient is \textbf{independent} of the number of input parameters. The method relies on constructing a Lagrangian, $\mathcal{L}$, which augments the objective functional with the governing PDEs, enforced as constraints via Lagrange multipliers (the adjoint variables, e.g., $\lambda_\Omega, \lambda_{\tilde{\nu}}$). By requiring the first variation of the Lagrangian ($\delta\mathcal{L}$) to be zero with respect to the state variables, a new set of linear PDEs—the \textbf{adjoint equations}—is derived.

These adjoint equations are solved \textbf{backward in time}, starting from a terminal condition derived from the objective functional at $t=T$. The resulting adjoint field, $\lambda(\mathbf{x}, 0)$, at the initial time $t=0$ (or integrated over time for forcing) provides the exact functional derivative (gradient) of $J$ with respect to the parameters. This allows the gradient for all parameters to be computed from just one forward solve of the primal equations and one backward solve of the adjoint equations.

\subsubsection*{Cost Functional}

The cost functional is the integrated final vorticity:
\begin{equation}
J = \int_{\mathcal{X}} \Omega(\mathbf{x}, T) \, d\mathbf{x}.
\end{equation}

\subsubsection*{Lagrangian Formulation}

The Lagrangian is
\begin{align}
\mathcal{L} &= \int_{\mathcal{X}} \Omega(T) \, d\mathbf{x} \notag \\[4pt]
&\quad + \int_0^T \int_{\mathcal{X}} \lambda_{\Omega} \left[
\frac{\partial \Omega}{\partial t}
+ (\mathbf{u} \cdot \nabla) \Omega
- \nabla \cdot \bigl( \nu_{\text{eff}}(\tilde{\nu}) \nabla \Omega \bigr)
- f
\right] d\mathbf{x} \, dt \notag \\[4pt]
&\quad + \int_0^T \int_{\mathcal{X}} \lambda_{\tilde{\nu}} \left[
\frac{\partial \tilde{\nu}}{\partial t}
+ (\mathbf{u} \cdot \nabla) \tilde{\nu}
- \frac{1}{\sigma} \nabla \cdot \bigl( (\nu + \tilde{\nu}) \nabla \tilde{\nu} \bigr)
- \frac{c_{b2}}{\sigma} (\nabla \tilde{\nu})^2
- R(\Omega, \tilde{\nu})
\right] d\mathbf{x} \, dt \notag \\[4pt]
&\quad + \int_0^T \int_{\mathcal{X}} \lambda_{\Psi} \left[
\nabla^2 \Psi + \Omega
\right] d\mathbf{x} \, dt,
\end{align}
where $\mathbf{u} = \nabla^\perp \Psi$, $\nu_{\text{eff}}(\tilde{\nu}) = \nu + \tilde{\nu} f_{v1}(\tilde{\nu})$, and
\[
R(\Omega, \tilde{\nu}) = c_{b1} \tilde{S} \tilde{\nu} - c_{w1} f_w \left( \frac{\tilde{\nu}}{d} \right)^2.
\]

\subsubsection*{Adjoint Equations}

The first variation $\delta \mathcal{L} = 0$ with respect to $(\Psi, \Omega, \tilde{\nu})$ yields the adjoint system. Integration by parts in space and time is performed. Boundary terms are eliminated using primal wall conditions.

\textbf{Adjoint Poisson Equation} (from $\delta \Psi$):
\begin{equation}
\nabla^2 \lambda_{\Psi}
= -\nabla^\perp \cdot (\lambda_{\Omega} \nabla \Omega)
  -\nabla^\perp \cdot (\lambda_{\tilde{\nu}} \nabla \tilde{\nu}).
\end{equation}

\textbf{Adjoint Vorticity Equation} (from $\delta \Omega$):
\begin{equation}
-\frac{\partial \lambda_{\Omega}}{\partial t}
= \mathbf{u} \cdot \nabla \lambda_{\Omega}
  + \nabla \cdot \bigl( \nu_{\text{eff}} \nabla \lambda_{\Omega} \bigr)
  + \lambda_{\Psi}
  + \lambda_{\tilde{\nu}} \frac{\partial R}{\partial \Omega},
\end{equation}
with
\[
\frac{\partial R}{\partial \Omega}
= c_{b1} \tilde{\nu} \frac{\partial \tilde{S}}{\partial \Omega},
\qquad
\frac{\partial \tilde{S}}{\partial \Omega}
= \operatorname{sgn}(\Omega)
\quad \text{(regularized as $\tanh(k \Omega)$ in practice)}.
\]

\textbf{Adjoint Spalart--Allmaras Equation} (from $\delta \tilde{\nu}$):
\begin{equation}
-\frac{\partial \lambda_{\tilde{\nu}}}{\partial t}
= \mathbf{u} \cdot \nabla \lambda_{\tilde{\nu}}
  + \frac{1}{\sigma} \nabla \cdot \bigl( (\nu + \tilde{\nu}) \nabla \lambda_{\tilde{\nu}} \bigr)
  - \nabla \cdot \left( \frac{2 c_{b2}}{\sigma} \lambda_{\tilde{\nu}} \nabla \tilde{\nu} \right)
  + \lambda_{\tilde{\nu}} \frac{\partial R}{\partial \tilde{\nu}}
  - \frac{\partial \nu_{\text{eff}}}{\partial \tilde{\nu}} (\nabla \lambda_{\Omega} \cdot \nabla \Omega).
\end{equation}

\subsubsection*{Adjoint Terminal and Boundary Conditions}

\textbf{Terminal conditions} (at $t = T$):
\begin{align}
\lambda_{\Omega}(\mathbf{x}, T) &= -1, \\
\lambda_{\tilde{\nu}}(\mathbf{x}, T) &= 0.
\end{align}

\textbf{Boundary conditions} (on $\partial \mathcal{X}$, $0 \leq t < T$):
\begin{itemize}
    \item $\lambda_{\Psi} = 0$ \quad (from primal $\Psi = 0$),
    \item $\lambda_{\tilde{\nu}} = 0$ \quad (from primal $\tilde{\nu} = 0$),
    \item $\lambda_{\Omega} = 0$ \quad (from primal $\Omega = 0$: $\delta \Omega = 0$ on boundary $\implies$ adjoint inherits Dirichlet).
\end{itemize}

The adjoint system is solved backward in time from $t = T$ to $t = 0$.

\subsubsection*{Sensitivity with Respect to Initial Vorticity $\Omega_0$}

From the time-boundary term at $t=0$:
\begin{equation}
\boxed{
\frac{\delta J}{\delta \Omega_0}(\mathbf{x})
= -\lambda_{\Omega}(\mathbf{x}, 0)
}
\end{equation}

\subsubsection*{Sensitivity with Respect to Forcing Field $f$}

From the explicit $f$ term in the Lagrangian:
\begin{equation}
\boxed{
\frac{\delta J}{\delta f}(\mathbf{x}, t)
= -\lambda_{\Omega}(\mathbf{x}, t),
\qquad
\frac{\delta J}{\delta f}(\mathbf{x})
= -\int_0^T \lambda_{\Omega}(\mathbf{x}, t) \, dt
}
\end{equation}

These sensitivities are exact in the continuous setting and provide the full gradient of the final vorticity field with respect to initial conditions and forcing.

\clearpage
\subsection*{Additional results for PDE2}
\begin{table}[h!]
\centering
\caption{\textbf{Comparison of Error Metrics Across Models Trained with Different Sample Sizes (Forward)}}
\label{tab:pde2_forward_nostd}
\scriptsize
\setlength{\tabcolsep}{3pt}
\begin{tabular}{lcccccccc}
\hline
\textbf{Model} & \multicolumn{2}{c}{\textbf{1000 Samples}} & \multicolumn{2}{c}{\textbf{500 Samples}} & \multicolumn{2}{c}{\textbf{200 Samples}} & \multicolumn{2}{c}{\textbf{100 Samples}} \\
 & \textbf{Rel. \(L^2\)} & \textbf{MAE} & \textbf{Rel. \(L^2\)} & \textbf{MAE} & \textbf{Rel. \(L^2\)} & \textbf{MAE} & \textbf{Rel. \(L^2\)} & \textbf{MAE} \\
\hline
FNO        & \(6.9 \times 10^{-2}\) & \(6.8 \times 10^{-3}\) & \(1.9 \times 10^{-1}\) & \(1.8 \times 10^{-2}\) & \(5.7 \times 10^{-1}\) & \(5.4 \times 10^{-2}\) & \(1.3 \times 10^{0}\) & \(1.3 \times 10^{-1}\) \\
WNO        & \(7.5 \times 10^{-2}\) & \(7.3 \times 10^{-3}\) & \(2.0 \times 10^{-1}\) & \(2.0 \times 10^{-2}\) & \(6.0 \times 10^{-1}\) & \(5.8 \times 10^{-2}\) & \(1.4 \times 10^{0}\) & \(1.4 \times 10^{-1}\) \\
DeepONet   & \(8.1 \times 10^{-2}\) & \(7.9 \times 10^{-3}\) & \(2.2 \times 10^{-1}\) & \(2.1 \times 10^{-2}\) & \(6.6 \times 10^{-1}\) & \(6.3 \times 10^{-2}\) & \(1.5 \times 10^{0}\) & \(1.5 \times 10^{-1}\) \\
SC-FNO     & \(\mathbf{2.9 \times 10^{-2}}\)& \(\mathbf{2.2 \times 10^{-3}}\) & \(\mathbf{3.9 \times 10^{-2}}\) & \(\mathbf{7.4 \times 10^{-3}}\) & \(\mathbf{6.6 \times 10^{-2}}\) & \(\mathbf{1.3 \times 10^{-2}}\) & \(\mathbf{1.5 \times 10^{-1}}\) & \(\mathbf{3.0 \times 10^{-2}}\) \\
SC-WNO     & \(3.4\times 10^{-2}\)& \(2.3 \times 10^{-3}\) & \(3.9 \times 10^{-2}\) & \(7.6 \times 10^{-3}\) & \(6.8 \times 10^{-2}\) & \(1.3 \times 10^{-2}\) & \(1.7 \times 10^{-1}\) & \(3.1 \times 10^{-2}\) \\
SC-DeepONet& \(4.1 \times 10^{-2}\)& \(2.9 \times 10^{-3}\) & \(5.1 \times 10^{-2}\) & \(9.7 \times 10^{-3}\) & \(8.6 \times 10^{-2}\) & \(1.7 \times 10^{-2}\) & \(2.0 \times 10^{-1}\) & \(3.9 \times 10^{-2}\) \\
\hline
\end{tabular}
\end{table}


\begin{table}[h!]
\centering
\caption{\scriptsize\textbf{Comparison of Error Metrics Across Models Trained with Different Sample Sizes (Inverse) PDE2}}
\label{tab:pde2_inverse_nostd}
\scriptsize
\setlength{\tabcolsep}{3pt}
\begin{tabular}{lcccccccc}
\hline
\textbf{Model} & \multicolumn{2}{c}{\textbf{1000 Samples}} & \multicolumn{2}{c}{\textbf{500 Samples}} & \multicolumn{2}{c}{\textbf{200 Samples}} & \multicolumn{2}{c}{\textbf{100 Samples}} \\
 & \textbf{Rel. \(L^2\)} & \textbf{MAE} & \textbf{Rel. \(L^2\)} & \textbf{MAE} & \textbf{Rel. \(L^2\)} & \textbf{MAE} & \textbf{Rel. \(L^2\)} & \textbf{MAE} \\
\hline
FNO        & \(4.7 \times 10^{-1}\) & \(5.0 \times 10^{-2}\) & \(6.7 \times 10^{-1}\) & \(7.2 \times 10^{-2}\) & \(9.6 \times 10^{-1}\) & \(1.0 \times 10^{-1}\) & \(1.4 \times 10^{0}\) & \(1.4 \times 10^{-1}\) \\
WNO        & \(5.6 \times 10^{-1}\) & \(5.9 \times 10^{-2}\) & \(7.9 \times 10^{-1}\) & \(8.4 \times 10^{-2}\) & \(1.1 \times 10^{0}\) & \(1.2 \times 10^{-1}\) & \(1.6 \times 10^{0}\) & \(1.7 \times 10^{-1}\) \\
DeepONet   & \(5.8 \times 10^{-1}\) & \(6.1 \times 10^{-2}\) & \(8.0 \times 10^{-1}\) & \(8.7 \times 10^{-2}\) & \(1.2 \times 10^{0}\) & \(1.2 \times 10^{-1}\) & \(1.6 \times 10^{0}\) & \(1.7 \times 10^{-1}\) \\
SC-FNO     & \(\mathbf{6.1 \times 10^{-2}}\) & \(\mathbf{5.2 \times 10^{-3}}\) & \(\mathbf{6.8 \times 10^{-2}}\) & \(\mathbf{5.8 \times 10^{-3}}\) & \(\mathbf{8.0 \times 10^{-2}}\) & \(\mathbf{6.8 \times 10^{-3}}\) & \(\mathbf{9.6 \times 10^{-2}}\) & \(\mathbf{8.2 \times 10^{-3}}\) \\
SC-WNO     & \(6.8 \times 10^{-2}\) & \(5.8 \times 10^{-3}\) & \(7.6 \times 10^{-2}\) & \(6.5 \times 10^{-3}\) & \(9.0 \times 10^{-2}\) & \(7.6 \times 10^{-3}\) & \(1.1 \times 10^{-1}\) & \(9.2 \times 10^{-3}\) \\
SC-DeepONet& \(8.3 \times 10^{-2}\) & \(7.0 \times 10^{-3}\) & \(9.3 \times 10^{-2}\) & \(7.9 \times 10^{-3}\) & \(1.1 \times 10^{-1}\) & \(9.3 \times 10^{-3}\) & \(1.3 \times 10^{-1}\) & \(1.1 \times 10^{-2}\) \\
\hline
\end{tabular}
\end{table}

\clearpage
\clearpage
\section{PDE3: 2D Shallow Water Equations for the Tohoku 2011 Tsunami}
\label{sec:SWE-Tohoku}

\renewcommand{\thefigure}{E.\arabic{figure}}  
\renewcommand{\thetable}{E.\arabic{table}}  
\setcounter{figure}{0}
\setcounter{table}{0}

\subsection*{Overview}
We evaluate the feasibility of near-real-time large-scale tsunami inversion and forecasting using SC-NOs, governed by the hyperbolic Shallow Water equations, based on the 2011 Tohoku event. Variability arises from earthquake-induced seafloor deformation computed using the Okada model \citep{grilli2013numerical} with different geometries and epicenter locations. We use noise-contaminated synthetic water-level observations at sea-buoy locations from the first 30 minutes as observations for the inverse source identification, followed by forward forecasting. The inversion first estimates an initial guess of the seafloor deformation field, restricts it to the Okada representation, and then optimizes the Okada model parameters via the pretrained neural operators (Methods). All neural operators (FNO and SC-FNO) are pretrained on high-resolution simulations from a observationally-validated finite-volume solver (Supplementary Figure~\ref{fig:tohoku_validation}), coarsened to the NO grid resolutions. They receive deformation fields $\Delta z_b(\mathbf{x})$ and an initial water stage ($t_u = 1$) as inputs and predict the remaining $t_G = 50$ time steps. We calculate the sensitivity (Jacobian) of the final water stage with respect to bed topography using adjoint-sensitivity analysis.

\subsection*{Problem setup}
We solve the shallow water equations (SWE) with Manning friction over a two-dimensional ocean domain \( \mathcal{X} = [0, L] \times [0, L] \) and time \( t \in [0, T] \). This formulation captures wave propagation, coastal inundation, and wet-dry dynamics. Given spatial coordinates \(\textbf{x} = (x, y) \in \mathcal{X} \), the governing SWE system is:

\begin{equation}
\frac{\partial}{\partial t}
\begin{pmatrix}
h \\[3pt]
m_x \\[3pt]
m_y
\end{pmatrix}
+ \nabla \cdot
\begin{pmatrix}
h \mathbf{u} \\[5pt]
h u \mathbf{u} + \tfrac{1}{2}\,g\,h^2 \,\mathbf{I}_x \\[3pt]
h v \mathbf{u} + \tfrac{1}{2}\,g\,h^2 \,\mathbf{I}_y
\end{pmatrix}
=
\begin{pmatrix}
0 \\[4pt]
-\,g\,h\,\frac{\partial z_b}{\partial x} \;-\; \bigl(\text{friction}_x\bigr) \\[6pt]
-\,g\,h\,\frac{\partial z_b}{\partial y} \;-\; \bigl(\text{friction}_y\bigr)
\end{pmatrix}.
\end{equation}

Here, \( h \) is the water depth, \( \mathbf{m} = (m_x, m_y) = (h u, h v) \) is the depth-integrated momentum, \( \mathbf{u} = (u, v) \) is the velocity, and \( z_b(x, y) \) represents the bed elevation. The source terms include the bed slope and Manning friction:

\begin{equation}
\text{friction}_x = g n^2 \|\mathbf{u}\| \frac{h u}{h^{7/3}}, 
\quad
\text{friction}_y = g n^2 \|\mathbf{u}\| \frac{h v}{h^{7/3}}.
\end{equation}
To represent wet-dry dynamics, dry regions are modeled by setting \( h = 0 \) when the water depth falls below a threshold. no reflection boundary conditions are imposed in both \( x \) and \( y \) directions to simulate an open ocean. 

\subsection*{Modeling Seafloor Deformation Using the Okada Dislocation Model}
\begin{figure}
    \centering
    \includegraphics[width=0.65\textwidth]{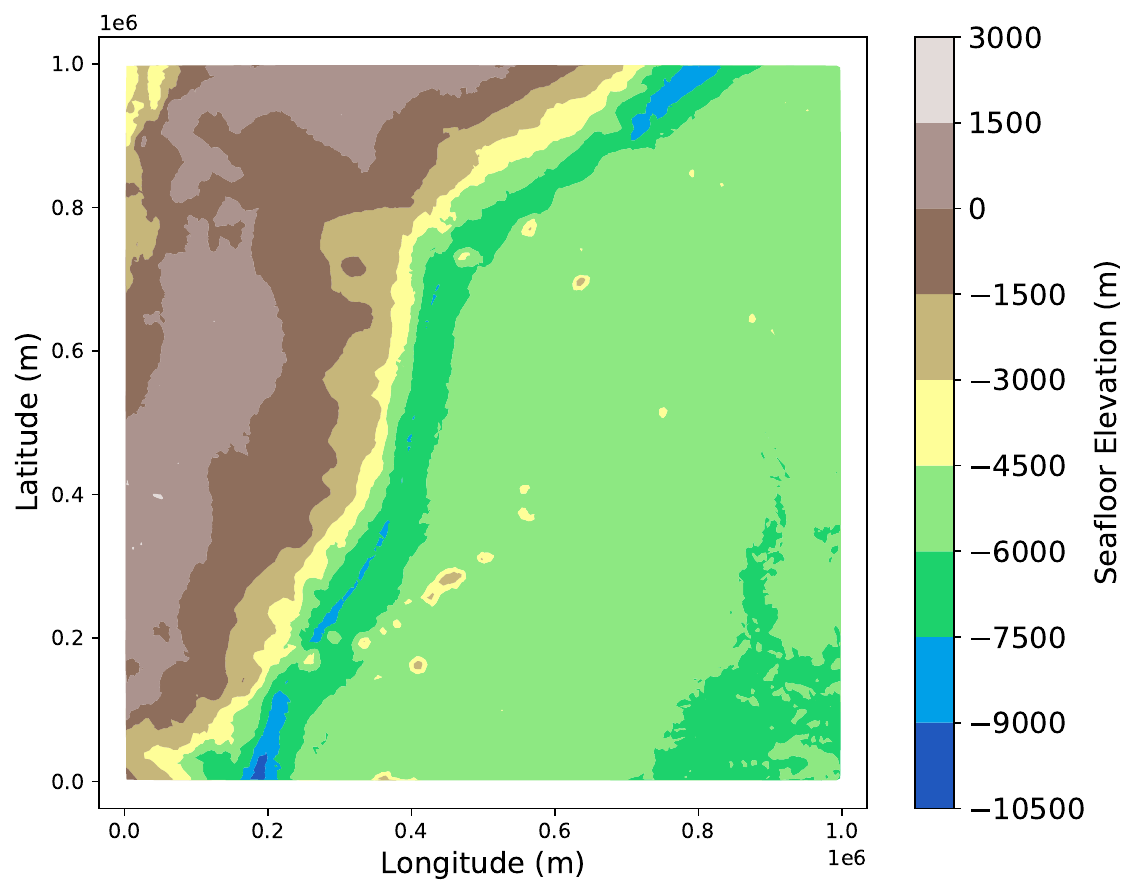} 
    \centering
    \caption{\scriptsize Bathymetry and Topography of the Computational Domain for the 2011 Tohoku Tsunami Simulations}
    \label{fig:bed_elevation}
\end{figure}

The Tohoku 2011 tsunami was triggered by an Mw 9.0 earthquake along the Japan Trench on March 11, 2011. To simulate this event, we investigate a 1000 km × 1000 km region as illustrated in Figure \ref{fig:bed_elevation}, using real bathymetry data to define the pre-earthquake seabed topography \( z_{b0}(\textbf{x}) \). This dataset, obtained from ETOPO 2022 at 30 Arc-Second resolution, provides an accurate baseline representation of the ocean floor before the earthquake.

As shown in Figure~\ref{fig:bed_elevation}, Bathymetry of the Computational Domain for the 2011 Tohoku Tsunami simulations. Before the earthquake, the water surface is assumed to be at rest, with an undisturbed sea level and no initial flow:

\begin{equation}
h(\textbf{x}, 0) = 0, \quad u(\textbf{x}, 0) = 0, \quad v(\textbf{x}, 0) = 0.
\label{eq:swe-initial-prequake}
\end{equation}
To model the seafloor deformation caused by the earthquake, we use the Okada displacement model \citep{okada1992internal}, which provides an analytical solution for surface deformation due to fault slip in an elastic half-space. This model computes the coseismic displacement \( \Delta z_b(\textbf{x}) \) based on key fault parameters that define the fault geometry and slip characteristics. The \textbf{depth} represents how deep the fault's top edge is below the surface, while the \textbf{length} and \textbf{width} describe the along-strike and down-dip dimensions of the fault plane. The \textbf{slip} quantifies the magnitude of displacement along the fault, whereas the \textbf{strike} specifies the fault's orientation relative to north, measured in degrees. The \textbf{dip} represents the inclination angle of the fault plane relative to a horizontal surface, while the \textbf{rake} defines the direction of slip along the fault plane. Together, these parameters determine the vertical displacement field, shaping the seafloor uplift or subsidence that drives tsunami generation. The earthquake's epicenter location is defined by its \textbf{\( x \)-coordinate} and \textbf{\( y \)-coordinate}, determining the origin of the deformation. Given these parameters, the Okada model computes the vertical displacement field \( \Delta z_b(x) \), representing the seafloor uplift or subsidence that alters the initial seabed topography. This deformation is incorporated into the model as:

\begin{equation}
z_b(\textbf{x}, 0) = z_{b0}(\textbf{x}) + \Delta z_b(\textbf{x}),
\end{equation}

where \( z_{b0}(\textbf{x}) \) represents the original seabed elevation before the earthquake, and \( \Delta z_b(\textbf{x}) \) accounts for the vertical displacement caused by the seismic event. This sudden deformation of the seabed displaces the overlying water column, generating the initial tsunami wave. The resulting modification of the water depth is expressed as:

\begin{equation}
h(\textbf{x}, 0) = \Delta z_b(\textbf{x}), \quad u(\textbf{x}, 0) = 0, \quad v(\textbf{x}, 0) = 0.
\label{eq:swe-initial}
\end{equation}

Thus, the initial tsunami wave is entirely defined by the seafloor deformation \( \Delta z_b(\textbf{x}) \), while the velocity field remains zero, assuming no initial horizontal flow. In this framework, the distributed coseismic deformation \( \Delta z_b(\textbf{x}) \) serves as the input parameter \( p \), which varies spatially over the domain \( \mathcal{X} \).


\subsection*{Finite Volume Method for Shallow Water Equations}
\label{sec:fv_sw}

We developed dTSUNAMI, a finite volume solver for the 2D shallow water equations (SWE), designed to efficiently handle tsunami propagation and inundation over complex bathymetry. The solver operates on an unstructured triangular mesh, ensuring mass conservation and allowing for local refinements in regions of interest, such as coastal areas and tsunami impact zones. The governing equations are expressed in their integral form over a control volume \( \Omega_i \):

\begin{equation}
\frac{d}{dt} \int_{\Omega_i} \mathbf{U} \, d\Omega
+ \oint_{\partial \Omega_i} \mathbf{F} \cdot \mathbf{n} \, dS
=
\int_{\Omega_i} \mathbf{S} \, d\Omega,
\end{equation}

where \( \mathbf{U} = (h, hu, hv)^T \) is the conserved state vector, \( \mathbf{F} \) is the flux tensor, \( \mathbf{n} \) is the outward unit normal at each face, and \( \mathbf{S} \) represents source terms.
To enhance accuracy in tsunami simulations, we developed a robust mesh generation approach within dTSUNAMI, leveraging GMSH to create an unstructured triangular mesh with adaptive local refinement.

Fluxes across cell faces are computed using a Riemann solver. We employ the HLLC (Harten-Lax-van Leer-Contact) scheme, which provides a balance between accuracy and robustness. The approximate Riemann solution at an interface is given by:

\begin{equation}
\mathbf{F}_{i+1/2} = 
\begin{cases}
\mathbf{F}_L, & \text{if } S_L \geq 0, \\[5pt]
\mathbf{F}^*, & \text{if } S_L < 0 < S_R, \\[5pt]
\mathbf{F}_R, & \text{if } S_R \leq 0,
\end{cases}
\end{equation}

where \( S_L \) and \( S_R \) are the wave speeds estimated from the characteristic structure of the system.

Time integration is performed using an adaptive second-order Heun’s method, an explicit two-stage Runge-Kutta scheme:

\begin{equation}
\mathbf{K}_1 = \mathbf{F}(t_n, \mathbf{U}_n),
\end{equation}

\begin{equation}
\widetilde{\mathbf{U}}_n = \mathbf{U}_n + \Delta t_n \mathbf{K}_1,
\end{equation}

\begin{equation}
\mathbf{K}_2 = \mathbf{F}(t_n + \Delta t_n, \widetilde{\mathbf{U}}_n),
\end{equation}

\begin{equation}
\mathbf{U}_{n+1} = \mathbf{U}_n + \frac{\Delta t_n}{2} (\mathbf{K}_1 + \mathbf{K}_2).
\end{equation}

The time step \( \Delta t_n \) is adaptively chosen based on the Courant-Friedrichs-Lewy (CFL) condition.


We validated dTSUNAMI against the 2011 Tohoku tsunami, triggered by the Mw 9.0 earthquake that occurred on March 11, 2011, at 05:46 UTC (14:46 JST), with its epicenter located at 37.490°\,N, 143.030°\,E. 

\begin{figure}
\centering
\includegraphics[width=0.9\textwidth]{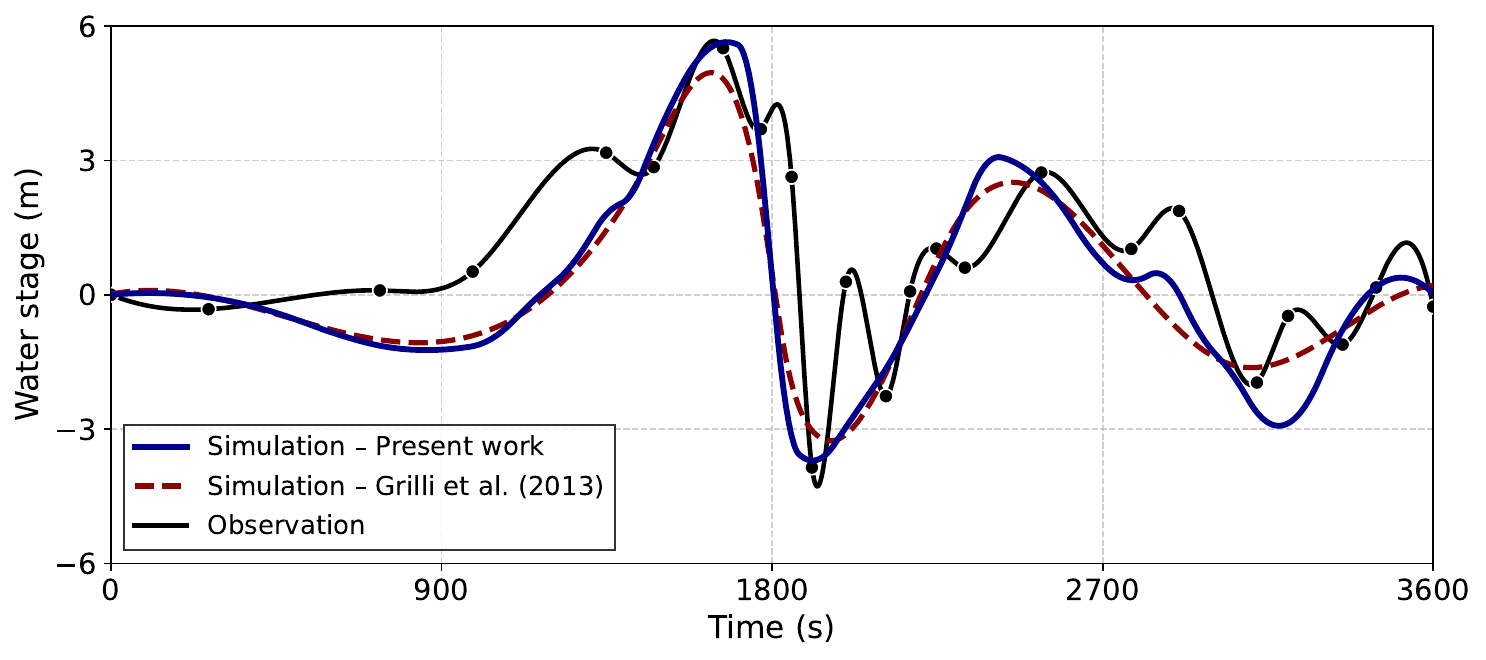}
\caption{\scriptsize Comparison of simulated free-surface elevation time series at North Miyagi offshore GPS wave gauge (station 803, NOWPHAS network) with recorded measurements during the 2011 Tohoku tsunami.}
\label{fig:tohoku_validation}
\end{figure}

The computational domain employs an unstructured triangular mesh consisting of 308,001 cells and 154,598 nodes. Figure~\ref{fig:tohoku_validation} shows time series of the simulated free-surface elevation compared to measurements recorded at the North Miyagi offshore GPS wave gauge (station 803) of the NOWPHAS network. For reference, the same figure includes the results reported by Grilli et al.~\citep{grilli2013numerical} using the Boussinesq model FUNWAVE-TVD.

The dTSUNAMI results exhibit very good agreement with both the observed data and the reference simulation, accurately capturing the arrival time, amplitude of the leading crest, and the subsequent wave train. This close match clearly demonstrates the validity and high accuracy of the solver for real-world large-scale tsunami propagation simulations.

\subsection*{Adjoint-Based Sensitivity Computation}
\label{sec:adjoint_sensitivity}

To quantify the sensitivity of the final water depth \( h_N \) to variations in bed topography \( b \), we employ the adjoint method. This approach efficiently computes gradients without requiring finite-difference perturbations, making it well-suited for high-dimensional tsunami simulations.

\subsubsection*{Adjoint System Formulation}

The forward SWE system evolves as:
\begin{equation}
\mathbf{U}_{n+1} = \mathbf{U}_n + \frac{\Delta t_n}{2} \left( \mathbf{K}_1^n + \mathbf{K}_2^n \right),
\end{equation}
where
\begin{equation}
\mathbf{K}_1^n = \mathbf{F}(\mathbf{U}_n, \mathbf{b}), \quad
\mathbf{K}_2^n = \mathbf{F}(\widetilde{\mathbf{U}}_n, \mathbf{b}),
\end{equation}
and the final objective function is:
\begin{equation}
J = \mathbf{h}_N = \mathbf{P} \mathbf{U}_N,
\end{equation}
where \( \mathbf{P} \) extracts water depth \( h \) from \( \mathbf{U} \).

The Lagrangian function incorporates the constraints imposed by the forward dynamics:
\begin{align}
\mathcal{L} = J + \sum_{n=0}^{N-1} \bigg[
\boldsymbol{\lambda}_{1,n}^T \left( \mathbf{K}_1^n - \mathbf{F}(\mathbf{U}_n, \mathbf{b}) \right)
+ \boldsymbol{\lambda}_{2,n}^T \left( \widetilde{\mathbf{U}}_n - \mathbf{U}_n - \Delta t_n \mathbf{K}_1^n \right)  \notag \\
+ \boldsymbol{\lambda}_{3,n}^T \left( \mathbf{K}_2^n - \mathbf{F}(\widetilde{\mathbf{U}}_n, \mathbf{b}) \right)
+ \boldsymbol{\lambda}_{4,n}^T \left( \mathbf{U}_{n+1} - \mathbf{U}_n - \frac{\Delta t_n}{2} (\mathbf{K}_1^n + \mathbf{K}_2^n) \right) \bigg].
\end{align}

\subsubsection*{Backward Recursive Computation of Adjoint Variables}

The adjoint system is solved by backward recursion, starting from the final time step.

\textbf{Initialization at Final Time Step:}
\begin{equation}
\boldsymbol{\lambda}_{4,N-1} = -\mathbf{P}^T.
\end{equation}

\textbf{Recursive Computation for \( n = N-1, N-2, \dots, 0 \):}
\begin{enumerate}
    \item Compute adjoint correction for \( \mathbf{K}_2^n \):
    \begin{equation}
    \boldsymbol{\lambda}_{3,n} = \frac{\Delta t_n}{2} \boldsymbol{\lambda}_{4,n}.
    \end{equation}

    \item Compute adjoint variable for the intermediate state \( \widetilde{\mathbf{U}}_n \):
    \begin{equation}
    \boldsymbol{\lambda}_{2,n} = \boldsymbol{\lambda}_{3,n}^T \frac{\partial \mathbf{F}}{\partial \mathbf{U}} \Big|_{\widetilde{\mathbf{U}}_n}.
    \end{equation}

    \item Compute adjoint correction for \( \mathbf{K}_1^n \):
    \begin{equation}
    \boldsymbol{\lambda}_{1,n} = \Delta t_n \boldsymbol{\lambda}_{2,n} + \frac{\Delta t_n}{2} \boldsymbol{\lambda}_{4,n}.
    \end{equation}

    \item Update adjoint state for the next iteration:
    \begin{equation}
    \boldsymbol{\lambda}_{4,n-1} = \boldsymbol{\lambda}_{1,n} \frac{\partial \mathbf{F}}{\partial \mathbf{U}} \Big|_{\mathbf{U}_n} + \boldsymbol{\lambda}_{2,n} + \boldsymbol{\lambda}_{4,n}.
    \end{equation}
\end{enumerate}
The recursive computation of \(\boldsymbol{\lambda}\) proceeds from the final time step \(N\) back to the initial time step \(0\). The pseudocode for efficient implementation of the recursion is presented in Algorithm~\ref{alg:adjoint}.

\begin{algorithm}[H]
\caption{Recursive Computation of \(\boldsymbol{\lambda}\)}
\label{alg:adjoint}
\begin{algorithmic}[1]
    \State \textbf{Input:} 
    \begin{itemize}
        \item Time steps \(\Delta t_n\)
        \item Jacobians \(\frac{\partial \mathbf{F}}{\partial \mathbf{b}}\big|_{\mathbf{U}_n}\), \(\frac{\partial \mathbf{F}}{\partial \mathbf{b}}\big|_{\widetilde{\mathbf{U}}_n}\)
        \item Output matrix \(\mathbf{P}\)
    \end{itemize}
    \State \textbf{Initialize:} 
    \[
    \boldsymbol{\lambda}_{4,N-1} \gets -\mathbf{P}^T
    \]
    \For{\(n = N-1, N-2, \dots, 0\)}
        \State \textbf{Step 1: Compute \(\boldsymbol{\lambda}_{3,n}\)}
        \[
        \boldsymbol{\lambda}_{3,n} \gets \frac{\Delta t_n}{2} \boldsymbol{\lambda}_{4,n}
        \]

        \State \textbf{Step 2: Compute \(\boldsymbol{\lambda}_{2,n}\)}
        \[
        \boldsymbol{\lambda}_{2,n} \gets \boldsymbol{\lambda}_{3,n}^T \frac{\partial \mathbf{F}}{\partial \mathbf{U}} \big|_{\widetilde{\mathbf{U}}_n}
        \]

        \State \textbf{Step 3: Compute \(\boldsymbol{\lambda}_{1,n}\)}
        \[
        \boldsymbol{\lambda}_{1,n} \gets \Delta t_n \boldsymbol{\lambda}_{2,n} + \frac{\Delta t_n}{2} \boldsymbol{\lambda}_{4,n}
        \]

        \State \textbf{Step 4: Update \(\boldsymbol{\lambda}_{4,n-1}\)}
        \[
        \boldsymbol{\lambda}_{4,n-1} \gets \boldsymbol{\lambda}_{1,n} \frac{\partial \mathbf{F}}{\partial \mathbf{U}} \big|_{\mathbf{U}_n} + \boldsymbol{\lambda}_{2,n} + \boldsymbol{\lambda}_{4,n}
        \]

        \State \textbf{Step 5: Store Intermediate Values:}
        \begin{itemize}
            \item Save \(\boldsymbol{\lambda}_{1,n}\) for sensitivity computation
            \item Save \(\boldsymbol{\lambda}_{3,n}\) for sensitivity computation
        \end{itemize}
    \EndFor
\end{algorithmic}
\end{algorithm}

The final sensitivity of the water depth at \( t = T \) with respect to bed topography is obtained as:

\begin{equation}
\frac{\partial J}{\partial b} = \sum_{n=0}^{N-1} \left[
-\boldsymbol{\lambda}_{1,n}^T \frac{\partial \mathbf{F}}{\partial b} \Big|_{\mathbf{U}_n}
- \boldsymbol{\lambda}_{3,n}^T \frac{\partial \mathbf{F}}{\partial b} \Big|_{\widetilde{\mathbf{U}}_n}
\right].
\end{equation}

Since \( J = \mathbf{h}_N = \mathbf{P} \mathbf{U}_N \), we obtain:

\begin{equation}
\frac{\partial h_N}{\partial b} = \sum_{n=0}^{N-1} \left[
-\boldsymbol{\lambda}_{1,n}^T \frac{\partial \mathbf{F}}{\partial b} \Big|_{\mathbf{U}_n}
- \boldsymbol{\lambda}_{3,n}^T \frac{\partial \mathbf{F}}{\partial b} \Big|_{\widetilde{\mathbf{U}}_n}
\right].
\end{equation}

This Jacobian \( \frac{\partial h_N}{\partial b} \in \mathbb{R}^{M \times M} \) quantifies how changes in bed topography influence final water depth, crucial for tsunami inversion and forecasting.

\subsubsection*{Derivation of $\frac{\partial \mathbf{F}}{\partial \mathbf{U}}$ for FVM}
To compute the adjoint sensitivity, we derive the local Jacobian \( \frac{\partial \mathbf{F}}{\partial \mathbf{U}} \) for the finite volume discretization. Differentiating the numerical flux at a face shared between cells \( i \) and \( j \):

\begin{equation}
\frac{\partial \mathbf{F}_f}{\partial \mathbf{U}_i} =
\begin{pmatrix}
\mathbf{u} \cdot \mathbf{n} & n_x & n_y \\
u (\mathbf{u} \cdot \mathbf{n}) + g h n_x & 2 u n_x & u n_y \\
v (\mathbf{u} \cdot \mathbf{n}) + g h n_y & v n_x & 2 v n_y
\end{pmatrix}.
\end{equation}

For source terms including bed slope and friction:
\begin{equation}
\frac{\partial \mathbf{S}_i}{\partial \mathbf{U}_i} =
\begin{pmatrix}
0 & 0 & 0 \\
- g \frac{\partial b}{\partial x} - c_f u & -c_f |\mathbf{u}| & 0 \\
- g \frac{\partial b}{\partial y} - c_f v & 0 & -c_f |\mathbf{u}|
\end{pmatrix},
\end{equation}
where \( c_f = g n^2 / h^{4/3} \) is the friction coefficient.

\paragraph{Jacobian Assembly}
The full local Jacobian is assembled as:
\begin{equation}
\frac{\partial \mathbf{F}_i}{\partial \mathbf{U}_i} = -\frac{1}{A_i} \sum_{f \in \text{faces}(i)} l_f \frac{\partial \mathbf{F}_f}{\partial \mathbf{U}_i} + \frac{\partial \mathbf{S}_i}{\partial \mathbf{U}_i}.
\end{equation}
For neighboring cell interactions:
\begin{equation}
\frac{\partial \mathbf{F}_i}{\partial \mathbf{U}_j} = \frac{1}{A_i} l_f \frac{\partial \mathbf{F}_f}{\partial \mathbf{U}_j}, \quad j \in \text{neighbors}(i).
\end{equation}

This formulation ensures consistency with the adjoint framework, capturing local sensitivities required for gradient-based inversion and optimization.

\subsubsection*{Derivation of $\frac{\partial \mathbf{F}}{\partial b}$ for FVM}
To compute the local sensitivity of the shallow water system to bed topography \( b \), we differentiate the right-hand side function \( \mathbf{F} \).

\paragraph{Flux Contributions}
The governing equations include:
\begin{align}
\frac{\partial \mathbf{F}_1}{\partial b} &= 0, \\[5pt]
\frac{\partial \mathbf{F}_2}{\partial b} &= -gh \frac{\partial b}{\partial x}, \\[5pt]
\frac{\partial \mathbf{F}_3}{\partial b} &= -gh \frac{\partial b}{\partial y}.
\end{align}
Only the momentum equations contain explicit \( b \)-dependence via the bed slope terms.

For a domain with \( M \) cells, the Jacobian matrix is:
\begin{equation}
\frac{\partial \mathbf{F}}{\partial b} =
\begin{pmatrix}
\mathbf{0}_{M \times M} \\[5pt]
- g \text{diag}(h) \cdot \text{diag}(S_x) \\[5pt]
- g \text{diag}(h) \cdot \text{diag}(S_y)
\end{pmatrix} \in \mathbb{R}^{3M \times M}.
\end{equation}

In this formulation, the first \( M \) rows correspond to continuity equations with no explicit \( b \)-dependence. The middle and last \( M \) rows represent the effects of bed slope on x- and y-momentum equations, respectively. The diagonal form ensures each grid cell's sensitivity is localized, aligning with the adjoint system.

\subsection*{Neural Operator Learning for the Shallow Water Equations}

We employ neural operator learning to model tsunami dynamics governed by the shallow water equations with Manning friction, applied to the 2011 Tohoku event setup. Variability across samples arises from earthquake-induced seafloor deformation, represented through the Okada dislocation framework. Okada fault parameters are sampled within the ranges reported in \citep{grilli2013numerical}, and random earthquake epicenters are assigned as illustrated in Supplementary Fig.~\ref{fig:nse_sampling_strategies}.  

To generate training data, we used the dTSUNAMI finite volume solver on an unstructured triangular mesh consisting of 308,001 cells and 154,598 nodes. Each simulation is run for 3600 seconds, storing 51 uniformly spaced snapshots. These high-resolution solutions and their Jacobians are subsequently coarsened to a 100×100 uniform grid for neural operator training.

Within the SC-NO framework, the input function consists of the first $t_u = 5$ time steps of the water stage field $h(\mathbf{x}, t)$ together with the seafloor deformation $\Delta z_b(\mathbf{x})$ computed from the Okada model. The neural operator $\mathcal{G}_\theta$ is then trained to predict the remaining $t_G = 46$ time steps, guided by the Jacobian of the final water stage with respect to the bed topography, computed via the adjoint method (Supplementary Section~\ref{sec:adjoint_sensitivity}). Formally, the operator learning problem is defined as:
\begin{equation}
\mathcal{G}_\theta : \left( \mathcal{X} \times [0, t_u], \, \Delta z_b(\mathbf{x}) \right) \;\to\; h(\mathbf{x}, t, \Delta z_b), 
\quad \text{for } (\mathbf{x}, t) \in \mathcal{X} \times [t_u, T].
\end{equation}

To evaluate both interpolation and extrapolation performance, we consider two testing scenarios: (i) an \textit{in-training} case, where earthquake epicenters fall within the range covered by the training set, and (ii) an \textit{out-of-training} case, where epicenters lie outside this region (Supplementary Fig.~\ref{fig:nse_sampling_strategies}). This setup enables us to assess not only the accuracy of SC-NO within the training distribution but also its ability to generalize tsunami propagation to previously unseen earthquake configurations, serving as an out-of-distribution (OOD) evaluation.

This approach integrates spatial-temporal input information with earthquake source parameters, enabling SC-NO to learn a solution operator that generalizes tsunami propagation and inundation across both in-training and out-of-distribution epicenter scenarios.

\begin{figure}
    \centering
    \includegraphics[width=0.8\textwidth]{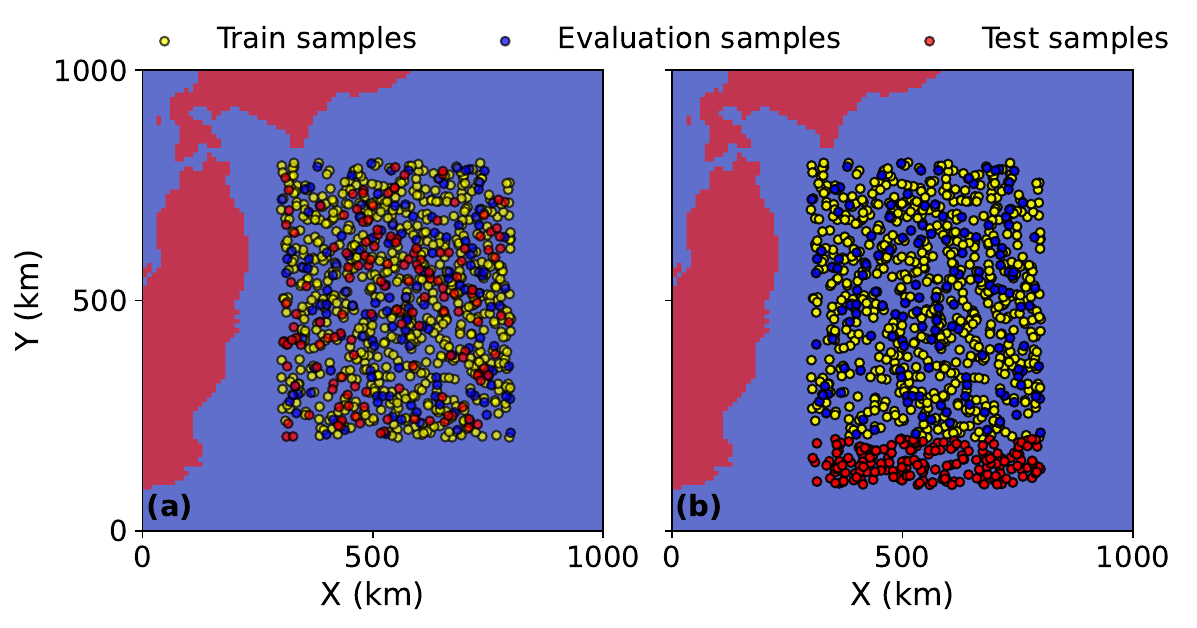}

    \caption{\scriptsize\textbf{Sampling strategy for PDE3 (tsunami scenario).} 
    (a) Training and evaluation samples are generated from random earthquake epicenters distributed across the central region of the domain. 
    (b) Out-of-distribution (OOD) test samples are generated by shifting epicenters into a geographically distinct southern band, ensuring that all test events occur in a region not represented in the training or evaluation sets.}     
        
        \label{fig:nse_sampling_strategies}
\end{figure}

\subsection*{Mapping Between Fine Solver Mesh and Coarse Neural Operator Grid}
\label{sec:coarse_fine_mapping}

For the shallow-water tsunami case (PDE3), neural operator training and inference are conducted on a fixed $100\times100$ uniform grid, while the governing dynamics are solved on a substantially finer unstructured mesh using the dTSUNAMI finite volume solver. This resolution disparity requires two complementary mappings: a \emph{coarsening} operator that restricts high-resolution solutions to the learning grid, and an \emph{upscaling} operator that reconstructs physically consistent fine-scale fields from coarse neural-operator predictions.

\paragraph{Coarsening of fine-scale solutions.}
Let $d(\mathbf{x}, t)$ denote the fine-scale water depth field produced by the solver on the continuous domain $\mathcal{X}$. We introduce a uniform partition $\{\Omega_{ij}\}$ of $\mathcal{X}$ corresponding to the $100\times100$ neural operator grid. Coarsening is performed via a cell-averaging operator,
\begin{equation}
\bar{d}_{ij}(t)
=
\frac{1}{|\Omega_{ij}|}
\int_{\Omega_{ij}} d(\mathbf{x}, t)\, d\mathbf{x},
\end{equation}
yielding a coarse depth field $\bar{d}_{ij}(t)$ that preserves the mean water depth within each cell. In practice, this corresponds to averaging solver values associated with fine mesh elements whose centroids lie within $\Omega_{ij}$. This projection provides a physically consistent low-resolution representation suitable for neural operator learning while retaining the dominant large-scale dynamics.

\paragraph{Physically based upscaling from coarse to fine resolution.}
The inverse mapping from coarse mean depth to fine-scale depth is not uniquely defined: many fine configurations yield the same coarse average. To obtain a physically meaningful reconstruction, the upscaling is defined as a \emph{cell-wise hydrostatic fill} over the known fine bathymetry.

Let $\{\Omega_{ij}\}$ denote the $100\times 100$ coarse partition of the horizontal domain, and let $\{(\mathbf{x}_p,z_{b,p})\}_{p=1}^{N}$ be the fine solver nodes (or vertices), where $\mathbf{x}_p\in\mathbb{R}^2$ and $z_{b,p}=z_b(\mathbf{x}_p)$ is the bed elevation. Each fine node is assigned to exactly one coarse cell via
\begin{equation}
c(p)= (i,j)\quad \text{s.t.}\quad \mathbf{x}_p\in\Omega_{ij}, 
\qquad 
\mathcal{P}_{ij}=\{p:\,c(p)=(i,j)\},\quad N_{ij}=|\mathcal{P}_{ij}|.
\end{equation}
The neural operator predicts the coarse mean depth $\bar d_{ij}(t)$ on each cell. The reconstructed fine depth is defined pointwise by
\begin{equation}
d_p(t) \;=\; \big(H_{ij}(t)-z_{b,p}\big)_+,
\qquad p\in\mathcal{P}_{ij},
\qquad (a)_+ := \max(a,0),
\label{eq:bathtub_pointwise}
\end{equation}
where $H_{ij}(t)$ is a \emph{cell-wise water level} (stage) chosen to satisfy the coarse mean-depth constraint in \emph{discrete} form:
\begin{equation}
\frac{1}{N_{ij}}\sum_{p\in\mathcal{P}_{ij}} \big(H_{ij}(t)-z_{b,p}\big)_+ \;=\; \bar d_{ij}(t).
\label{eq:bathtub_mean_constraint}
\end{equation}
Equation~\eqref{eq:bathtub_mean_constraint} defines $H_{ij}(t)$ implicitly. The left-hand side is a continuous, non-decreasing function of $H_{ij}(t)$, hence the solution exists and is unique whenever $\bar d_{ij}(t)$ is attainable over the local bathymetric distribution $\{z_{b,p}\}_{p\in\mathcal{P}_{ij}}$. In practice, $H_{ij}(t)$ is obtained by bisection with brackets
\begin{equation}
H_{ij}^{\mathrm{low}} = \min_{p\in\mathcal{P}_{ij}} z_{b,p},
\qquad
H_{ij}^{\mathrm{high}} = \max_{p\in\mathcal{P}_{ij}} z_{b,p} + \bar d_{ij}(t) + \varepsilon,
\end{equation}
and the cell is treated as dry whenever $\bar d_{ij}(t)\le d_{\min}$, in which case $d_p(t)=0$ for all $p\in\mathcal{P}_{ij}$.

Finally, to ensure that the reconstruction preserves the \emph{total} water volume implied by the coarse prediction at each time $t$, an optional per-time-step global rescaling is applied. Let $A_{ij}=|\Omega_{ij}|$ be the coarse cell area and approximate each node in $\Omega_{ij}$ by an equal area weight $A_{ij}/N_{ij}$. Then the target and reconstructed volumes are
\begin{equation}
V_{\mathrm{target}}(t)=\sum_{i,j} \bar d_{ij}(t)\,A_{ij},
\qquad
V_{\mathrm{rec}}(t)=\sum_{i,j}\sum_{p\in\mathcal{P}_{ij}} d_p(t)\,\frac{A_{ij}}{N_{ij}},
\end{equation}
and we set a scalar factor $\alpha(t)=V_{\mathrm{target}}(t)/V_{\mathrm{rec}}(t)$ (when $V_{\mathrm{rec}}(t)>0$) and replace $d_p(t)\leftarrow \alpha(t)\,d_p(t)$.
This upscaling is therefore non-negative, bathymetry-aware, and exactly consistent with the coarse mean-depth constraints by construction, while avoiding the unphysical shoreline behavior that can arise from purely geometric interpolation in shallow regions.

\begin{figure}
    \centering
    \includegraphics[width=0.8\textwidth]{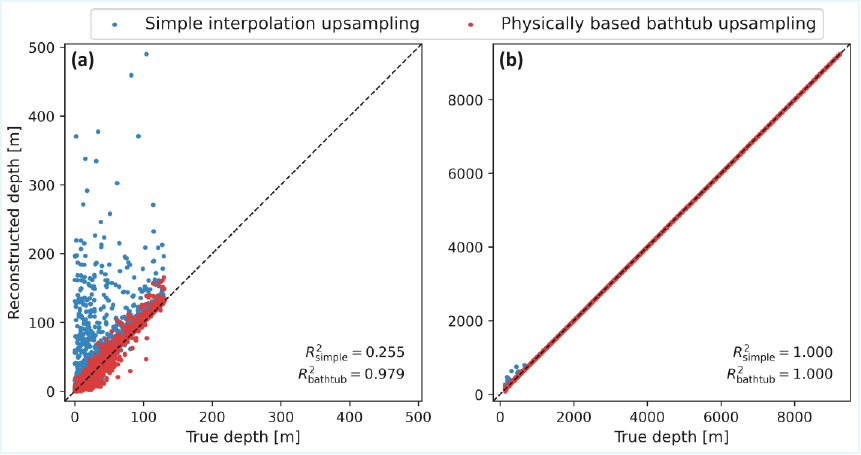}
    \caption{\scriptsize\textbf{Comparison of depth upscaling methods.}
    (a) Shallow-water regime (0--5th percentile of true depth).
    (b) Moderate-to-deep regime (5--100th percentile of true depth).}
    \label{fig:upsampling_comparison}
\end{figure}

A common alternative is to upsample coarse predictions using standard interpolation methods (e.g., nearest-neighbor or bilinear interpolation of depth or stage). Such approaches ignore bathymetric variability and do not enforce physical constraints, often leading to spurious inundation over dry land and degraded accuracy in shallow regions.

Figure~\ref{fig:upsampling_comparison} illustrates a controlled comparison between a standard interpolation-based reconstruction and the proposed physically based upscaling procedure. Starting from a high-resolution reference solution obtained from the tsunami solver, the water depth field is first \emph{coarsened} to the $100\times100$ uniform grid used by the neural operator. This coarse representation is then independently upscaled back to the original fine resolution using (i) a purely geometric interpolation scheme and (ii) the proposed bathtub-based reconstruction. The resulting reconstructed fields are compared against the original high-resolution solution.

In shallow-water regions near the shoreline (0–5th percentile of true depth), interpolation-based reconstruction fails to recover physically meaningful depths and exhibits poor agreement with the reference solution (Figure~\ref{fig:upsampling_comparison}a). This failure arises because interpolation does not account for local bathymetric structure and cannot enforce essential physical constraints such as non-negativity of depth or shoreline emergence. In contrast, the bathtub-based reconstruction explicitly incorporates the underlying bed elevation and enforces consistency with the coarse mean depth, yielding accurate recovery of shallow-water states.

In moderate-to-deep regions, both reconstruction approaches perform comparably well, indicating that simple interpolation is sufficient only away from the shoreline (Figure~\ref{fig:upsampling_comparison}b). These results demonstrate that physically informed upscaling is a necessary complement to coarse neural-operator predictions when reconstructing near-shore and inundation-relevant tsunami dynamics, while purely geometric interpolation is inadequate in depth regimes most critical for hazard assessment.
Taken together, this comparison establishes the upscaling procedure as a reliable and physically consistent complement to the neural-operator framework. While SC-NO is trained and evaluated entirely on coarse-resolution fields, the proposed reconstruction enables faithful recovery of fine-scale depth distributions required for solver-level validation, visualization, and downstream hazard analysis. Importantly, this post-processing step operates independently of the learned operator and does not modify its predictions; instead, it provides a principled bridge between coarse neural-operator outputs and high-resolution physical fields, ensuring consistency with bathymetry and shallow-water physics.

\subsection*{Three-Stage Inversion Framework for Tsunami Source Reconstruction}
\label{sec:inversion_framework}

\begin{figure}
    \centering
    \includegraphics[width=0.9\textwidth, trim={0 0 0 0}, clip]{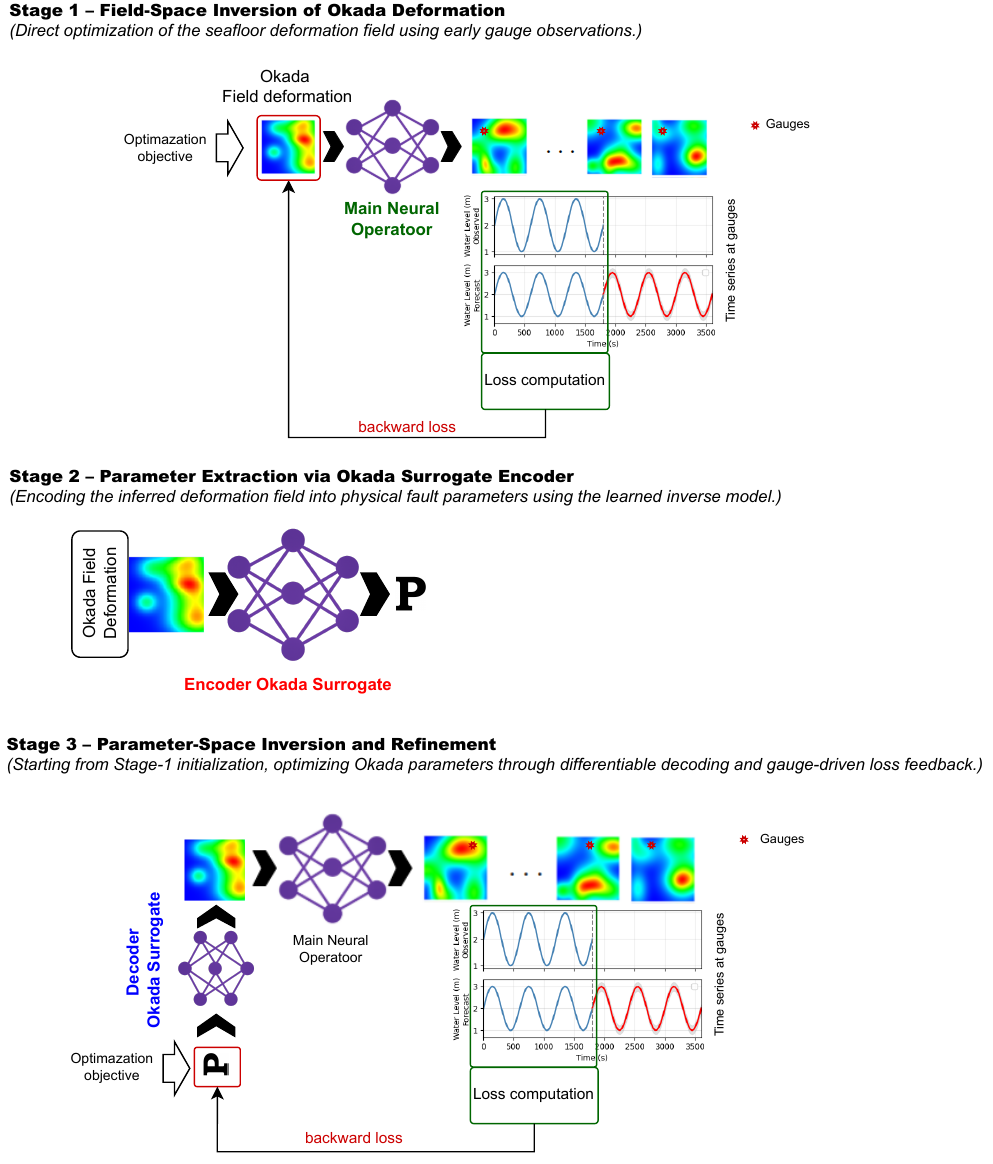}
    \caption{\scriptsize\textbf{Three-stage tsunami inversion workflow.}  
    The framework integrates neural operator surrogates and Okada-based encoder–decoder networks for rapid and interpretable tsunami source reconstruction.  
    \textbf{Stage~1:} Field-space inversion optimizes the Okada deformation field using early tsunami gauge observations via neural operator surrogates (FNO or SC-FNO).  
    \textbf{Stage~2:} The inferred deformation is passed through a U-FNO--based Okada Surrogate Encoder that maps the 2D field to a nine-dimensional fault parameter vector $\mathbf{P}$.  
    \textbf{Stage~3:} A U-FNO Decoder reconstructs the deformation field from $\mathbf{P}$, enabling parameter-space refinement through differentiable feedback and gauge-driven loss propagation.
    }
    
    \label{fig:inversion_stages}
\end{figure}

Rapid and reliable estimation of tsunami sources is vital for the effectiveness of real-time early-warning systems. When a major undersea earthquake occurs, the resulting seafloor deformation initiates tsunami waves that can traverse ocean basins and reach coastal regions within minutes, demanding immediate and accurate source characterization.  
To meet this need, we introduce a \textbf{data-driven, gradient-based inversion framework} that integrates \textbf{Neural Operators} with \textbf{Okada surrogate networks} to reconstruct the earthquake-induced seafloor deformation and infer the corresponding fault parameters directly from \textbf{early-stage tsunami gauge observations}.  
These early observations, typically spanning the first 30 minutes after the event onset, contain limited yet highly informative signals that guide the inversion toward physically consistent, interpretable, and computationally efficient source reconstructions suitable for real-time forecasting.
The framework combines the differentiability and efficiency of neural operators with the interpretability of Okada fault parameterization, enabling near--real-time source reconstruction with both accuracy and physical coherence. The inversion proceeds in three sequential stages, progressively transitioning from field-level reconstruction to interpretable parameter refinement (Figure~\ref{fig:inversion_stages}).

\begin{figure}]
    \centering
    \includegraphics[width=\textwidth, trim={0 0 0 0}, clip]{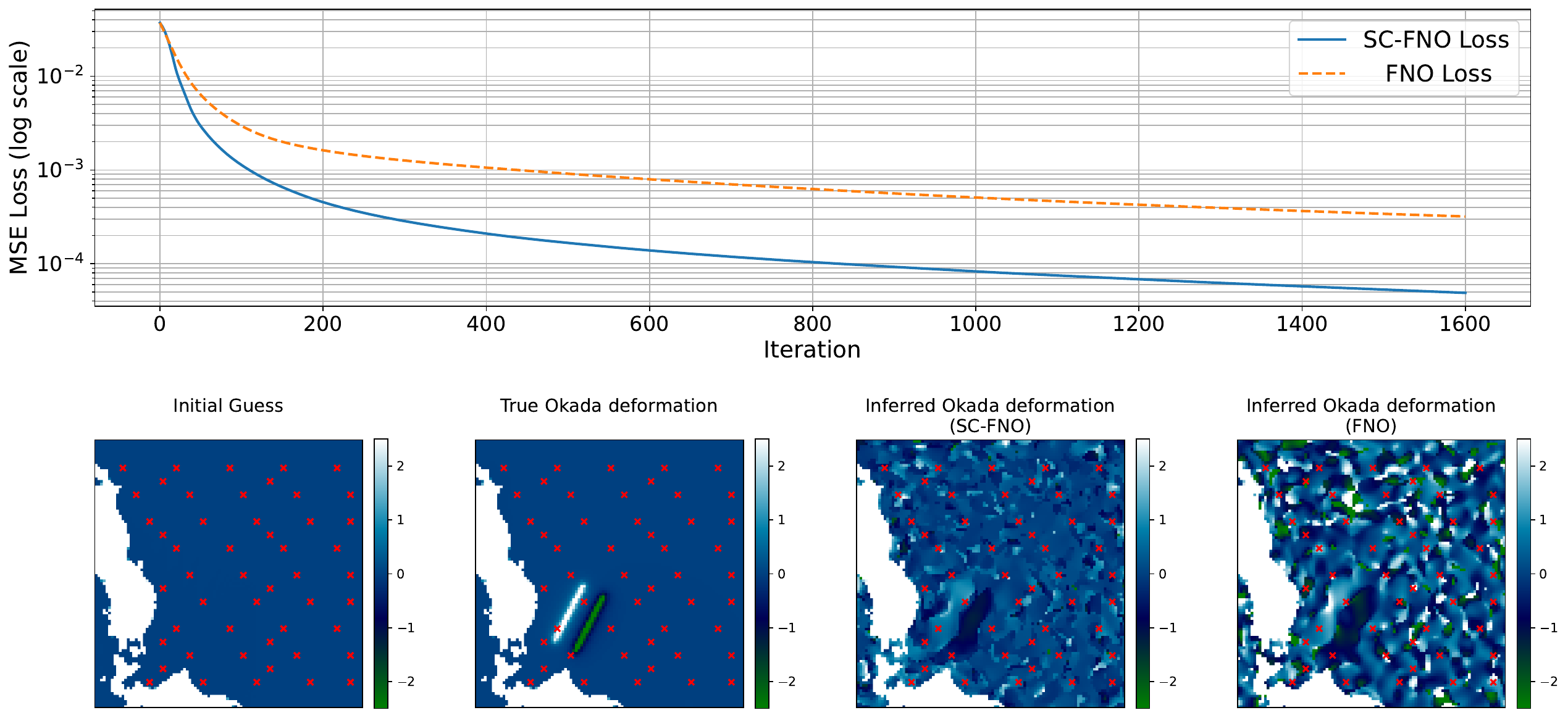}
    \caption{\scriptsize\textbf{Field-space inversion of Okada deformation (Stage~1).}  
    Evolution of the mean squared error (MSE) loss during inversion using SC-FNO and FNO (top), and comparison of deformation fields (bottom).  
    Red dots indicate the locations of tsunami gauges.}
        \label{fig:stage1_field_inversion}
\end{figure}

\paragraph{Stage 1 --- Field-Space Inversion of Okada Deformation}
\textit{(Direct gradient-based optimization of the seafloor deformation field using early gauge observations.)}

In the first stage, inversion is performed directly in the \textbf{field domain} to reconstruct the Okada deformation field---the vertical displacement of the seafloor responsible for tsunami initiation.  
Starting from an initial guess, the deformation field is iteratively updated through gradient descent to minimize the mean squared error between simulated and observed water-stage time series at selected gauge locations during the early event window ($T_u = 30$ minutes).  

Forward simulations are produced by a pretrained Neural Operator---either the baseline FNO or the Sensitivity-Constrained FNO ---which acts as a differentiable surrogate for the shallow water equations.  
This stage yields an initial deformation estimate that reproduces the observed signals but may contain \textbf{low-frequency artifacts} or \textbf{nonphysical features}, since the optimization acts directly on spatial pixels without explicit parameter constraints. At the end of Stage~1, both models converge toward deformation fields that reproduce the early tsunami gauge observations with reasonable accuracy (Figure~\ref{fig:stage1_field_inversion}). Although residual discrepancies remain in both reconstructions, the \textbf{SC-FNO} demonstrates faster convergence, smoother gradients, and more accurate localization of the primary rupture zone compared to the baseline \textbf{FNO}. The SC-FNO–inferred field better captures the overall geometry, polarity, and region of occurrence of the true Okada deformation, while the FNO result remains affected by high-frequency noise and scattered artifacts.

\paragraph{Stage 2 --- Parameter Extraction via Okada Surrogate Encoder}
\textit{(Encoding the inferred deformation field into compact Okada fault parameters through a learned inverse mapping.)}

To restore physical interpretability, the deformation field recovered in Stage~1, denoted as $\mathbf{F} \in \mathbb{R}^{H \times W}$, is passed through a pretrained \textbf{Okada Surrogate Encoder} $E_{\theta}$.  
This network performs the nonlinear mapping  
\[
E_{\theta} : \mathbf{F} \; \rightarrow \; \mathbf{P} \in \mathbb{R}^{9},
\]
where $\mathbf{P}$ represents the compact vector of Okada fault parameters, including the fault epicenter coordinates ($x$, $y$), depth, length, width, strike, dip, rake, and slip magnitude.

\begin{figure}
    \centering
    \includegraphics[width=0.98\textwidth, trim={0 0 0 0}, clip]{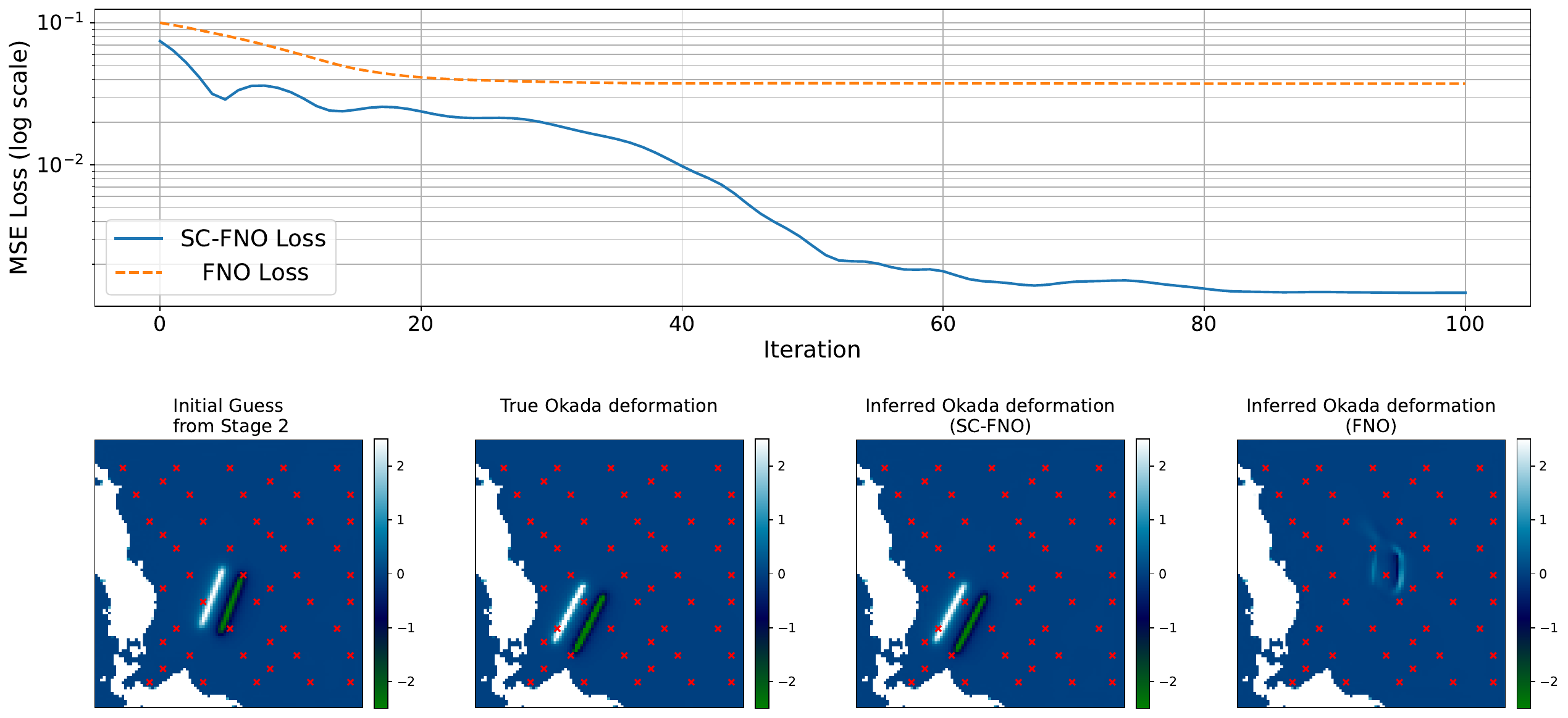}
    \caption{\scriptsize\textbf{Parameter-space inversion and refinement (Stage~3).}  
    Convergence of the mean squared error (MSE) loss during parameter-space optimization (top) and comparison of deformation fields (bottom).  
    The optimization is initialized using parameters inferred from Stage~2 and refined through the differentiable Decoder Okada Surrogate.  
    The sensitivity-constrained neural operator (SC-FNO) achieves faster convergence and higher reconstruction fidelity compared to the baseline FNO, closely reproducing the true Okada deformation.  
    Red dots indicate tsunami gauge locations.}
    \label{fig:stage3_param_inversion}
\end{figure}

\paragraph{Stage 3 --- Parameter-Space Inversion and Refinement via Decoder Okada Surrogate}
\textit{(Gradient-based optimization of Okada parameters using differentiable decoding and gauge-driven feedback.)}

In the final stage, inversion operates in the \textbf{parameter domain}. The vector $\mathbf{P}$ obtained from Stage~2 serves as the initial guess for a new optimization loop. A pretrained \textbf{Decoder Okada Surrogate} maps $\mathbf{P}$ back into its corresponding seafloor deformation field, providing a differentiable link between the parameters and simulated tsunami signals.  

At each iteration, the decoded deformation field is propagated through the main neural operator (FNO or SC-FNO) to generate gauge predictions. The resulting loss between predicted and observed signals is backpropagated through the full pipeline---decoder and forward model included---to update all nine Okada parameters simultaneously via gradient descent.  
This stage refines the parameter estimates within a physically meaningful subspace, producing deformation fields that are dynamically consistent with observed data.

At the end of Stage~3, the inversion converges toward a stable and physically consistent solution (Figure~\ref{fig:stage3_param_inversion}).
While the baseline \textbf{FNO} fails to achieve meaningful refinement, the \textbf{SC-FNO} continues to reduce the residual mismatch between predicted and observed gauge signals, converging toward the true Okada deformation. The resulting field accurately recovers both the location and polarity of the rupture with minimal spatial artifacts, confirming that sensitivity-constrained learning enables effective parameter-space optimization and physically reliable reconstruction.

With the final Okada deformation field inferred from Stage~3, the complete tsunami evolution can now be simulated through the trained neural operator model. By propagating the recovered deformation through the \textbf{FNO} and  \textbf{SC-FNO} forward models, we reconstruct the full spatiotemporal dynamics of the event from the initial seafloor motion to the final coastal impact.
Representative tsunami gauge time series are shown in Figure~\ref{fig:gauge_timeseries}, demonstrating that while both models reproduce the overall waveform trends, the \textbf{SC-FNO} achieves closer agreement with observed amplitudes and phases, maintaining stability and reducing residual errors throughout the simulation.


\clearpage
\subsection*{Additional results for PDE3}

\begin{table}[htbp]
\centering
\caption{\scriptsize\textbf{Comparison of Error Metrics Across Models Trained with Different Sample Sizes (in-training) PDE3}}
\label{tab:pde3_forward_nostd}
\scriptsize
\setlength{\tabcolsep}{3pt}
\begin{tabular}{lcccccccc}
\hline
\textbf{Model} & \multicolumn{2}{c}{\textbf{1200 Samples}} & \multicolumn{2}{c}{\textbf{1000 Samples}} & \multicolumn{2}{c}{\textbf{700 Samples}} & \multicolumn{2}{c}{\textbf{200 Samples}} \\
 & \textbf{Rel. \(L^2\)} & \textbf{MAE} & \textbf{Rel. \(L^2\)} & \textbf{MAE} & \textbf{Rel. \(L^2\)} & \textbf{MAE} & \textbf{Rel. \(L^2\)} & \textbf{MAE} \\
\hline
FNO        & \(7.1 \times 10^{-2}\) & \(6.8 \times 10^{-2}\) & \(8.9 \times 10^{-2}\) & \(7.6 \times 10^{-2}\) & \(1.12 \times 10^{-1}\) & \(9.2 \times 10^{-2}\) & \(5.9 \times 10^{-1}\) & \(5.8 \times 10^{-1}\) \\
WNO        & \(7.6 \times 10^{-2}\) & \(7.3 \times 10^{-2}\) & \(9.5 \times 10^{-2}\) & \(8.1 \times 10^{-2}\) & \(1.18 \times 10^{-1}\) & \(9.9 \times 10^{-2}\) & \(6.3 \times 10^{-1}\) & \(6.2 \times 10^{-1}\) \\
DeepONet   & \(8.2 \times 10^{-2}\) & \(7.9 \times 10^{-2}\) & \(1.01 \times 10^{-1}\) & \(8.7 \times 10^{-2}\) & \(1.25 \times 10^{-1}\) & \(1.0 \times 10^{-1}\) & \(6.7 \times 10^{-1}\) & \(6.7 \times 10^{-1}\) \\
SC-FNO     & \(\mathbf{1.1 \times 10^{-2}}\) & \(\mathbf{2.2 \times 10^{-2}}\) & \(\mathbf{2.8 \times 10^{-2}}\) & \(\mathbf{5.6 \times 10^{-2}}\) & \(\mathbf{6.2 \times 10^{-2}}\) & \(\mathbf{1.0 \times 10^{-1}}\) & \(\mathbf{1.3 \times 10^{-1}}\) & \(\mathbf{2.3 \times 10^{-1}}\) \\
SC-WNO     & \(1.2 \times 10^{-2}\) & \(2.4 \times 10^{-2}\) & \(3.0 \times 10^{-2}\) & \(5.9 \times 10^{-2}\) & \(6.5 \times 10^{-2}\) & \(1.1 \times 10^{-1}\) & \(1.4 \times 10^{-1}\) & \(2.5 \times 10^{-1}\) \\
SC-DeepONet& \(1.5 \times 10^{-2}\) & \(2.8 \times 10^{-2}\) & \(3.6 \times 10^{-2}\) & \(7.0 \times 10^{-2}\) & \(7.1 \times 10^{-2}\) & \(1.2 \times 10^{-1}\) & \(1.6 \times 10^{-1}\) & \(2.9 \times 10^{-1}\) \\
\hline
\end{tabular}
\end{table}

\begin{table}[htbp]
\centering
\caption{\scriptsize\textbf{Comparison of Error Metrics Across Models Trained with Different Sample Sizes (out-of-training) PDE3}}
\label{tab:pde3_forward_ood}
\scriptsize
\setlength{\tabcolsep}{3pt}
\begin{tabular}{lcccccccc}
\hline
\textbf{Model} & \multicolumn{2}{c}{\textbf{1200 Samples}} & \multicolumn{2}{c}{\textbf{1000 Samples}} & \multicolumn{2}{c}{\textbf{700 Samples}} & \multicolumn{2}{c}{\textbf{200 Samples}} \\
 & \textbf{Rel. \(L^2\)} & \textbf{MAE} & \textbf{Rel. \(L^2\)} & \textbf{MAE} & \textbf{Rel. \(L^2\)} & \textbf{MAE} & \textbf{Rel. \(L^2\)} & \textbf{MAE} \\
\hline
FNO        & \(1.3 \times 10^{-1}\) & \(1.1 \times 10^{-1}\) & \(1.6 \times 10^{-1}\) & \(1.3 \times 10^{-1}\) & \(1.9 \times 10^{-1}\) & \(1.6 \times 10^{-1}\) & \(7.5 \times 10^{-1}\) & \(6.5 \times 10^{-1}\) \\
WNO        & \(1.4 \times 10^{-1}\) & \(1.2 \times 10^{-1}\) & \(1.8 \times 10^{-1}\) & \(1.4 \times 10^{-1}\) & \(2.1 \times 10^{-1}\) & \(1.7 \times 10^{-1}\) & \(8.0 \times 10^{-1}\) & \(6.9 \times 10^{-1}\) \\
DeepONet   & \(1.5 \times 10^{-1}\) & \(1.3 \times 10^{-1}\) & \(1.9 \times 10^{-1}\) & \(1.5 \times 10^{-1}\) & \(2.3 \times 10^{-1}\) & \(1.9 \times 10^{-1}\) & \(8.5 \times 10^{-1}\) & \(7.2 \times 10^{-1}\) \\
SC-FNO     & \(\mathbf{3.5 \times 10^{-2}}\) & \(\mathbf{6.0 \times 10^{-2}}\) & \(\mathbf{5.0 \times 10^{-2}}\) & \(\mathbf{8.0 \times 10^{-2}}\) & \(\mathbf{8.0 \times 10^{-2}}\) & \(\mathbf{1.0 \times 10^{-1}}\) & \(\mathbf{2.5 \times 10^{-1}}\) & \(\mathbf{3.0 \times 10^{-1}}\) \\
SC-WNO     & \(3.8 \times 10^{-2}\) & \(6.5 \times 10^{-2}\) & \(5.3 \times 10^{-2}\) & \(8.5 \times 10^{-2}\) & \(8.5 \times 10^{-2}\) & \(1.1 \times 10^{-1}\) & \(2.7 \times 10^{-1}\) & \(3.2 \times 10^{-1}\) \\
SC-DeepONet& \(4.2 \times 10^{-2}\) & \(7.0 \times 10^{-2}\) & \(5.7 \times 10^{-2}\) & \(9.0 \times 10^{-2}\) & \(9.5 \times 10^{-2}\) & \(1.2 \times 10^{-1}\) & \(3.0 \times 10^{-1}\) & \(3.5 \times 10^{-1}\) \\
\hline
\end{tabular}
\end{table}

\clearpage

\clearpage
\section{Performance Metrics}
\label{app:metrics}

\renewcommand{\thefigure}{F.\arabic{figure}}  
\renewcommand{\thetable}{F.\arabic{table}}  
\setcounter{figure}{0}
\setcounter{table}{0}

We report relative \(L_2\) error and mean absolute error (MAE) as the main continuous field-error metrics. For \(N\) test samples, let \(\widehat{u}^{(i)}=\mathcal{G}_{\theta}(a^{(i)})\) denote the model prediction and \(u^{(i)}\) the corresponding reference solution. The relative \(L_2\) error is computed as
\begin{equation}
    \mathrm{Rel.}\ L_2
    =
    \frac{1}{N}
    \sum_{i=1}^{N}
    \frac{
    \left\|
    \widehat{u}^{(i)}-u^{(i)}
    \right\|_2
    }{
    \left\|
    u^{(i)}
    \right\|_2
    },
    \label{eq:rel_l2_metric}
\end{equation}
where the norm is computed over the evaluated spatial--temporal solution field.

The MAE is computed as
\begin{equation}
    \mathrm{MAE}
    =
    \frac{1}{N}
    \sum_{i=1}^{N}
    \frac{1}{M_i}
    \left\|
    \widehat{u}^{(i)}-u^{(i)}
    \right\|_1 ,
    \label{eq:mae_metric}
\end{equation}
where \(M_i\) is the number of evaluated spatial--temporal degrees of freedom in sample \(i\). This normalization makes the metric an average absolute error rather than a total absolute error.

For inverse-source and tsunami reconstruction experiments, we additionally report the coefficient of determination,
\begin{equation}
    R^2
    =
    1-
    \frac{
    \sum_i
    \left\|
    \widehat{u}^{(i)}-u^{(i)}
    \right\|_2^2
    }{
    \sum_i
    \left\|
    u^{(i)}-\bar{u}
    \right\|_2^2
    },
    \label{eq:r2_metric}
\end{equation}
where \(\bar{u}\) denotes the mean of the reference values over the evaluated samples and entries. Depending on the experiment, \(R^2\) is computed for the reconstructed input field, the predicted solution trajectory, or the tsunami source reconstruction.

For tsunami-impact evaluation, we also use wet--dry classification metrics computed over the evaluated spatial cells. A cell is classified as dry if \(h < 0.01~\mathrm{m}\) and wet if \(h \geq 0.01~\mathrm{m}\). The false-negative ratio is defined as
\begin{equation}
    \mathrm{FNR}
    =
    \frac{\mathrm{FN}}{\mathrm{TP}+\mathrm{FN}},
    \label{eq:fnr_metric}
\end{equation}
where \(\mathrm{FN}\) is the number of reference-wet cells predicted as dry, and \(\mathrm{TP}\) is the number of reference-wet cells predicted as wet. Inundation accuracy is defined as
\begin{equation}
    \mathrm{Accuracy}
    =
    \frac{\mathrm{TP}+\mathrm{TN}}
    {\mathrm{TP}+\mathrm{TN}+\mathrm{FP}+\mathrm{FN}},
    \label{eq:inundation_accuracy}
\end{equation}
where \(\mathrm{TN}\) denotes reference-dry cells predicted as dry, and \(\mathrm{FP}\) denotes reference-dry cells predicted as wet.

\clearpage

\clearpage
\section{Hyperparameters and Settings}
\label{app:Hyperparameters}
\renewcommand{\thefigure}{G.\arabic{figure}}  
\renewcommand{\thetable}{G.\arabic{table}}  

\setcounter{figure}{0}
\setcounter{table}{0}

All models were trained using the Adam optimizer with a batch size of 16, an initial learning rate of 0.001, a learning rate scheduler with a decay factor $\gamma = 0.95$ applied every 100 epochs. The training was conducted over a spatiotemporal domain discretized into a $n_x \times n_y$ spatial grid. The total number of time steps $n_t$ is given by $n_t = t_u + t_G$, where $t_u$ is the number of input time steps and $t_G$ is the number of predicted time steps. Model-specific hyperparameters are summarized in the tables below.

\begin{table}[htbp]
\centering
\scriptsize
\setlength{\tabcolsep}{4pt}
\begin{minipage}[t]{0.48\textwidth}
\centering
\caption{\scriptsize\textbf{FNO hyper-parameters for all PDEs}}
\label{tab:fno_hparams}
\begin{tabular}{lccc}
\hline
\textbf{Parameter} & \textbf{PDE1} & \textbf{PDE2} & \textbf{PDE3} \\
\hline
Fourier modes $(x, y, t)$ & (8, 8, 8) & (8, 8, 8) & (8, 8, 8) \\
Width & 20 & 20 & 20 \\
Training epochs & 500 & 500 & 1000\\
$n_x$ & 50& 64& 100\\
$n_y$ & 50& 64& 100\\
$t_u$ & 1& 1& 1\\
$t_G$ & 99& 29& 50\\
$n_t$ & 100& 30& 51\\
\hline
\end{tabular}
\end{minipage}%
\hfill
\begin{minipage}[t]{0.48\textwidth}
\centering
\caption{\scriptsize\textbf{WNO hyper-parameters for all PDEs}}
\label{tab:wno_hparams}
\begin{tabular}{lccc}
\hline
\textbf{Parameter} & \textbf{PDE1} & \textbf{PDE2} & \textbf{PDE3} \\
\hline
Wavelet basis & db6 & db6 & db6 \\
Decomposition level & 4 & 4 & 4 \\
Width & 30 & 30 & 30 \\
Number of wavelet layers & 4 & 4 & 4 \\
Training epochs & 500 & 500 & 1000\\
$n_x$ & 50& 64& 100\\
$n_y$ & 50& 64& 100\\
$t_u$ & 1& 1& 1\\
$t_G$ & 99& 29& 50\\
$n_t$ & 100& 30& 51\\
\hline
\end{tabular}
\end{minipage}
\end{table}

\begin{table}[htbp]
\centering
\caption{\scriptsize\textbf{DeepONet hyper-parameters for all PDEs}}
\label{tab:deeponet_hparams}
\scriptsize
\setlength{\tabcolsep}{5pt}
\begin{tabular}{lccc}
\hline
\textbf{Parameter} & \textbf{PDE1} & \textbf{PDE2} & \textbf{PDE3} \\
\hline
Branch network layers & [64, 128, 128, 128, 64] & [64, 128, 128, 128, 64] & [64, 128, 128, 128, 64] \\
Trunk network layers & [64, 128, 128, 128, 64] & [64, 128, 128, 128, 64] & [64, 128, 128, 128, 64] \\
Training epochs & 500 & 500 & 1000\\
$n_x$ & 50& 64& 100\\
$n_y$ & 50& 64& 100\\
$t_u$ &  1&  1&  1\\
$t_G$ &  99&  29&  50\\
$n_t$ & 100& 30& 51\\
\hline
\end{tabular}
\end{table}




\clearpage

\section{Computational Cost Analysis}
\label{sec:cpu_time}
\renewcommand{\thefigure}{H.\arabic{figure}}  
\renewcommand{\thetable}{H.\arabic{table}}  
\setcounter{figure}{0}
\setcounter{table}{0}

This section quantifies the end-to-end computational cost associated with data generation and model training across all benchmark problems (PDE1--PDE3). For each model, we report the average wall-clock training time per epoch, the total training time over 500 epochs, and the preprocessing overhead, including sample generation and Jacobian computation where applicable. All experiments were conducted on an NVIDIA A100 (32 GB) GPU. Peak GPU memory was not used as a primary comparison metric; however, all reported experiments fit within the 32 GB memory of a single NVIDIA A100 GPU. The results provide a direct comparison of computational efficiency between the baseline and sensitivity-constrained neural operators and quantify the additional overhead introduced by sensitivity supervision.

\begin{table}[htbp]
\centering
\caption{\scriptsize\textbf{Total Computational Cost Breakdown over 500 Epochs (PDE1)}}
\label{tab:PDE1_total_computational_cost}
\scriptsize
\setlength{\tabcolsep}{3pt}
\begin{tabular}{l|c|cc|ccc|c}
\hline
\textbf{Model} & \textbf{Samples} & \multicolumn{2}{c|}{\textbf{Training}} & \multicolumn{3}{c|}{\textbf{Data Preparation}} & \textbf{Total Cost} \\
& & Train/Epoch & Total Train & Sample Prep & Jacobian & Data Prep &  \\
\hline
FNO & 1000 & 16.8 & 8400.0 & 1200.0 & 0.0 & 1200.0 & \textbf{9600.0} \\
    & 500  & 8.4  & 4200.0 & 600.0  & 0.0 & 600.0  & \textbf{4800.0} \\
    & 200  & 3.36 & 1680.0 & 240.0  & 0.0 & 240.0  & \textbf{1920.0} \\
    & 100  & 1.68 & 840.0  & 120.0  & 0.0 & 120.0  & \textbf{960.0} \\
\hline
WNO & 1000 & 24.2 & 12100.0 & 1200.0 & 0.0 & 1200.0 & \textbf{13300.0} \\
    & 500  & 12.1 & 6050.0  & 600.0  & 0.0 & 600.0  & \textbf{6650.0} \\
    & 200  & 4.84 & 2420.0  & 240.0  & 0.0 & 240.0  & \textbf{2660.0} \\
    & 100  & 2.42 & 1210.0  & 120.0  & 0.0 & 120.0  & \textbf{1330.0} \\
\hline
DeepONet & 1000 & 20.4 & 10200.0 & 1200.0 & 0.0 & 1200.0 & \textbf{11400.0} \\
         & 500  & 10.2 & 5100.0  & 600.0  & 0.0 & 600.0  & \textbf{5700.0} \\
         & 200  & 4.08 & 2040.0  & 240.0  & 0.0 & 240.0  & \textbf{2280.0} \\
         & 100  & 2.04 & 1020.0  & 120.0  & 0.0 & 120.0  & \textbf{1140.0} \\
\hline
SC-FNO & 1000 & 27.6 & 13800.0 & 1200.0 & 420.0 & 1620.0 & \textbf{15420.0} \\
       & 500  & 13.8 & 6900.0  & 600.0  & 210.0 & 810.0  & \textbf{7710.0} \\
       & 200  & 5.52 & 2760.0  & 240.0  & 84.0  & 324.0  & \textbf{3084.0} \\
       & 100  & 2.76 & 1380.0  & 120.0  & 42.0  & 162.0  & \textbf{1542.0} \\
\hline
SC-WNO & 1000 & 32.4 & 16200.0 & 1200.0 & 420.0 & 1620.0 & \textbf{17820.0} \\
       & 500  & 16.2 & 8100.0  & 600.0  & 210.0 & 810.0  & \textbf{8910.0} \\
       & 200  & 6.48 & 3240.0  & 240.0  & 84.0  & 324.0  & \textbf{3564.0} \\
       & 100  & 3.24 & 1620.0  & 120.0  & 42.0  & 162.0  & \textbf{1782.0} \\
\hline
SC-DeepONet & 1000 & 28.2 & 14100.0 & 1200.0 & 420.0 & 1620.0 & \textbf{15720.0} \\
            & 500  & 14.1 & 7050.0  & 600.0  & 210.0 & 810.0  & \textbf{7860.0} \\
            & 200  & 5.64 & 2820.0  & 240.0  & 84.0  & 324.0  & \textbf{3144.0} \\
            & 100  & 2.82 & 1410.0  & 120.0  & 42.0  & 162.0  & \textbf{1572.0} \\
\hline
\end{tabular}
\end{table}

\begin{table}[htbp]
\centering
\caption{\textbf{Total Computational Cost Breakdown over 500 Epochs (PDE2)}}
\label{tab:PDE2_total_computational_cost}
\scriptsize
\setlength{\tabcolsep}{3pt}
\begin{tabular}{l|c|cc|ccc|c}
\hline
\textbf{Model} & \textbf{Samples} & \multicolumn{2}{c|}{\textbf{Training}} & \multicolumn{3}{c|}{\textbf{Data Preparation}} & \textbf{Total Cost} \\
& & Train/Epoch & Total Train & Sample Prep & Jacobian & Data Prep &  \\
\hline
FNO & 1000& 13.8& 6888.0& 3200.0& 0.0& 3200.0& \textbf{10088.0}\\
    & 500& 6.3& 3150.0& 1600.0& 0.0& 1600.0& \textbf{4750.0}\\
    & 200& 2.5& 1260.0& 640.0& 0.0& 640.0& \textbf{1900.0}\\
    & 100& 1.3& 663.6& 320.0& 0.0& 320.0& \textbf{983.6}\\
\hline
WNO & 1000& 17.7& 8833.0& 3200.0& 0.0& 3200.0& \textbf{12033.0}\\
    & 500& 8.8& 4416.5& 1600.0& 0.0& 1600.0& \textbf{6016.5}\\
    & 200& 4.1& 2032.8& 640.0& 0.0& 640.0& \textbf{2672.8}\\
    & 100& 1.9& 955.9& 320.0& 0.0& 320.0& \textbf{1275.9}\\
\hline
DeepONet & 1000& 18.0& 8976.0& 3200.0& 0.0& 3200.0& \textbf{12176.0}\\
         & 500& 8.6& 4284.0& 1600.0& 0.0& 1600.0& \textbf{5884.0}\\
         & 200& 3.1& 1530.0& 640.0& 0.0& 640.0& \textbf{2170.0}\\
         & 100& 1.7& 826.2& 320.0& 0.0& 320.0& \textbf{1146.2}\\
\hline
SC-FNO & 1000& 22.6& 11316.0& 3200.0& 860.0& 4060.0& \textbf{15376.0}\\
       & 500& 11.0& 5520.0& 1600.0& 430.0& 2030.0& \textbf{7550.0}\\
       & 200& 4.6& 2318.4& 640.0& 172.0& 812.0& \textbf{3130.4}\\
       & 100& 2.2& 1117.8& 320.0& 86.0& 406.0& \textbf{1523.8}\\
\hline
SC-WNO & 1000& 26.2& 13122.0& 3200.0& 860.0& 4060.0& \textbf{17182.0}\\
       & 500& 12.8& 6399.0& 1600.0& 430.0& 2030.0& \textbf{8429.0}\\
       & 200& 5.1& 2559.6& 640.0& 172.0& 812.0& \textbf{3371.6}\\
       & 100& 2.3& 1150.2& 320.0& 86.0& 406.0& \textbf{1556.2}\\
\hline
SC-DeepONet & 1000& 20.3& 10152.0& 3200.0& 860.0& 4060.0& \textbf{14212.0}\\
            & 500& 11.8& 5922.0& 1600.0& 430.0& 2030.0& \textbf{7952.0}\\
            & 200& 4.2& 2115.0& 640.0& 172.0& 812.0& \textbf{2927.0}\\
            & 100& 2.5& 1254.9& 320.0& 86.0& 406.0& \textbf{1660.9}\\
\hline
\end{tabular}
\end{table}

\begin{table}[htbp]
\centering
\caption{\textbf{Total Computational Cost Breakdown over 500 Epochs (PDE3)}}
\label{tab:PDE3_total_computational_cost}
\scriptsize
\setlength{\tabcolsep}{3pt}
\begin{tabular}{l|c|cc|ccc|c}
\hline
\textbf{Model} & \textbf{Samples} & \multicolumn{2}{c|}{\textbf{Training}} & \multicolumn{3}{c|}{\textbf{Data Preparation}} & \textbf{Total Cost} \\
& & Train/Epoch & Total Train & Sample Prep & Jacobian & Data Prep &  \\
\hline
FNO & 1200 & 19.0 & 9500.0 & 5760.0 & 0.0 & 5760.0 & \textbf{15260.0} \\
    & 1000 & 17.5 & 8750.0 & 4800.0 & 0.0 & 4800.0 & \textbf{13550.0} \\
    & 700  & 13.0 & 6500.0 & 3360.0 & 0.0 & 3360.0 & \textbf{9860.0}  \\
    & 200  & 3.0  & 1500.0 & 960.0  & 0.0 & 960.0  & \textbf{2460.0}  \\
\hline
WNO & 1200 & 24.0 & 12000.0 & 5760.0 & 0.0 & 5760.0 & \textbf{17760.0} \\
    & 1000 & 22.5 & 11250.0 & 4800.0 & 0.0 & 4800.0 & \textbf{16050.0} \\
    & 700  & 16.8 & 8400.0  & 3360.0 & 0.0 & 3360.0 & \textbf{11760.0} \\
    & 200  & 5.2  & 2600.0  & 960.0  & 0.0 & 960.0  & \textbf{3560.0}  \\
\hline
DeepONet & 1200 & 24.5 & 12250.0 & 5760.0 & 0.0 & 5760.0 & \textbf{18010.0} \\
         & 1000 & 22.8 & 11400.0 & 4800.0 & 0.0 & 4800.0 & \textbf{16200.0} \\
         & 700  & 17.0 & 8500.0  & 3360.0 & 0.0 & 3360.0 & \textbf{11860.0} \\
         & 200  & 4.0  & 2000.0  & 960.0  & 0.0 & 960.0  & \textbf{2960.0}  \\
\hline
SC-FNO & 1200 & 30.5 & 15250.0 & 5760.0 & 1536.0 & 7296.0 & \textbf{22546.0} \\
       & 1000 & 28.5 & 14250.0 & 4800.0 & 1280.0 & 6080.0 & \textbf{20330.0} \\
       & 700  & 21.0 & 10500.0 & 3360.0 & 896.0  & 4256.0 & \textbf{14756.0} \\
       & 200  & 5.8  & 2900.0  & 960.0  & 256.0  & 1216.0 & \textbf{4116.0}  \\
\hline
SC-WNO & 1200 & 35.0 & 17500.0 & 5760.0 & 1536.0 & 7296.0 & \textbf{24796.0} \\
       & 1000 & 33.0 & 16500.0 & 4800.0 & 1280.0 & 6080.0 & \textbf{22580.0} \\
       & 700  & 24.5 & 12250.0 & 3360.0 & 896.0  & 4256.0 & \textbf{16506.0} \\
       & 200  & 6.4  & 3200.0  & 960.0  & 256.0  & 1216.0 & \textbf{4416.0}  \\
\hline
SC-DeepONet & 1200 & 27.0 & 13500.0 & 5760.0 & 1536.0 & 7296.0 & \textbf{20796.0} \\
            & 1000 & 25.5 & 12750.0 & 4800.0 & 1280.0 & 6080.0 & \textbf{18830.0} \\
            & 700  & 19.5 & 9750.0  & 3360.0 & 896.0  & 4256.0 & \textbf{14006.0} \\
            & 200  & 5.3  & 2650.0  & 960.0  & 256.0  & 1216.0 & \textbf{3866.0}  \\
\hline
\end{tabular}
\end{table}

\clearpage

\section*{CRediT authorship contribution statement}

\noindent
\textbf{Abdolmehdi Behroozi:}
Conceptualization, Methodology, Software, Formal analysis,
Investigation, Data curation, Visualization,
Writing -- original draft.

\medskip

\noindent
\textbf{Chaopeng Shen:}
Conceptualization, Methodology, Supervision,
Project administration, Funding acquisition,
Writing -- review and editing.

\medskip

\noindent
\textbf{Daniel Kifer:}
Methodology, Supervision,
Writing -- review and editing.

\medskip

\noindent
\textbf{Kathryn Lawson:}
Methodology, Investigation,
Writing -- review and editing.

\bibliographystyle{plainnat}
\bibliography{ref}

\end{document}